\documentclass[preprints,article,accept,moreauthors,pdftex,letterpaper]{Definitions/mdpi} 

\firstpage{1} 
\pubvolume{xx}
\issuenum{1}
\articlenumber{5}
\pubyear{2020}
\copyrightyear{2020}

\history{Received: date; Accepted: date; Published: date}

\usepackage{amsmath} 
\usepackage{amssymb}  
\usepackage[linesnumbered,ruled]{algorithm2e}
\usepackage{booktabs}
\usepackage{subcaption}
\usepackage[tableposition=top]{caption}
\graphicspath{{Figures/}}
\usepackage{mathtools}
\usepackage{enumitem}
\setlist[enumerate]{label*=\arabic*.}

\usepackage[utf8]{inputenc}
\usepackage{pgfplots}
\DeclareUnicodeCharacter{2212}{−}
\usepgfplotslibrary{groupplots,dateplot}
\usetikzlibrary{patterns,shapes.arrows}
\pgfplotsset{compat=newest}
\usepackage{adjustbox}

\Title{Polylidar3D - Fast Polygon Extraction from 3D Data}

\Author{Jeremy Castagno $^{1,*}$\orcidA and Ella Atkins  $^{2}$\orcidB}

\AuthorNames{Jeremy Castagno, Ella Atkins}

\address{%
$^{1}$ \quad University of Michigan; jdcasta@umich.edu\\
$^{2}$ \quad University of Michigan; ematkins@umich.edu}

\corres{Correspondence: jdcasta@umich.edu}

\abstract{Flat surfaces captured by 3D point clouds are often used for localization, mapping, and modeling.  Dense point cloud processing has high computation and memory costs making low-dimensional representations of flat surfaces such as polygons desirable.  We present Polylidar3D, a non-convex polygon extraction algorithm which takes as input unorganized 3D point clouds (e.g.,  LiDAR data), organized point clouds (e.g., range images), or user-provided meshes. Non-convex polygons represent flat surfaces in an environment with interior cutouts representing obstacles or holes. The Polylidar3D front-end transforms input data into a half-edge triangular mesh.  This representation provides a common level of input data abstraction for subsequent back-end processing. The Polylidar3D back-end is composed of four core algorithms: mesh smoothing, dominant plane normal estimation, planar segment extraction, and finally polygon extraction.  Polylidar3D is shown to be quite fast, making use of CPU multi-threading and GPU acceleration when available. We demonstrate Polylidar3D's versatility and speed with real-world datasets including aerial LiDAR point clouds for rooftop mapping, autonomous driving LiDAR point clouds for road surface detection, and RGBD cameras for indoor floor/wall detection. We also evaluate Polylidar3D on a challenging planar segmentation benchmark dataset.  Results consistently show excellent speed and accuracy. }

\begin{document}



\section{Introduction}
Flat surfaces are pervasive in engineered structures and also occur in natural terrain. For example, structures such as walls, floors, rooftops, and roadways are often flat or "flat-like". Similarly, home and office furnishings are typically composed of multiple flat surfaces. Sensors such as LiDAR and RGBD cameras generate dense 3D point clouds of these predominately flat surface environments. This observation has been exploited for tasks in localization and mapping \cite{pathak_online_2010}, 3D building modelling \cite{2016ISPAr49B3}, and point cloud registration \cite{rusinkiewicz_efficient_2001}. Planar segmentation techniques are often used to group points together belonging to a flat surface \cite{feng_fast_2014, pham_geometrically_2016, schaefer_maximum_2019}. However points clouds are dense incurring a high computational cost when used directly in higher level tasks. Planar point clouds can be converted to lower dimensional representations such as polygons. Polygons reduce map size, accelerate matching for localization \cite{lee_indoor_2012}, and support model reconstruction and object detection \cite{cao_roof_2017}. 

Planar points clouds may be converted to convex polygons \cite{biswas_planar_2012}. Convex polygons are simple and efficient to generate but often do not represent the true shape of a point set. Non-convex polygons may be generated using techniques such as $\alpha$-shapes but operate strictly on 2D data, requiring the projection of each 3D planar point cloud and expensive triangulation \cite{lee_fast_2013, 1056714}. Pixel-level boundary following of organized point clouds can be used to extract non-convex polygons but often only captures the exterior shell of the polygon \cite{lee_indoor_2012}. These methods are not able to capture \emph{interior} holes in a polygon representing the shape of obstacles on flat surfaces. Finally, speed is an important consideration for many of the applications mentioned previously. Parallel algorithms written for multi-core CPUs and GPUs should be used to reduce latency. 

\begin{figure}[ht]
    \centering
    \includegraphics[width=.95\linewidth]{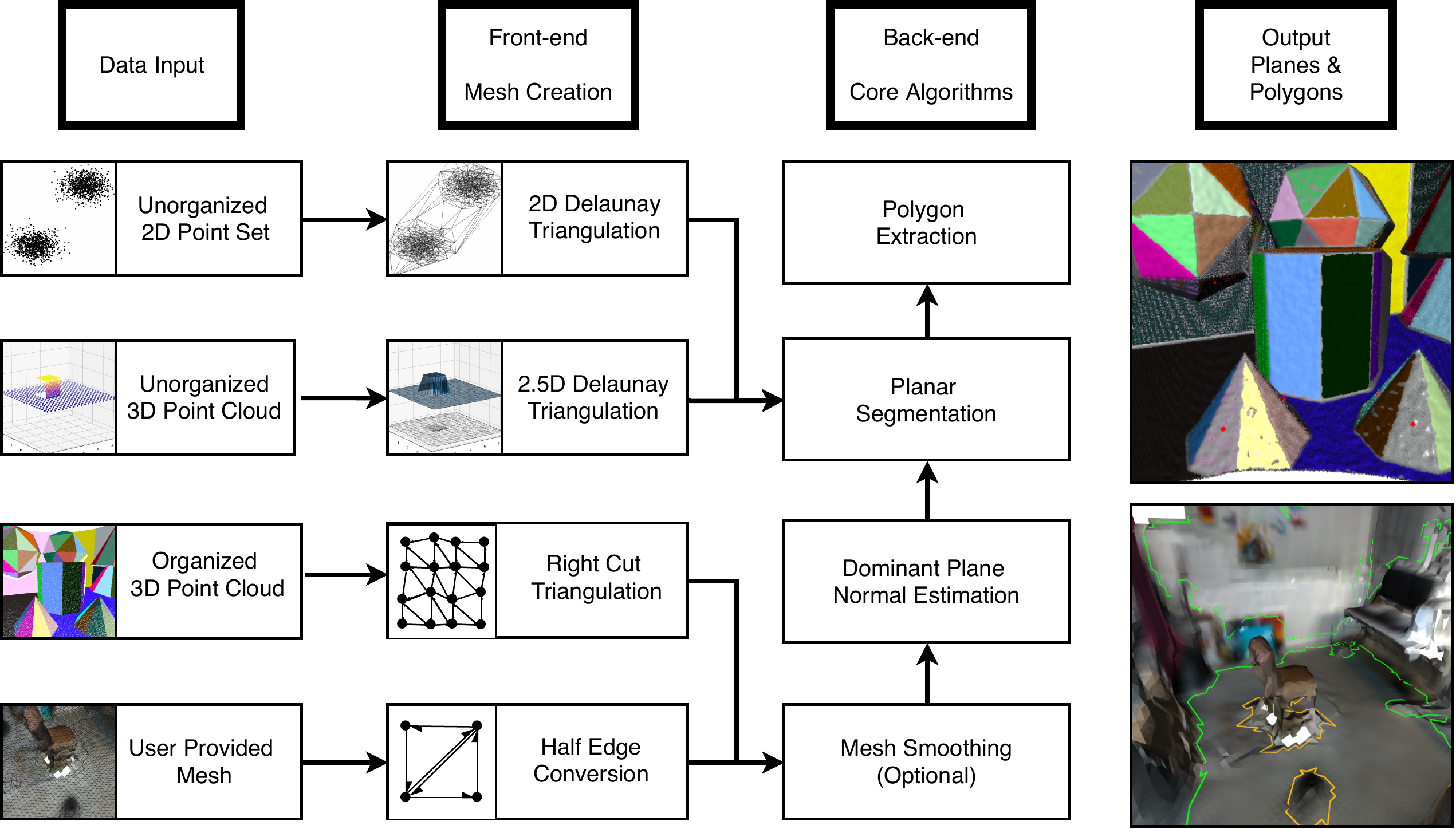}
    \caption{Overview of Polylidar3D. Input data can be 2D point sets, unorganized/organized 3D point clouds, or user-provided meshes. Polylidar3D's front-end transforms input data to a half-edge triangulation structure. The back-end is responsible for mesh smoothing, dominant plane normal estimation, planar segmentation, and polygon extraction. Polylidar3D outputs both planes (sets of spatially connected triangles) and corresponding polygonal representations. An example output of color-coded extracted planes from organized point clouds is shown (top right). An example of extracted polygons from a user-provided mesh is shown (bottom right). The green line represents the concave hull; orange lines show interior holes representing obstacles.} 
    \label{fig:polylidar_overview}
\end{figure}

We present Polylidar3D, a non-convex polygon extraction algorithm which takes as input either unorganized 3D point clouds (e.g., airborne LiDAR point clouds), organized point clouds (e.g., range images), or user provided meshes. The non-convex polygons extracted represent flat surfaces in an environment, while interior holes represent obstacles on these surfaces.  Figure \ref{fig:polylidar_overview} provides an overview of Polylidar3D's data input, front-end, back-end, and output. The front-end transforms input data into a half-edge triangular mesh.  This representation provides a common level of abstraction offering increased efficiency for back-end operations. The back-end is composed of four core algorithms: mesh smoothing, dominant plane normal estimation, planar segment extraction, and polygon extraction.  Polylidar3D outputs planar triangular segments, sets of flat connected triangles, and their polygonal representations. Polylidar3D is extremely fast, typically executing in a few milliseconds. It makes use of CPU multi-threading and GPU acceleration when available. Polylidar3D is a substantial extension to Polylidar, the authors' 2D algorithm that transforms 2D point sets into polygons \cite{polylidar2D}. Our previous Polylidar algorithm only operated on 2D point sets and offered no parallelism. 
The primary contributions of this paper are:

\begin{itemize}
  \item An efficient and versatile open source \cite{polylidarcode} framework for concave (multi)polygon extraction for 3D data. Input can be unorganized/organized 3D point clouds or user-provided meshes.
  \item A fast open source \cite{fastgacode} dominant plane normal estimation procedure using a Gaussian Accumulator that can also be used as a stand-alone algorithm.
  \item Multiple diverse open source experiments showing qualitative and quantitative benchmark results from data sources including LiDAR and RGBD cameras \cite{polylidar_kitti, polylidar_realsense, polylidar_synpeb}.
  \item Improved half-edge triangulation efficiency for organized point clouds; CPU multi-threaded and GPU accelerated mesh smoothing \cite{fastgacode}. 
  \item Planar segmentation and polygon extraction performed in tandem using task-based parallelism to reduce latency for time-critical applications. 
\end{itemize}

Below, Sections \ref{sec:background} and \ref{sec:prelim} provide background and mathematical preliminaries, respectively. Section \ref{sec:methods_mesh_creation} describes Polylidar3D's front-end methods for mesh creation. Section \ref{sec:methods_mesh_smoothing} outlines optional mesh smoothing while Section \ref{sec:methods_fastga} introduces our dominant plane normal estimation algorithm. Section \ref{sec:methods_polylidar} describes plane and polygon extraction with parallelization techniques. Section \ref{sec:methods_polylidar_polygon_filtering} proposes optional post-processing methods to refine and simplify the polygons.  Section \ref{sec:results} provides qualitative results as well as quantitative benchmarks. Sections \ref{sec:discussion} and \ref{sec:conclusion} provide discussion and conclusion, respectively.

\section{Background}\label{sec:background}

This section summarizes baseline methods on which Polylidar3D is constructed.  Plane segmentation and polygon extraction background is followed by a description of 3D data denoising techniques such as mesh smoothing and plane normal estimation with Gaussian accumulators.

\subsection{Planar Segmentation}\label{sec:bg_plane_extraction}

Planar segmentation processes an input 3D point cloud and segments it into groups of points representing flat surfaces. These point groups are often informally called  ``planes'' but differ from the geometric definition.  A geometric plane is defined by a unit normal $\hat{n} \in \mathbb{R}^3$ and a single point on the plane $p \in \mathbb{R}^3$. Flat surface representation as a point set is advantageous because:
\begin{enumerate}
    \item Point sets are naturally bounded (i.e.,  have finite extent). Bounded surfaces better correspond with most real-world flat surfaces.
    \item Holes inside a plane may be represented implicitly by the absence of points. This representation can also indicate obstacles embedded in a flat surface.
    \item Best-fit geometric planes can also be computed after segmentation using least-squares, principal component analysis (PCA), or RANSAC based methods \cite{kaiser_survey_2019}.
    \item Merging similar planes can be rapidly performed by combining their points sets.
\end{enumerate}

Planar segmentation can be performed with region growing methods.  Algorithms can exploit the spatial structure of an organized point cloud for which data is arranged into rows/columns like an image (e.g., range images).  Region growing algorithms extract connected components in point clouds with neighborhood information (e.g., pixel neighbors or $k$ nearest neighbors). A seed is chosen and assigned a unique label, then its neighbors are iteratively analyzed and assigned the seed’s label if their characteristics are sufficiently similar to the seed's \cite{kaiser_survey_2019}. Characteristics such as normal orientation, color, or Euclidean distance from each other may be used. A unique label assigned to each point denotes a grouping in a planar surface.

Ref. \cite{lee_fast_2013} outlines planar segmentation by employing approximate polygonal meshing from organized point clouds. Mesh construction exploits the organized structure of a range image. Point normals are computed from the mesh and smoothed using bilateral filtering techniques which preserve edges. Region growing is performed sequentially until all possible points have been examined. Points are merged based on differences in normal angles and Euclidean distances. Reported benchmarks show that a 320X240 range image can be segmented in approximately 125 ms. 
Work by Ref. \cite{feng_fast_2014} proposes the use of agglomerative hierarchical clustering (AHC) on organized point clouds to perform fast planar segmentation.  The algorithm first creates a graph by uniformly dividing the points in image space. Initial node size (e.g., 4X4 pixel group) is user-configurable and allows a trade-off between execution speed and the detail of extracted planes. Nodes belonging to the same plane are merged through AHC until plane fitting error exceeds a user-defined threshold. A final refinement is done though pixel-wise region growing with possible plane merging.  The open-source algorithm is extremely fast; a 640X480 image with an initial node size of 10X10 can be segmented in $\approx$30ms.

Ref. \cite{schaefer_maximum_2019} details a probabilistic plane extraction (PPE) algorithm to detect planes in organized 3D laser range scans. The algorithm utilizes AHC with individual laser reflection in the scene to define an initial candidate plane set. Each plane is then iteratively merged with adjacent planes that maximize the measurement likelihood of the scan. Measurement likelihood is computed using a Gaussian probability density function modelling ray length. The algorithm is implemented in MATLAB with GPU acceleration but has execution times exceeding one hour for a 500X500 scan.  Work by Ref. \cite{trevor2013efficient} similarly operates on organized point clouds and exploits neighbor information for merging. Surface normals are estimated for every point using methods in \cite{lee_fast_2013}.  Points are merged if their normals and orthogonal distances are below a threshold which creates locally planar segments later refined through plane fitting and filtering out segments that exceed curvature constraints.

Ref. \cite{salas-moreno_dense_2014} outlines a fast GPU accelerated planar segmentation method for use in simultaneous localization and mapping (SLAM) from range images. A range image is first converted into surfels (surface elements) describing a point position and orientation. A surfel similarity bit mask is created and marked 1 if neighbor surfels to the left and above are similar in normals and plane distance, 0 otherwise.  This bit mask is used to perform region growing where similar and contiguous pixels are merged to the same plane. The algorithm executes in real-time and can perform planar extraction with SLAM in 66 ms.
Ref. \cite{oesau_planar_2016} presents a parallel plane extraction method specifically designed for unorganized point clouds.  First, points are organized using an octree from which multiple seed points are uniformly selected. Region growing occurs in parallel for each seed point with points inside the same cell in the octree used for plane fitting. Region growing is periodically interrupted to perform regularization of the planar shapes detected. Regularization is carried out by merging planes captured on the same flat surface by refitting through PCA. The implementation is GPU accelerated and can segment point clouds with 1.1 million points in approximately three seconds. 

Polylidar3D segments points clouds through region growing but operates on triangles instead of the points themselves. Region growing is regularized and parallelized by first identifying dominant plane normals in the mesh. Triangles having similar normals to a dominant plane are grouped. Each group (in parallel) then performs region growing accounting for normal orientation, Euclidean distance, and point to plane distance. Note this method relies upon the data to be properly denoised.



\subsection{Polygonal Shape Extraction}\label{sec:bg_shape_extraction}

\begin{figure}[ht]

  \begin{subfigure}{.24\linewidth}
    \centering\includegraphics[width=.95\linewidth]{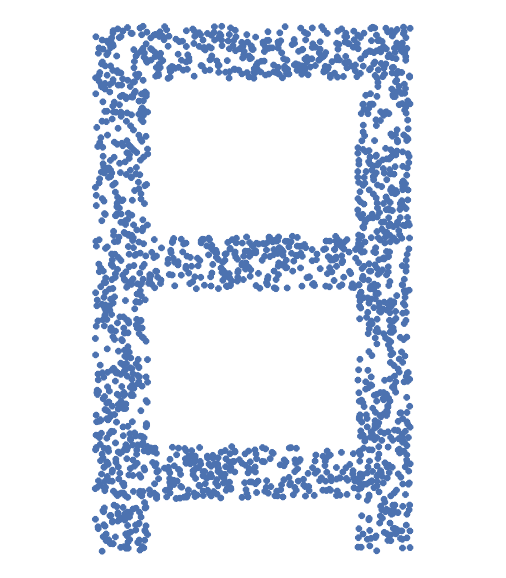}
    \caption{\label{fig:convex_concave_1}2D Point Set}
  \end{subfigure}
  \begin{subfigure}{.24\linewidth}
    \centering\includegraphics[width=.95\linewidth]{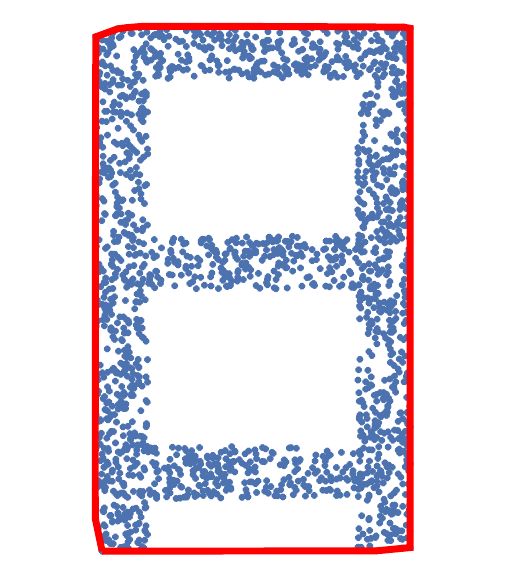}
    \caption{\label{fig:convex_concave_2}Convex}
  \end{subfigure}
  \begin{subfigure}{.24\linewidth}
    \centering\includegraphics[width=.95\linewidth]{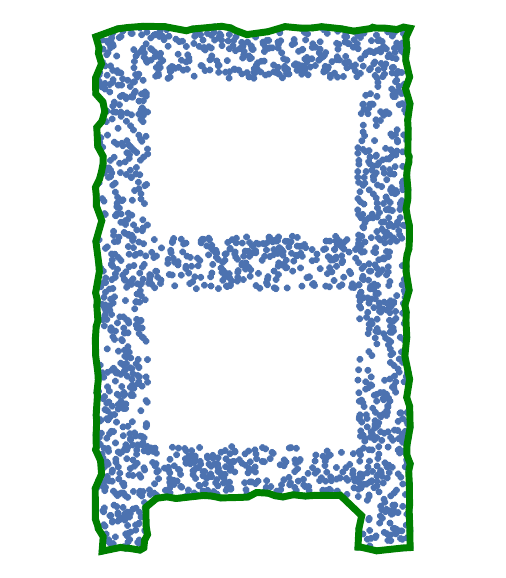}
    \caption{\label{fig:convex_concave_3}Non-convex}
  \end{subfigure}
  \begin{subfigure}{.24\linewidth}
    \centering\includegraphics[width=.95\linewidth]{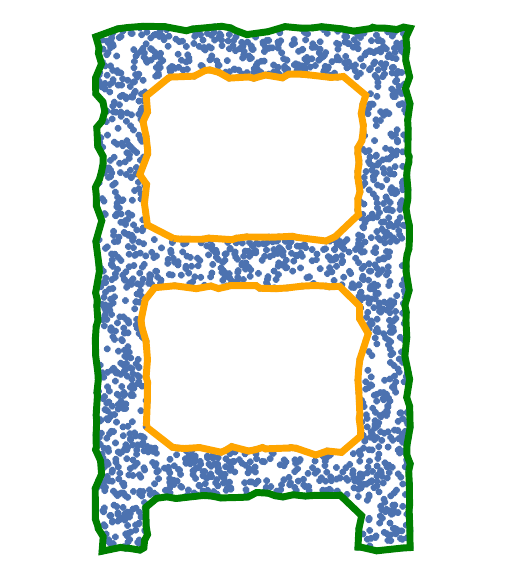}
    \caption{\label{fig:convex_concave_4}Non-convex with holes}
  \end{subfigure}
  \caption{Example polygons that can be generated from plane segmented point clouds. (\subref{fig:convex_concave_1}) 2D point set representation of a floor diagram with interior offices.  (\subref{fig:convex_concave_2}) Convex polygon, (\subref{fig:convex_concave_3}) non-convex polygon, (\subref{fig:convex_concave_4}) and non-convex polygon with holes. The exterior hull (green) and interior holes (orange) are indicated. }\label{fig:convex_concave}
\end{figure}

Representing planar surfaces as point sets has the disadvantage of high memory and computational overhead.  Dense planar point sets have redundant information about the underlying surface they represent. Polygonal representations of point sets removes redundant information. We consider convex polygons, non-convex polygons, and non-convex polygons with holes per Figure \ref{fig:convex_concave}.

Ref. \cite{biswas_planar_2012} represents flat surfaces as convex polygons extracted from range images. First, points are randomly sampled in the image with nearby pixel neighbors used for fast RANSAC plane fitting. This returns numerous sparse point subsets which may be coplanar. The convex hull is computed for each of these subsets generating many convex polygons in the scene. Polygons belonging to the same surface are then merged with GPU accelerated correspondence matching. The sparse random sampling of the point cloud and efficient generation of convex hulls allows the algorithm to run in real-time (less than 2 ms). However, convex hulls ignore boundary concavities, overestimate the area of the enclosed point set, and do no not account for holes per Figure \ref{fig:convex_concave_2}. Ref. \cite{poppinga_fast_2008} outlines a  method to convert plane segmented range images into convex polygons. 
Each plane segment is decomposed into  a set of convex polygons. Each polygon is progressively built through scan-lines; a new polygon is generated when convexity constraints are not met. This allows a concave plane to be represented by multiple convex polygons. 
Ref. \cite{lee_indoor_2012} generates non-convex polygons from range images. The range image is first planar-segmented using an eight-way flood fill algorithm. This involves region growing which accounts for the normal vector to each point. Each of the planar segments is then converted to a non-convex polygon. Exterior boundary pixels of a plane segment are sampled and neighboring samples connected to create a non-convex polygon. However interior holes in the plane segments are not explicitly captured as shown in \ref{fig:convex_concave_3}. Ref. \cite{trevor2013efficient} performs a similar polygon extraction procedure through boundary tracing of the exterior hull. 

Non-convex polygons with holes may be generated through a variety of methods, many under the name of concave hulls \cite{1056714, spatialite, postgis}. Many of these methods strictly operate on 2D data, requiring the 3D planar point cloud segments be projected to the best fit geometric plane to produce 2D point sets. Ref. \cite{lee_fast_2013} proposes this technique and the use of $\alpha$-shapes to extract such polygons \cite{1056714}.  We developed a faster open source polygon extraction algorithm, Polylidar, which extracts non-convex polygons with holes from 2D point sets \cite{polylidar2D}. The point set is converted to a 2D mesh through Delaunay triangulation, and triangles are subsequently filtered by edge length creating the ``shape'' of the point set. This filtered mesh is then converted to a polygon through boundary following while accounting for holes. Benchmarks demonstrate that our algorithm is a minimum of four times faster than leading methods  \cite{polylidar_benchmark_concave}. This paper extends Polylidar to operate directly on 3D data, performing planar segmentation and polygon extraction in parallel. This integration allows Polylidar3D to skip expensive Delaunay triangulation previously required for organized point clouds.  Planar segments represented as non-convex polygons with holes gives the following advantages:

\begin{enumerate}
    \item Significantly reduced memory requirements, on the order of square root (perimeter vs area).
    \item Faster computation of geometric values of interest, e.g., centroid, area, perimeter.
    \item Ability to dilate, erode, and simplify polygons through computational geometry routines.
    \item Holes inside a polygon account for gaps or obstacles on flat surfaces.
\end{enumerate}

\subsection{3D Data Denoising}\label{sec:bg_mesh_smoothing}


This section discusses two methods to smooth a triangular mesh. The Laplacian filter performs weighted averaging of nearby vertex neighbors to reduce noise \cite{taubin_curve_1995}. Vertices are updated according to 
\begin{align}
  v_o = v_i \cdot \frac{\lambda}{W} \sum_{j=1}^{N} w_j \cdot (v_i - v_j) \label{eq:laplacian_vertex}  \\
  w_j = ||v_i - v_j||^{-1} \quad \label{eq:laplacian_weight}
  W = \sum_{j=1}^{N} w_j
\end{align}
where $v_o$, $v_i$, $v_j$, and $N$ denote the output (smoothed) vertex, input vertex, neighboring vertex, and total number of neighbors, respectively. Weighting for each neighbor vertex $w_j$ is the inverse of its Euclidean distance and is normalized with $W$.  Parameter $\lambda$ adjusts smoothing [0-1], though multiple iterations may be performed to increase smoothing. The Laplacian filter is not edge-preserving. 

Ref. \cite{zheng_bilateral_2011} proposes a bilateral filtering technique on triangular meshes that is analogous to images.   Filtering occurs in two stages: normal smoothing and vertex updating. The first stage performs local iterative normal filtering to smooth normals but preserves edges as given by:

\begin{align}
  n_o = K \sum_{j=1}^{N} W_c(||c_i - c_j||) \cdot W_s(||n_i - n_j||) \cdot n_j \label{eq:bilateral} \\
  W_c(||c_i - c_j||) = \operatorname{exp}(-||c_i - c_j||^2/2\sigma_c^2 ) \label{eq:bilateral_centroid} \\
  W_s(||n_i - n_j||) = \operatorname{exp}(-||n_i - n_j||^2/2\sigma_s^2 ) \label{eq:bilateral_normal}\\
  K = 1 / \sum_{j=1}^{N} W_c(||c_i - c_j||) \cdot W_s(||n_i - n_j||)
\end{align}
where $n_o$, $n_i$, $c_i$, $n_j$, $c_j$, and $N$ denote smoothed triangle normal, input normal, centroid, neighbor normal, neighbor centroid, and number of neighbors, respectively. Triangle weights $W_c$ and $W_s$ exponentially decay based upon deviation from the triangle position and normal and parameters $\sigma_c$ and $\sigma_s$. Sharp edges can be preserved. Ref. \cite{lee_fast_2013} performs a similar normal filtering technique but includes an optional intensity term for colored point clouds. A second stage updates vertices using a method proposed by \cite{sun_fast_2007} which executes weighted averaging of neighboring vertices using the newly smoothed normals. Note that smoothing may not be possible if a triangular mesh is so noisy that neighboring triangles have significantly different normals. 

Both Laplacian and bilateral filtering rely upon neighboring triangles for smoothing.  The neighbors are often limited to their 1-ring neighbors defined by vertex or edge neighbors. In Section \ref{sec:bg_mesh_smoothing} we provide accelerated implementations of these algorithms for use with organized point clouds. The organized structure allows an implicit triangular mesh to be defined (i.e., no data structures is needed to store the graph) with the ability to use arbitrary kernel sizes to expand the neighorhood graph, a necessary feature with noisy point clouds. 

\subsection{Dominant Plane Normal Estimation}\label{sec:bg_dominant_plane_normal}

The Gaussian accumulator (spherical histogram) is a widely used method for detecting planes \cite{borrmann_3d_2011}. It discretizes the surface of the unit sphere (S2) into individual cells, creating ``bins'' or ``buckets'' of a histogram. A ``vote'' for a possible plane, often in the form of unit normal and origin offset, are accumulated into this histogram. Peak detection strategies on the histogram can then find dominant plane normals. Many discretization strategies of S2 exist with tradeoffs in speed, memory requirements, and subsequent peak detection.

\begin{figure}[ht]
  \begin{subfigure}[t]{.23\linewidth}
    \centering\includegraphics[width=.95\linewidth]{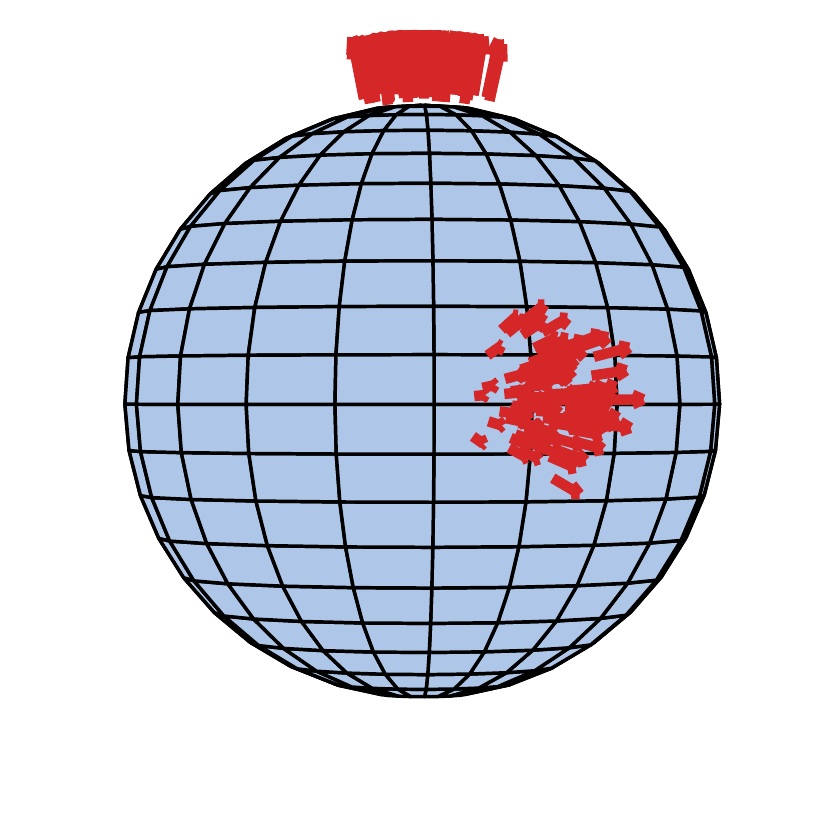}
    \caption{UV Sphere, Equator\label{fig:bg_ga_a}}
  \end{subfigure}
  \begin{subfigure}[t]{.23\linewidth}
    \centering\includegraphics[width=.95\linewidth]{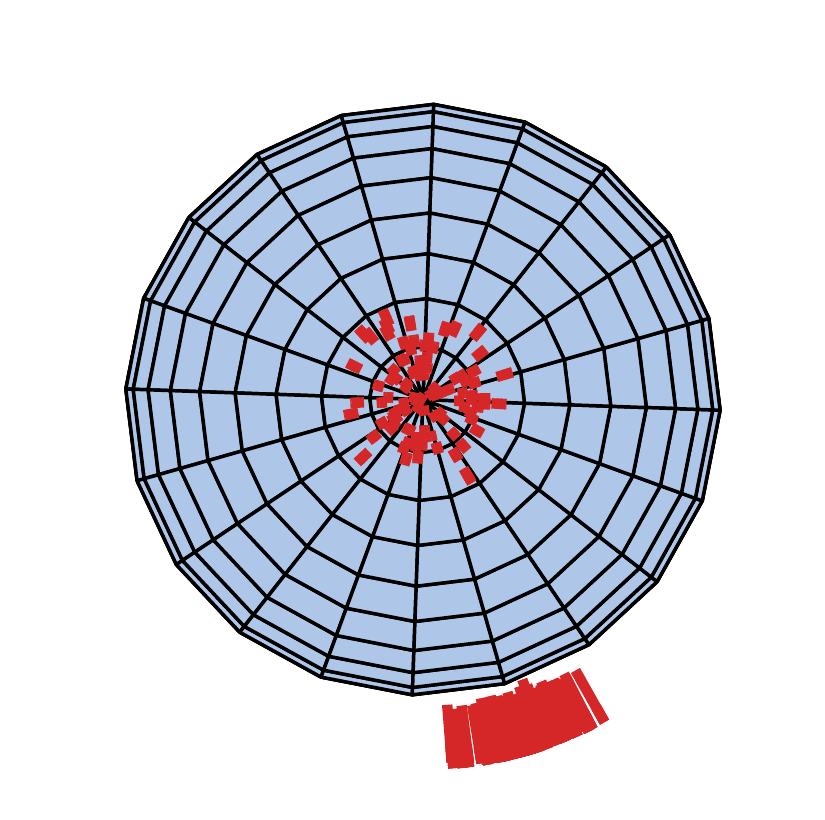}
    \caption{UV Sphere, Pole\label{fig:bg_ga_b}}
  \end{subfigure}
  \begin{subfigure}[t]{.23\linewidth}
    \centering\includegraphics[width=.95\linewidth]{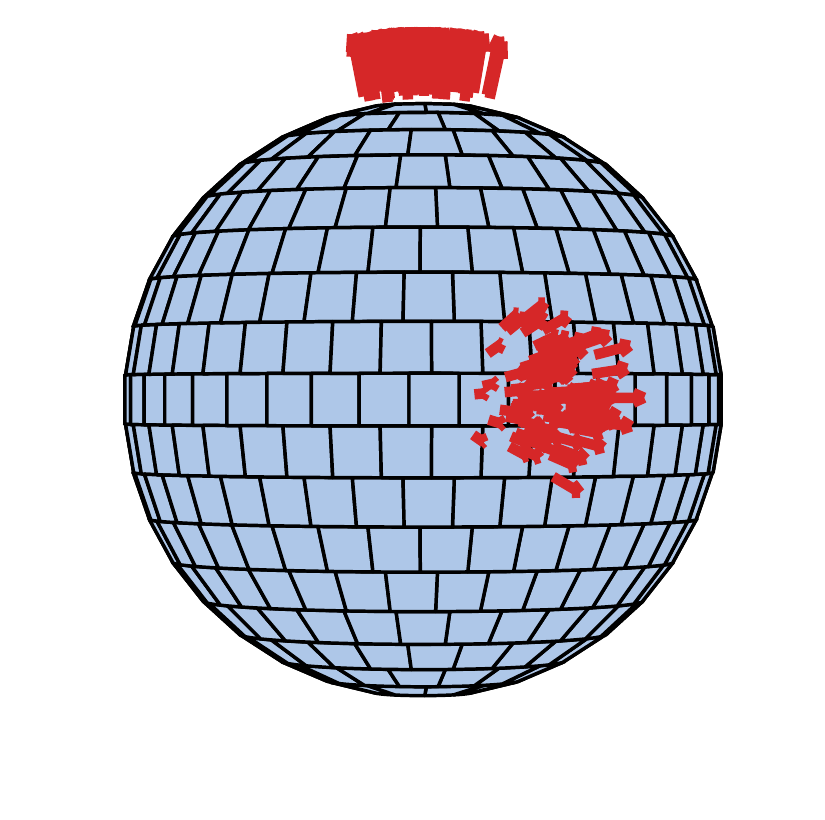}
    \caption{Ball Sphere, Equator\label{fig:bg_ga_c}}
  \end{subfigure}
   \begin{subfigure}[t]{.23\linewidth}
    \centering\includegraphics[width=.95\linewidth]{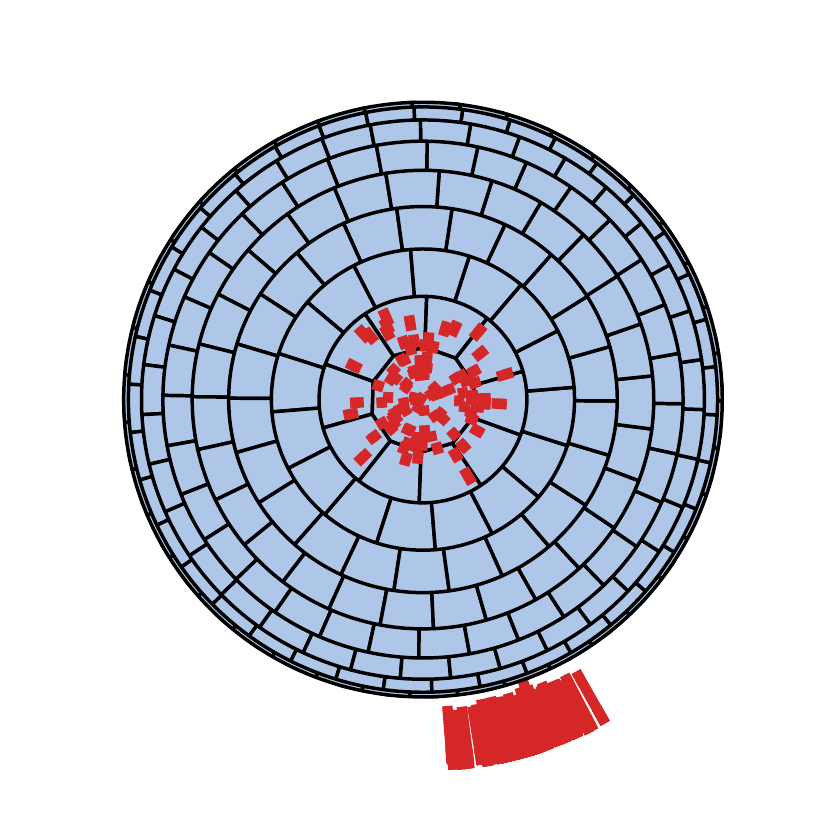}
    \caption{Ball Sphere, Pole\label{fig:bg_ga_d}}
  \end{subfigure}
  \par\bigskip
    \centering
  \begin{subfigure}[t]{.49\linewidth}
    \centering\includegraphics[width=.99\linewidth]{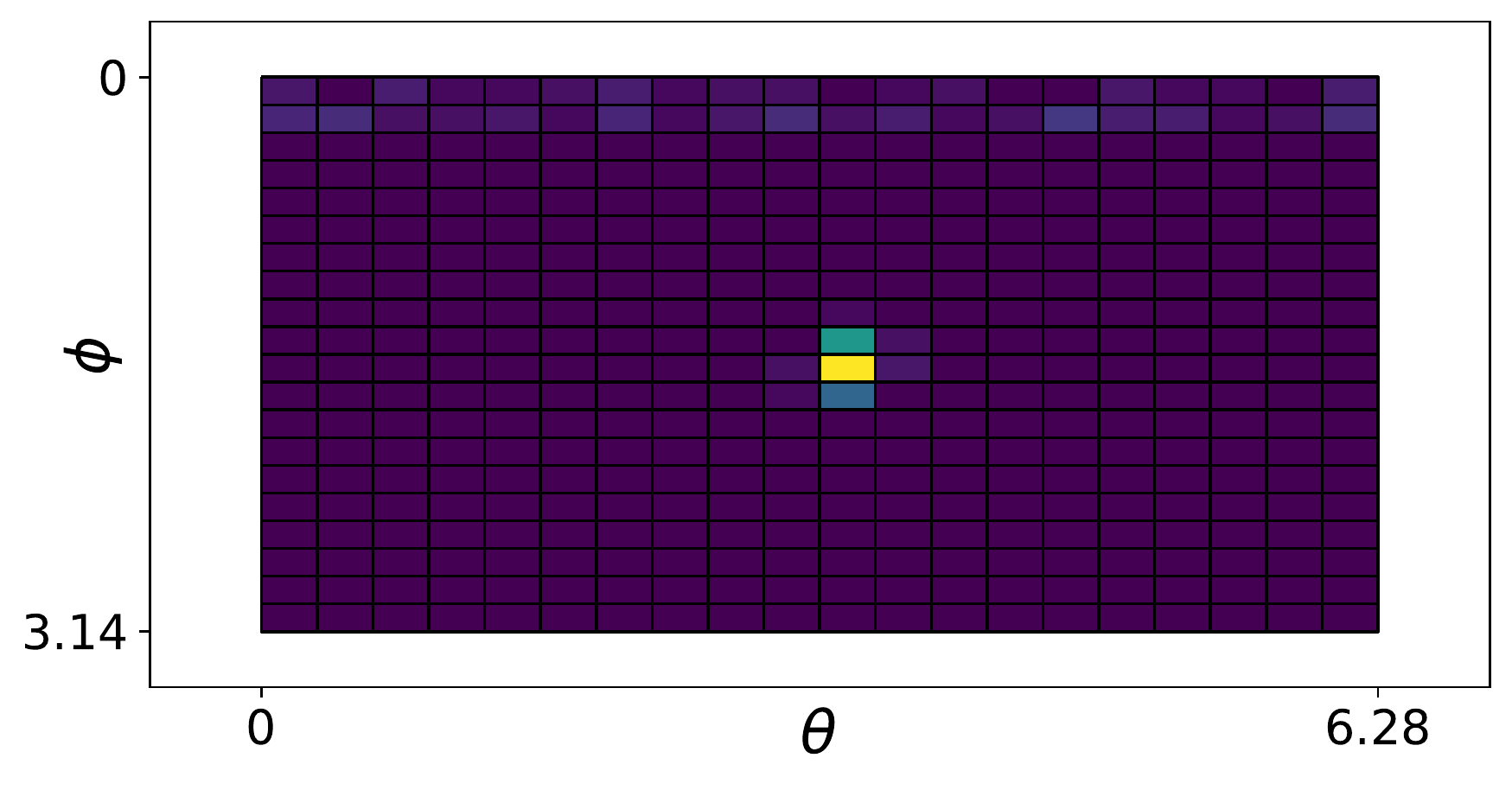}
    \caption{UV Sphere, Histogram\label{fig:bg_ga_e}}
  \end{subfigure}
  \begin{subfigure}[t]{.49\linewidth}
    \centering\includegraphics[trim=0.1cm 0cm 0.99cm 0.6cm,width=.99\linewidth]{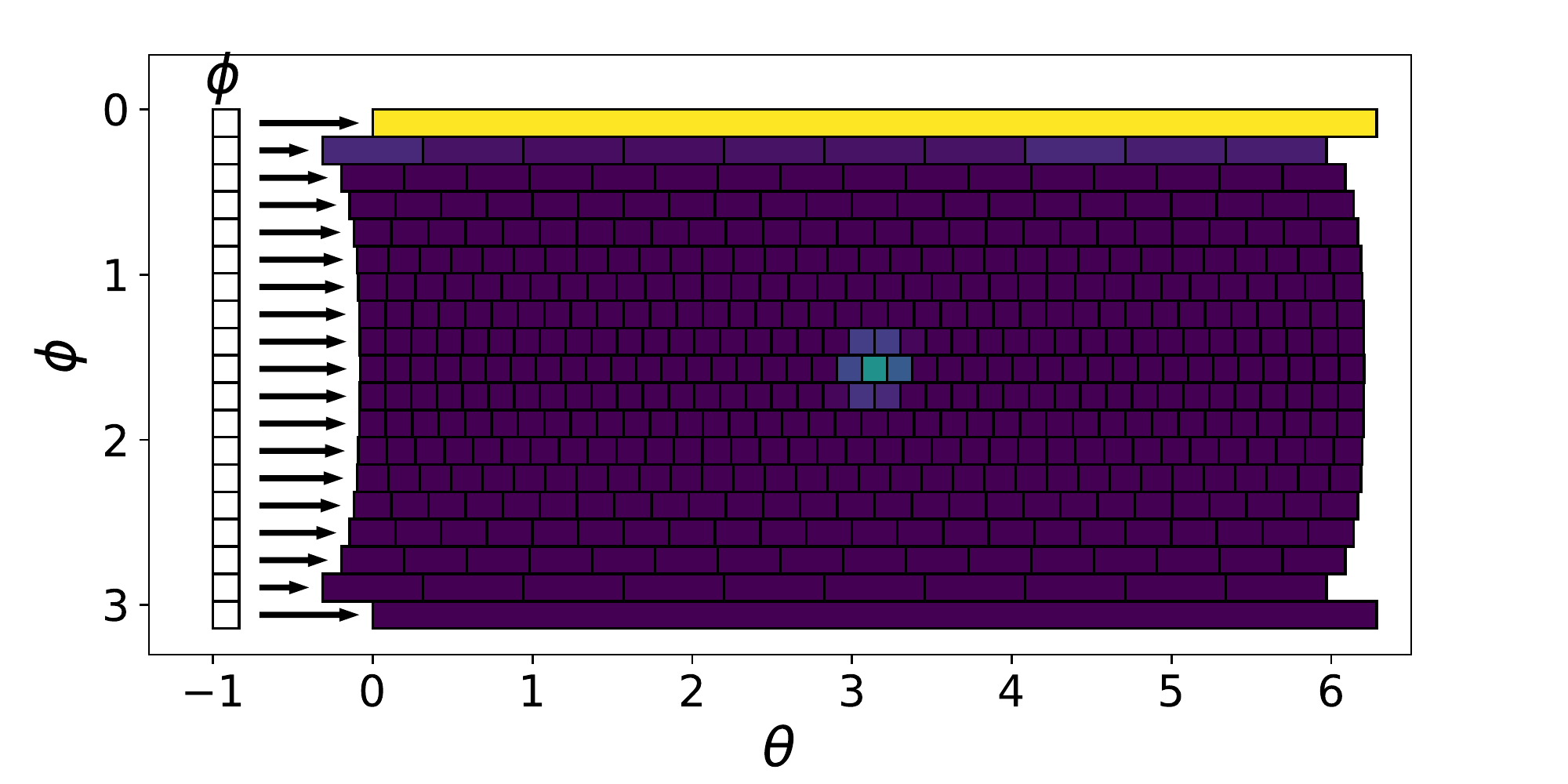}
    \caption{Ball Sphere, Histogram\label{fig:bg_ga_f}}
  \end{subfigure}
  \caption{Examples of integrating normals from two identically ``noisy'' planes (red arrows, Gaussian distributed) into a UV Sphere (\subref{fig:bg_ga_a}, \subref{fig:bg_ga_b}) and Ball Sphere (\subref{fig:bg_ga_c}, \subref{fig:bg_ga_d}). Note the anisotropic property of sphere cells caused by unequal area and shape. (\subref{fig:bg_ga_e}) The UV Sphere histogram is unable to detect the peak at the pole. (\subref{fig:bg_ga_f}) The Ball sphere is able to detect both peaks, but the north pole cell is significantly larger
  leading to an incorrectly higher value than the equator cell.}\label{fig:bg_ga}
\end{figure}

The UV Sphere discretization strategy (Figures \ref{fig:bg_ga_a} and \ref{fig:bg_ga_b}) decomposes S2 into a 2D array by polar coordinates $\theta$ and $\phi$. Each dimension is discretized in equal steps creating a fixed number of cells, $n_{\theta}$ $\times$ $n_{\phi}$. A unit normal to be integrated, $\hat{n}_i \in \mathbb{R}^3$, is converted to polar coordinates $\theta_i \in [0, 360],\phi_i \in [0, 180]$ to identify its corresponding histogram cell in the 2D array. Finding the 2D array index requires a simple operation, e.g.  $\theta_{index} = \theta_i / 360^{\circ} \cdot n_{\theta}$. However UV Sphere cells have different shapes and area with very small cells at the poles resulting in three issues: unequal weighting (voting) during accumulation,  singularities at the poles, and non-equivariant kernels for peak detection. Figure \ref{fig:bg_ga_e} shows an example UV sphere histogram that fails to detect a plane at the top (North) pole.

Refs. \cite{borrmann_3d_2011} and \cite{limberger_real-time_2015} recommend adjusting azimuth step size based upon elevation angle leading to more uniform cell area.  This creates a `Ball'' Sphere with strips with a varying number of cells for each elevation angle stored as a list of lists (Figures \ref{fig:bg_ga_c} and \ref{fig:bg_ga_d}). Cell areas are similar but have different shapes; a substantially larger cap is placed at the poles. Ref. \cite{limberger_real-time_2015} attempts to handle the singularity near the poles after peak detection through a vote weighting scheme. However any discretization strategy by polar coordinates will not have equivariant kernels during peak detection caused by anisotropic cells \cite{cohen_gauge_2019}. 
Ref. \cite{toony_describing_2015} proposes unit sphere tessellation into 1996 equilateral triangle cells. This approach gives near uniform cells in area and shape, resolving previous issues with unequal weighting, pole singularities, and non-equivariant kernels. The process of integrating a unit normal into the histogram is no longer an indexing scheme. They propose to use a $K$-$D$ tree to spatially index each cell using its triangle normal.  A nearest neighbor search must be conducted for every unit normal integrated into the histogram. Peak detection is not performed, instead the sorted histogram distribution is analyzed to predict the shape of the object being integrated (e.g., circle, plane, or torus). 

In Polylidar3D we tessellate the unit sphere with triangles by recursively subdividing the primary faces of an icosahedron. The recursion level dictates the approximation of the unit sphere. Our search strategy does not rely upon $K$-$D$ trees but instead uses a global index from space filling curves followed by local neighborhood search. We unfold the icosahedron into a 2D image in a particular way that guarantees equivariant kernels as outlined in \cite{cohen_gauge_2019}. Standard 2D image peak detection is performed with nearby peaks clustered using AHC.










\section{Preliminaries}\label{sec:prelim}

A 3D point $ \vec{{p}}$ is defined in a Cartesian reference frame by orthogonal bases $\hat{\mathbf{e}}_x$,  $\hat{\mathbf{e}}_y$,  and $\hat{\mathbf{e}}_z$: 
\begin{equation}
\label{eq:point}
    \vec{{p}}=x\,\hat{\mathbf{e}}_x + y\, \hat{\mathbf{e}}_y + z\, \hat{\mathbf{e}}_z= [x,y,z]
\end{equation}
An \emph{unorganized} 3D point cloud is an arbitrarily ordered array of points denoted $\mathcal{P}= \{ \vec{{p}_{0}}, \vec{{p}_{i}}, \ldots, \vec{{p}}_{n-1} \}$ with an index $i \in [0, n-1]$.
An \emph{organized} 3D point cloud is structured with 2D indices $u \in [0, M -1], v \in [1, N -1]$ such that $\vec{p}_{u,v} = [\vec{x}_{u,v}, \vec{y}_{u,v}, \vec{z}_{u,v}]$.   Neighboring 2D indices ($u,v$) and ($u+1, v+1$) represent 3D proximity relationships between $p_{u,v}$ and $p_{u+1,v+1}$ when they lie on the same surface \cite{feng_fast_2014}. These 2D indices create an \emph{image space} with $M$ and $N$ denoting the rows and columns. 
Note that the 2D indices can be collapsed to a 1D stacked array by $i = u \cdot N + v$.
 A triangular mesh $ \mathcal{T}$ with $k$ triangles is defined by
\begin{equation}
\label{eq:tri}
    \mathcal{T} = \{ t_0, t_i, \ldots, t_{k-1} \}
\end{equation}
where each $t_i$ is a triangle with vertices defined by three point indices $\{i_1, i_2, i_3\} \in \left[0,n-1\right]$ referencing points in $\mathcal{P}$. A half-edge triangulation further decomposes each triangle into three individual half-edges. Specifically each edge in the mesh is split into two oriented half-edges, often called twin or opposite edges \cite{paris_modified_2013}. Each half-edge is represented by a unique id $he_j$ in triangle $t_i = \operatorname{floor}(he_j/3)$. An ordered array $\mathcal{HE}$ is created to find corresponding twin edges. Specifically the twin edge of $he_i$ can be found at index $i$ in $\mathcal{HE}$. If no twin exists, i.e., the edge is on a border, then -1 is returned.


The Open Geospatial Consortium (OGC) standard \cite{herring2006opengis} defines a \textit{linear ring} as a consecutive list of points that is both closed and simple. This mandates the linear ring to have non-intersecting line segments that join to form a closed path. A valid \textit{polygon} must have a single exterior linear ring representing the hull of the polygon and a set of linear rings (possibly empty) representing holes inside the polygon. 


\section{Mesh Creation}\label{sec:methods_mesh_creation}

Polylidar3D requires a half-edge triangulated mesh to perform plane and polygon extraction. Mesh generation for unorganized and organized 3D points clouds is described below, followed by details on converting a user-provided triangular mesh to half-edge form.

\subsection{Unorganized 3D Point Clouds}\label{sec:methods_mesh_upc}

\begin{figure}[t]

  \begin{subfigure}{.48\linewidth}
    \centering\includegraphics[width=.95\linewidth]{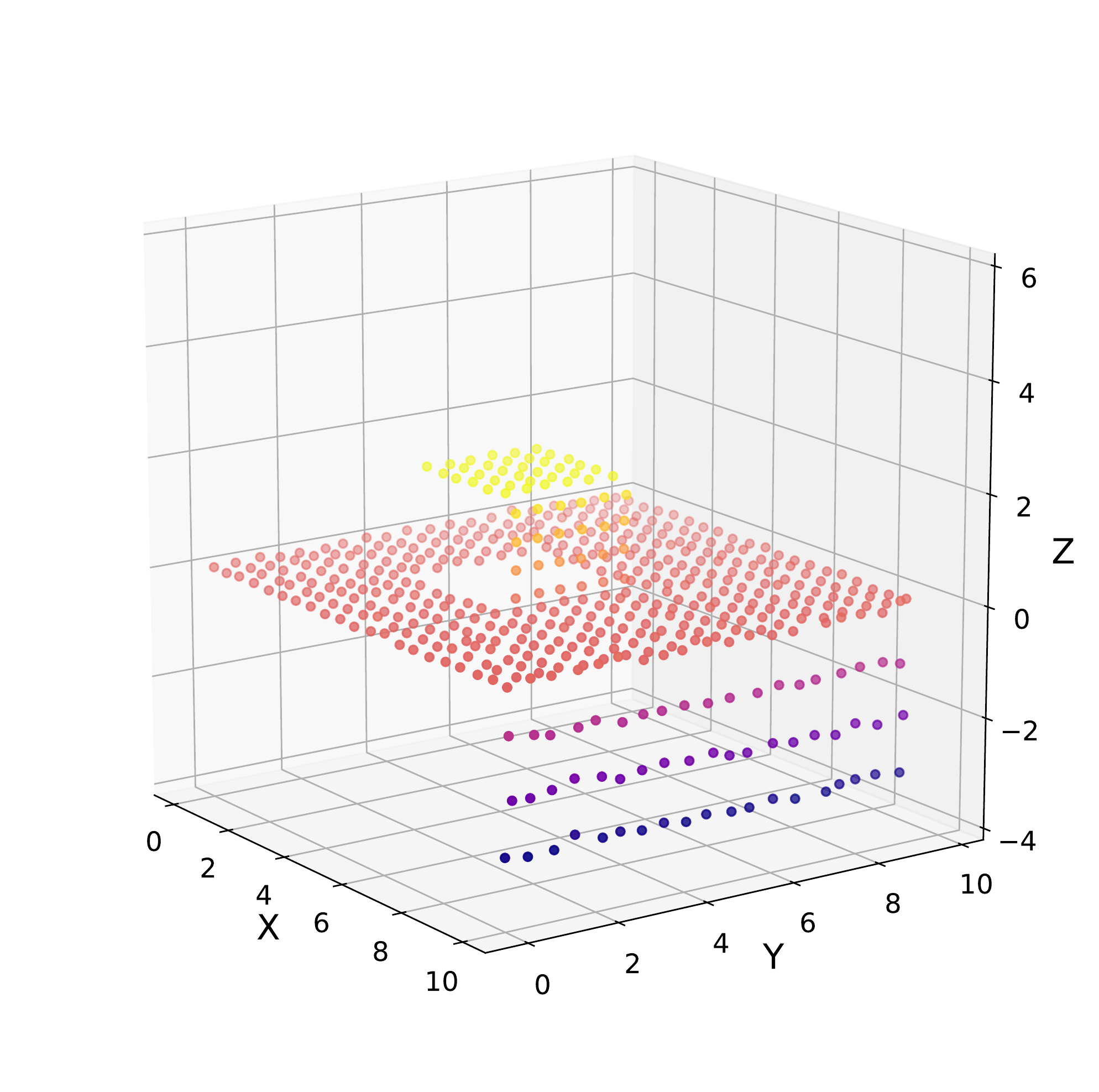}
    \caption{\label{fig:25DPoints}Unorganized 3D Point Cloud}
  \end{subfigure}
  \begin{subfigure}{.48\linewidth}
    \centering\includegraphics[width=.95\linewidth]{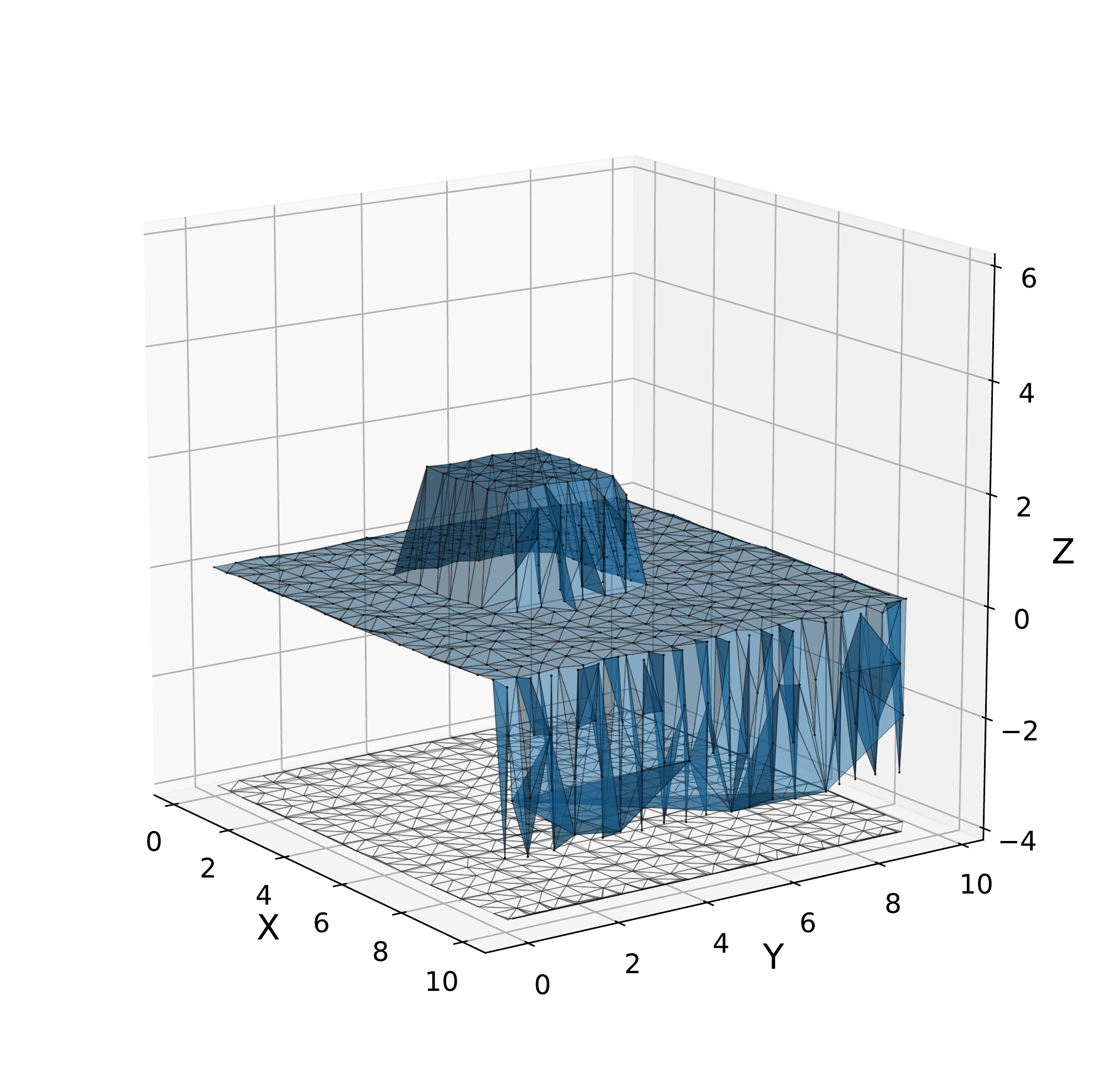}
    \caption{\label{fig:25DMesh}2D and 3D Mesh}
  \end{subfigure}
  \caption{Example conversion of an unorganized 3D point cloud to a 3D triangular mesh. (\subref{fig:25DPoints}) Synthetic point cloud of a  rooftop scene generated from an overhead laser scanner. A single wall is captured because the scanner is slightly angled. (\subref{fig:25DMesh}) The point cloud is projected to the $xy$ plane and triangulated, generating the dual 3D mesh. Only planes aligned with the $xy$ plane can be captured.  }\label{fig:25DMeshCreation}
\end{figure}

We convert an unorganized 3D point cloud $\mathcal{P}$ into a 3D triangular mesh through 2.5D Delaunay triangulation \cite{de2008delaunay}. $\mathcal{P}$ is projected to the $xy$ plane, creating a corresponding 2D point set that is subsequently triangulated.  Half-edge triangulation is provided by the Delaunator library with robust geometric predicates \cite{dealaunator, richard_shewchuk_adaptive_1997}. Although triangulation is performed in 2D, both 2D and 3D point sets have 1:1 correspondence, allowing dual construction of the 3D mesh. Figure \ref{fig:25DMeshCreation} demonstrates this technique applied to a synthetic rooftop scene with noisy point clouds from an overhead sensor. The rooftop is captured at a slight angle providing points of one side of a building wall.  Only planar segments roughly aligned with $xy$ plane can be extracted with this technique, i.e., only the rooftop can be extracted, no walls. This type of conversion is most suitable for 3D points clouds generated from a top down viewpoint, such as airborne LiDAR point clouds as shown in Section \ref{sec:results_rooftop}. In this situation the plane normal to be extracted is already aligned with the $xy$ plane. 

Plane normals may not be aligned with the inertial $xy$ plane, e.g., 3D laser scanner rigidly mounted on an automobile. The point cloud, generated in the sensor frame, must then be rotated such that desired plane to be extracted is aligned with $xy$ plane. This requires a priori knowledge of the plane normal and rigid body transformation necessary to align the sensor frame point cloud as demonstrated in Section \ref{sec:results_kitti} where the ground plane (road) is extracted from point clouds generated by a spinning LiDAR sensor mounted on a car.

\subsection{Organized 3D Point Clouds}\label{sec:methods_mesh_opc}

\begin{figure}[t]
  \begin{subfigure}[t]{.33\linewidth}
    \centering\includegraphics[width=.95\linewidth]{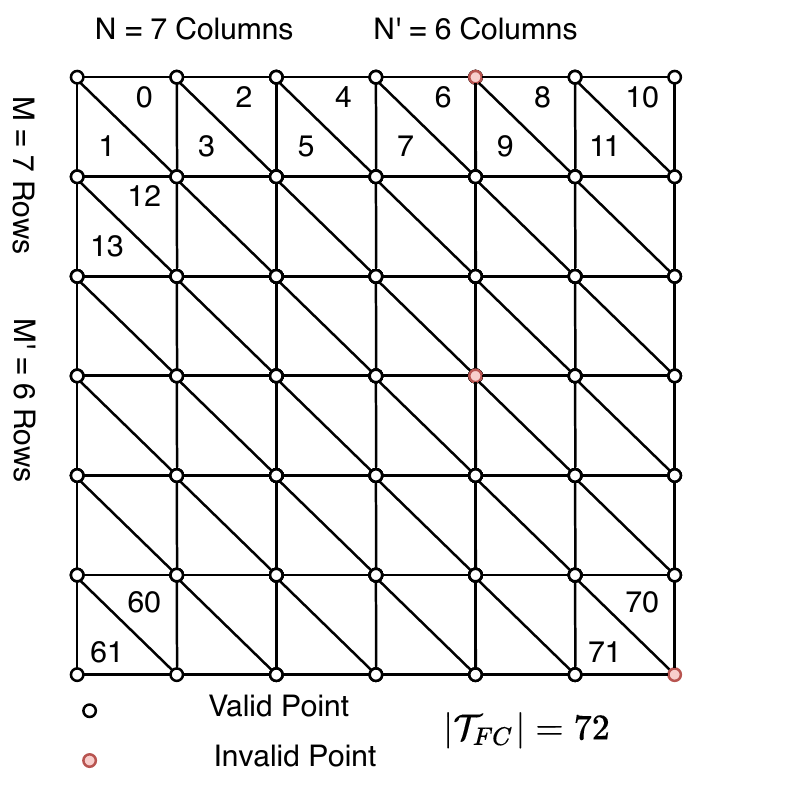}
    \caption{\label{fig:OPC_FC}Implicit Mesh, $\mathcal{T}_{FC}$ }
  \end{subfigure}
  \begin{subfigure}[t]{.33\linewidth}
    \centering\includegraphics[trim=0.1cm 1.7cm 1.0cm 0cm,width=.99\linewidth]{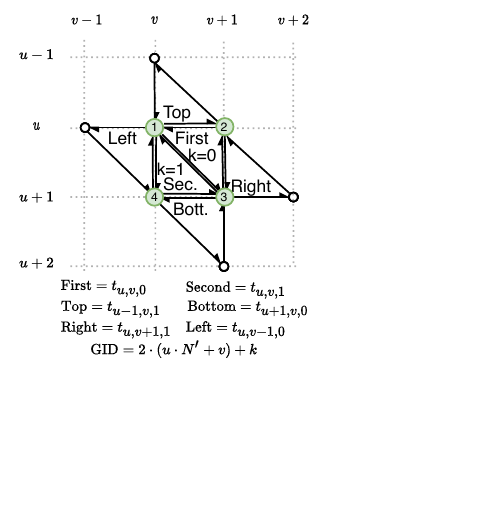}
    \caption{\label{fig:OPC_HE} Triangle indexing scheme}
  \end{subfigure}
  \begin{subfigure}[t]{.33\linewidth}
    \centering\includegraphics[trim=0cm 0.0cm 0cm 0cm,width=.95\linewidth]{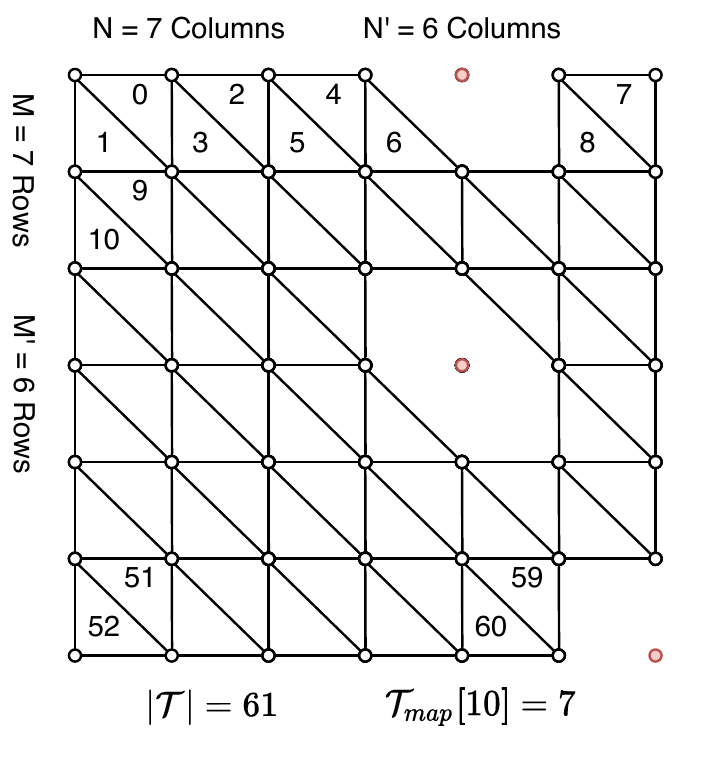}
    \caption{\label{fig:OPC_Mesh} Final Mesh, $\mathcal{T}$ }
  \end{subfigure}
  \caption{Example conversion of a 7X7 Organized Point Cloud (OPC) to a half-edge triangular mesh. Points are represented by circles; red indicates an invalid value (e.g., 0 depth measurement). (\subref{fig:OPC_FC}) Implicit mesh of OPC with right cut triangles, $\mathcal{T}_{FC}$. $GID$s for each triangle are marked.  (\subref{fig:OPC_HE}) Indexing scheme to define $GID$s for triangles in $\mathcal{T}_{FC}$. (\subref{fig:OPC_Mesh}) Final mesh $\mathcal{T}$ with triangles created if and only if all vertices are valid. Unique indices into $\mathcal{T}$  are marked.  $\mathcal{T}_{map}$ maps between $GID$s in $\mathcal{T}_{FC}$ to $\mathcal{T}$. }\label{fig:OPC_Mesh_all}
\end{figure}

Half-edge triangulation of an $M \times N$ organized point cloud can be quickly computed using spatial relationships from the image space. Our procedure is similar to Ref. \cite{lee_fast_2013} except our method creates an explicit half-edge triangulation and only uses right-cut triangles. Our half-edge triangulation allows efficient triangle region growing which is a requirement for real-time polygon extraction.  Ref. \cite{lee_fast_2013} performs adaptive meshing, switching between right and left cut triangles, to better handle missing data at the expense of increased computational demand. Figure \ref{fig:OPC_Mesh_all} demonstrates an example conversion of a 7X7 organized point cloud to a half-edge mesh using our procedure. The procedure creates the triangle set $\mathcal{T}$, half-edge array $ \mathcal{HE}$, and a triangle map $\mathcal{T}_{map}$ as documented below.

First, any invalid data in the point cloud is set to an NaN floating point value. This keeps the point cloud organized and prompts removal of invalid triangles in the mesh. An implicit fully-connected right-cut mesh triangulates all points, including invalid points marked red in Figure \ref{fig:OPC_FC} and denoted $\mathcal{T}_{FC}$.  Each 2X2 grid in the OPC creates two triangles, which we denote first and second as shown in Figure \ref{fig:OPC_HE}. Each triangle in $\mathcal{T}_{FC}$ is indexed by $t_{u,v,k}$ where  $k \in \{0,1\}$ represents the first or second triangle, respectively. A unique global id $GID = 2 \cdot (u \cdot (N-1) + v) + k$ is shown inside each triangle in Figure \ref{fig:OPC_FC}. The final mesh returned $\mathcal{T}$ is shown in Figure \ref{fig:OPC_Mesh} with construction outlined in Algorithm \ref{alg:opc_mesh_creation}. The data structure $\mathcal{T}_{map}$ defines a mapping of global ids in $\mathcal{T}_{FC}$ to their index positions in $\mathcal{T}$ (if they exist, else -1) which is used later in half-edge extraction. Algorithm \ref{alg:opc_mesh_creation} begins by iterating over all 2X2 point grids and constructs the first and second triangle for each. These respective triangles are only added to $\mathcal{T}$ if all three points are valid. Triangle point indices are added counter-clockwise with the three half-edges implicitly defined by the ordered traversal of point indices, e.g., the first triangle's half-edges are [PI3 $\rightarrow$ PI2, PI2 $\rightarrow$ PI1, PI1 $\rightarrow$ PI3].


The half-edge array $\mathcal{HE}$ is constructed 
using the previously calculated $\mathcal{T}_{map}$ and is shown in Algorithm \ref{alg:opc_halfedge_creation}. The algorithm begins at Line 3 by setting all half-edges in $\mathcal{HE}$ to the default sentinel value of -1 indicating no shared edge.  Line 4 and 5 then begin iterating through every 2X2 grid in the OPC inspecting the first and second triangles in $\mathcal{T}_{FC}$. Line 6 and 7 retrieve the index of these triangles in $\mathcal{T}$ using $\mathcal{T}_{map}$ if they exist. If these triangles exist then their neighboring triangles may be assigned in Lines 9 and 18 respectively. For example in Line 13 if the right triangle neighbor exists then its first edge corresponding to a half-edge id of $3 \cdot Right_{idx}$ will be linked to the first half-edge of the first triangle.



\begin{algorithm}[ht]
    \SetKwInOut{Input}{Input}
    \SetKwInOut{Output}{Output}
    \SetStartEndCondition{ }{}{}%
    \SetKwProg{Fn}{def}{\string:}{}
    \SetKwIF{If}{ElseIf}{Else}{if}{:}{elif}{else:}{}%

    \Input{Organized Point Cloud: $\mathcal{P}$ \\ 
           Rows: M, Columns: N }
    \Output{Triangle Set: $\mathcal{T}$ \\
           Triangle Map: $\mathcal{T}_{map}$}
    $N' = N - 1$, 
    $M' = M - 1$ \\
    $\mathcal{T} = \emptyset$ \\
    $SV = -1 $ \tcc*[h]{Sentinel Value indicating invalid triangle} \\
    $\mathcal{T}_{map} = [SV, SV, \ldots, SV]$ \tcc*[h]{$|T_{map}| = 2 \cdot M' \cdot N' $} \\
    $n_{tri}$ = 0 \\
    \For{$u \leftarrow 0$ \KwTo  $M'-1$ \hspace{.1cm} }{
        \For{$v \leftarrow 0$ \KwTo  $N'-1$ \hspace{.1cm} }{
            $First_{GID} = 2 \cdot (u \cdot N' + v)$, 
            $Second_{GID} = 2 \cdot (u \cdot N' + v) + 1$ \\
            p1, p2, p3, p4 = GetPointIndices(u, v)\\
            \tcc{First Triangle}
            \uIf{$\operatorname{NotNan}$(p1, p2, p3, $\mathcal{P}$) }{
                $\mathcal{T}$ = $\mathcal{T}$ + \{ p3, p2, p1 \} \\
                $\mathcal{T}_{map}[First_{GID}] =  n_{tri}$ \\
                $n_{tri} = n_{tri} + 1$ \\
            }
            \tcc{Second Triangle}
            \uIf{$\operatorname{NotNan}$(p1, p3, p4, $\mathcal{P}$) }{
                $\mathcal{T}$ = $\mathcal{T}$ + \{ p1, p4, p3 \} \\
                $\mathcal{T}_{map}[Second_{GID}] =  n_{tri}$ \\
                $n_{tri} = n_{tri} + 1$ \\
            }
        }
    }
    return $\mathcal{T}, \mathcal{T}_{map}$
    \caption{Extract Triangles from OPC}
    \label{alg:opc_mesh_creation}
\end{algorithm}

\begin{algorithm}[ht]
    \SetKwInOut{Input}{Input}
    \SetKwInOut{Output}{Output}
    \SetStartEndCondition{ }{}{}%
    \SetKwProg{Fn}{def}{\string:}{}
    \SetKwIF{If}{ElseIf}{Else}{if}{:}{elif}{else:}{}%

    \Input{Triangle Map: $\mathcal{T}_{map}$ \\ 
          Rows: M, Columns: N }
    \Output{Half-Edge Set: $\mathcal{HE}$}
    $N' = N - 1$, 
    $M' = M - 1$ \\
    $SV = -1$ \tcc*[h]{Sentinel value indicating no shared edge } \\
    $\mathcal{HE} = [SV, SV, \ldots, SV]$ \tcc*[h]{$|\mathcal{HE}| = 3 \cdot |\mathcal{T}| $} \\
    \For{$u \leftarrow 0$ \KwTo  $M'-1$ \hspace{.1cm} }{
        \For{$v \leftarrow 0$ \KwTo  $N'-1$ \hspace{.1cm} }{
            $First_{idx}$ = $\mathcal{T}_{map}[2 \cdot (u \cdot N' + v)] $\\
            $Second_{idx}$ = $\mathcal{T}_{map}[2 \cdot (u \cdot N' + v) + 1] $\\
            $Top_{GID}$, $Right_{GID}$, $Bottom_{GID}$, $Left_{GID}$ = GetNeighborsGID(u,v)\\
            \uIf{$First_{idx}$ != $SV$ }{
                $Top_{idx}$ = $\mathcal{T}_{map}[Top_{GID}] $\\
                $Right_{idx}$ = $\mathcal{T}_{map}[Right_{GID}] $\\

                \uIf{$Right_{idx}$ != $SV$ }{
                    $\mathcal{HE}[First_{idx} \cdot 3]$ = $Right_{idx} \cdot 3$
                }
                \uIf{$Top_{idx}$ != $SV$ }{
                    $\mathcal{HE}[First_{idx} \cdot 3 + 1]$ = $Top_{idx}$ $\cdot 3 + 1$
                }
                \uIf{$Second_{idx}$ != $SV$ }{
                    $\mathcal{HE}[First_{idx} \cdot 3 + 2]$ = $Second_{idx}$ $\cdot 3 + 2$
                }
            }
            \uIf{$Second_{idx}$ != $SV$ }{
                $Bottom_{idx}$ = $\mathcal{T}_{map}[Bottom_{GID}] $\\
                $Left_{idx}$ = $\mathcal{T}_{map}[Left_{GID}] $\\

                \uIf{$Left_{idx}$ != $SV$ }{
                    $\mathcal{HE}[Second_{idx} \cdot 3]$ = $Left_{idx}$ $\cdot 3$
                }
                \uIf{$Bottom_{idx}$ != $SV$ }{
                    $\mathcal{HE}[Second_{idx} \cdot 3 + 1]$ = $Bottom_{idx}$ $\cdot 3 + 1$
                }
                \uIf{$First_{idx}$ != $SV$ }{
                    $\mathcal{HE}[Second_{idx} \cdot 3 + 2]$ = $First_{idx}$ $\cdot 3 + 2$
                }
            }
        }
    }
    return $\mathcal{T}, \mathcal{T}_{map}$
    \caption{Extract Half-Edges from OPC}
    \label{alg:opc_halfedge_creation}
\end{algorithm}

\subsection{User Provided Meshes}\label{sec:methods_mesh_user}

We define a user-provided triangle mesh as a triangle set $\mathcal{T}$ with a corresponding 3D point cloud $\mathcal{P}$. These meshes can be generated from 3D data using a variety of methods \cite{kazhdan_screened_2013,bernardini_ball-pivoting_1999,10.1145/2461912.2461919}. The front-end of Polylidar3D  creates the half-edge set $\mathcal{HE}$ of this mesh to determine shared edges in similar manner to \cite {zhou2018open3d}. This entails first constructing half-edge hashmaps where the key is each half-edge's ordered point indices and value its half-edge ID. Opposite half-edges for any half-edge can then be found by reversing the order of its point indices and performing a hashmap lookup. If a shared half-edge is found then its half-edge ID is mapped into $\mathcal{HE}$.

Certain forms of non-manifold meshes must be explicitly handled. We focus on a subclass of meshes that are not two-manifold. First we define two key properties of a two-manifold mesh: 
\begin{enumerate}
    \item Every vertex connects to a single edge-connected set of triangles.
    \item Every edge is shared by one or two triangles.
\end{enumerate}

\begin{figure}[ht]
  \begin{subfigure}{.24\linewidth}
    \centering\includegraphics[width=.95\linewidth]{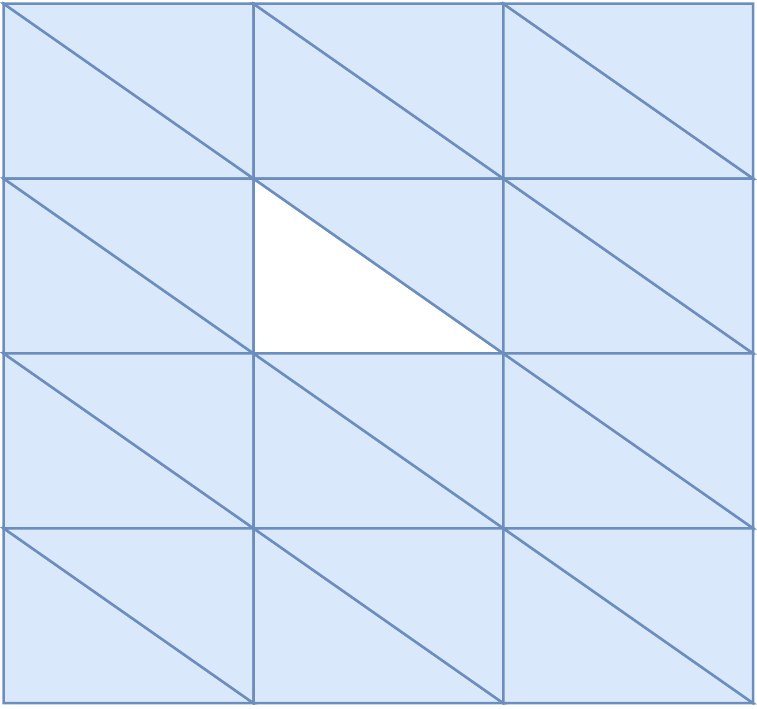}
    \caption{\label{fig:NonMainifoldCond1A}}
  \end{subfigure}
  \begin{subfigure}{.24\linewidth}
    \centering\includegraphics[width=.95\linewidth]{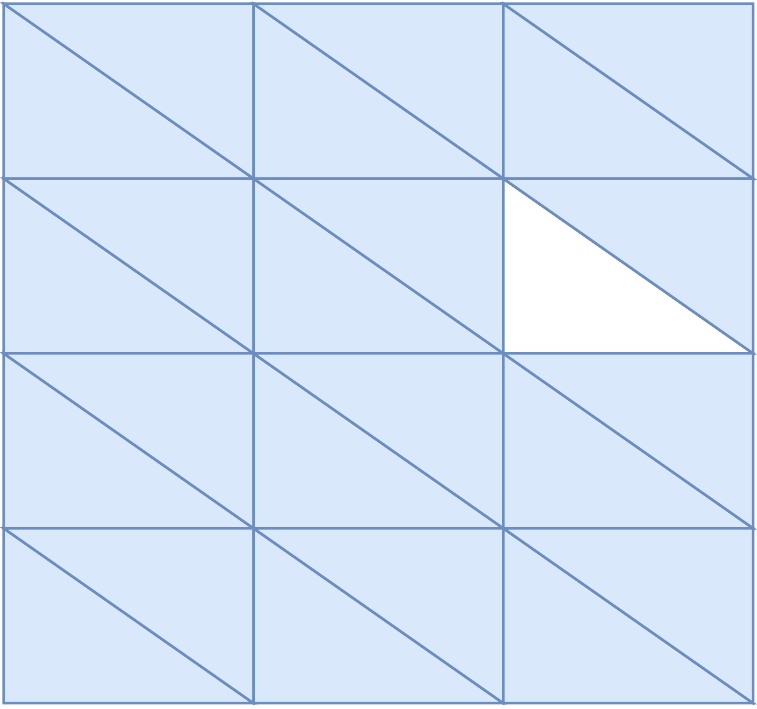}
    \caption{\label{fig:NonMainifoldCond1B}}
  \end{subfigure}
  \begin{subfigure}{.24\linewidth}
    \centering\includegraphics[width=.95\linewidth]{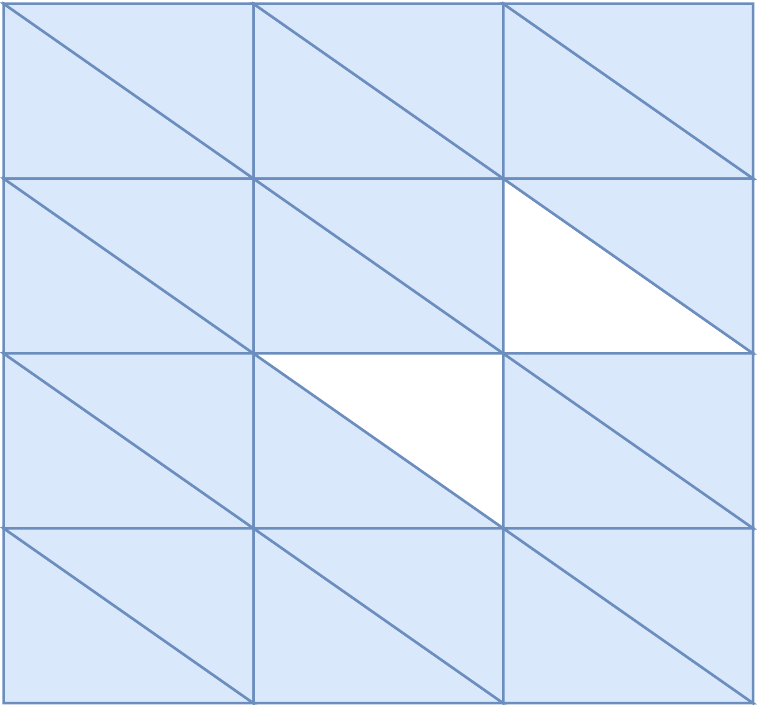}
    \caption{\label{fig:NonMainifoldCond1C}}
  \end{subfigure}
  \begin{subfigure}{.24\linewidth}
    \centering\includegraphics[width=.95\linewidth]{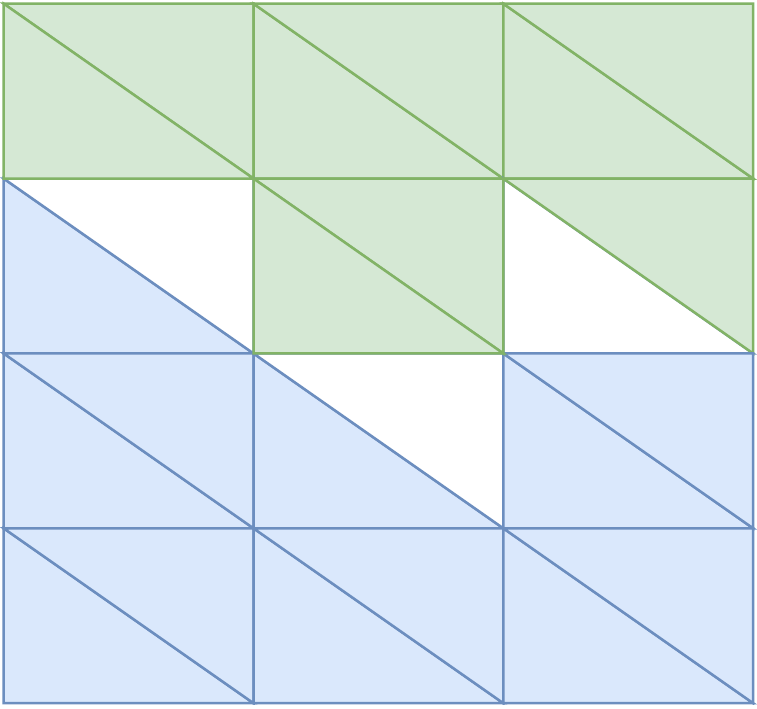}
    \caption{\label{fig:NonMainifoldCond1D}}
  \end{subfigure}
  \caption{Example non-manifold meshes that include holes.}\label{fig:NonMainifoldCond1}
\end{figure}

\begin{figure}[ht]
  \begin{subfigure}{.30\linewidth}
    \centering\includegraphics[clip,trim=1.1cm 5.1cm 7.1cm 5.1cm, width=.95\linewidth]{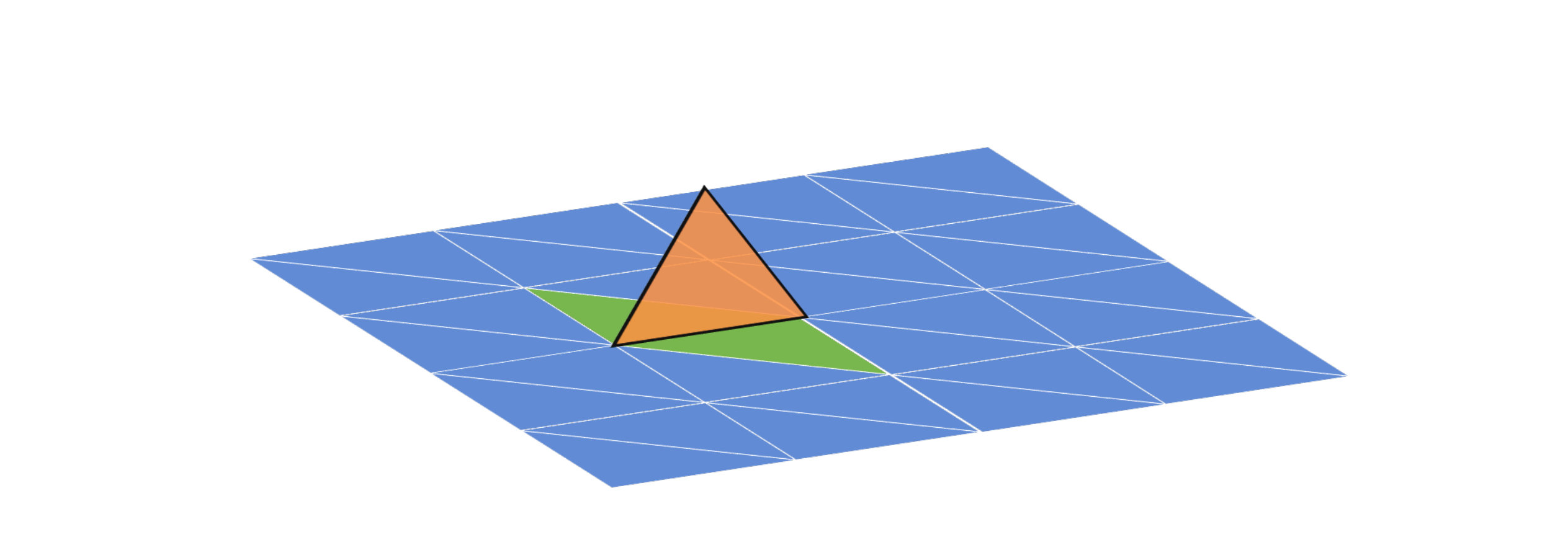}
    \caption{Three shared edges\label{fig:NonMainifoldCond2A}}
  \end{subfigure}
  \begin{subfigure}{.30\linewidth}
    \centering\includegraphics[clip,trim=1.1cm 5.1cm 7.1cm 5.1cm,width=.95\linewidth]{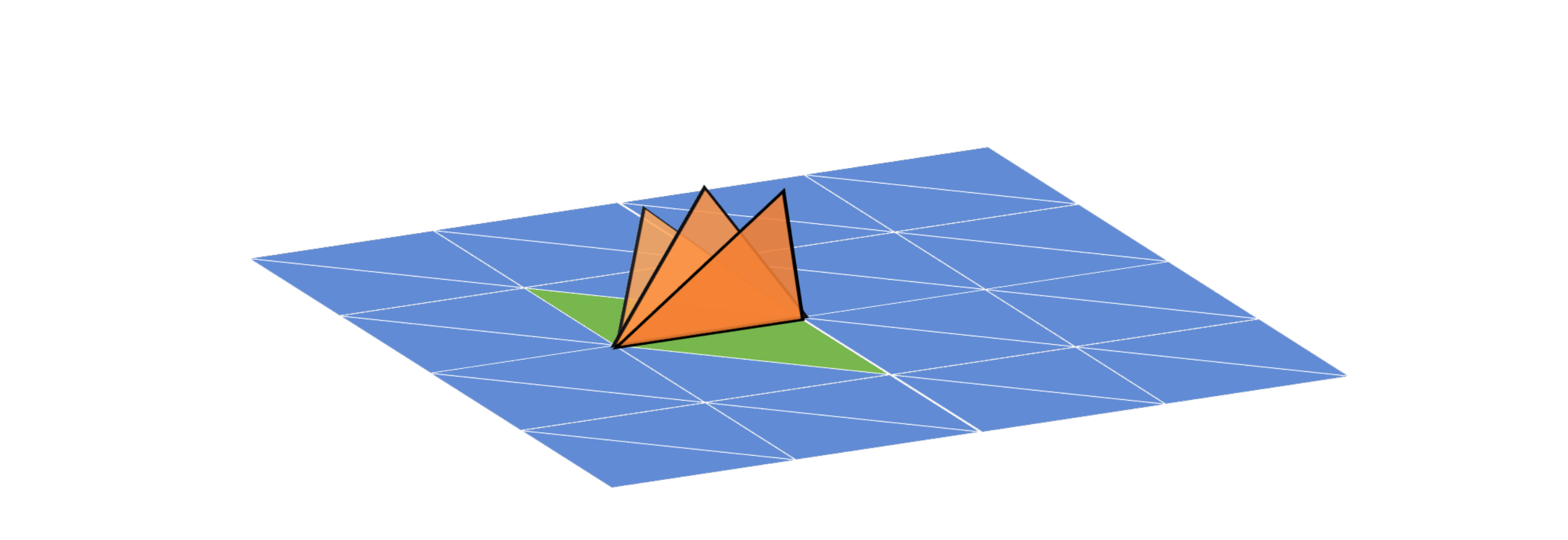}
    \caption{Five shared edges\label{fig:NonMainifoldCond2B}}
  \end{subfigure}
  \begin{subfigure}{.30\linewidth}
    \centering\includegraphics[clip,trim=1.1cm 5.1cm 9.1cm 5.1cm,width=.95\linewidth]{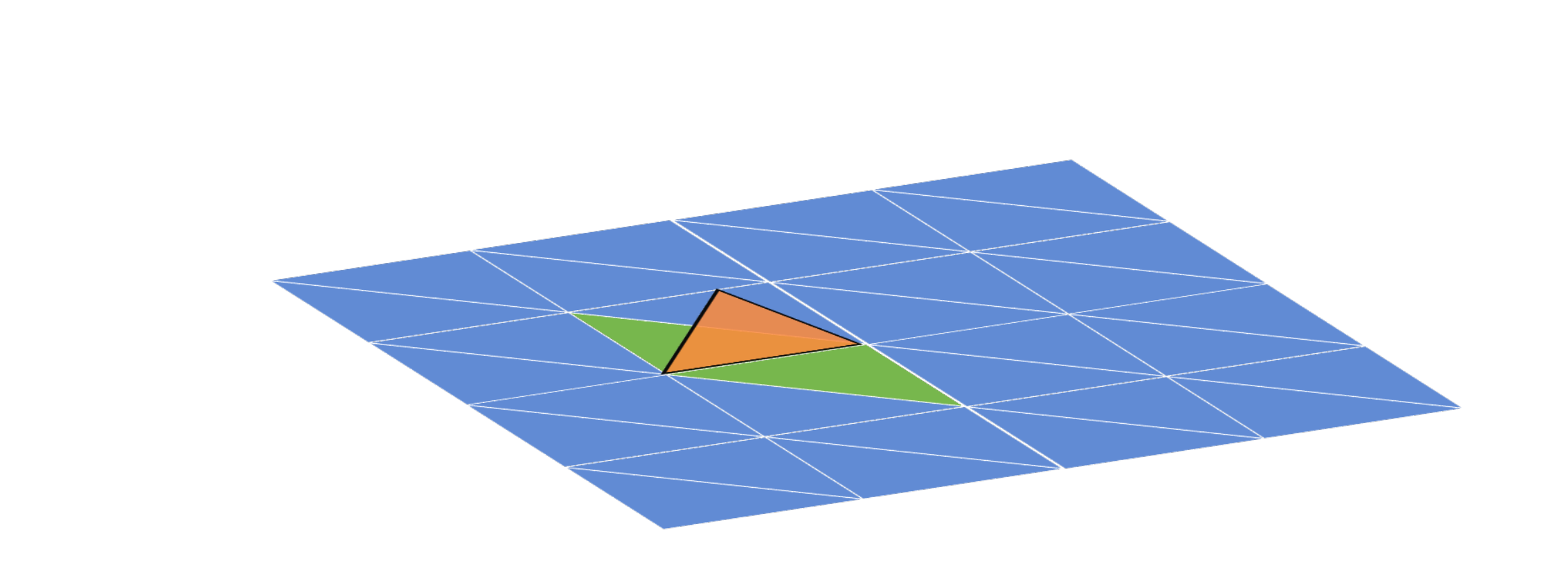}
    \caption{Three shared edges\label{fig:NonMainifoldCond2C}}
  \end{subfigure}
  \caption{Examples of non-manifold meshes where an edge is shared by more than two triangles. This common edge is shared by green and orange triangles. The green triangles form a two-manifold mesh with the blue triangles while the orange triangle(s) do not. The orange triangles in (\subref{fig:NonMainifoldCond2A}) and (\subref{fig:NonMainifoldCond2B}) have sufficiently different normals such the green triangles half-edges can be easily linked. However all triangles in (\subref{fig:NonMainifoldCond2C}) have nearly equal normals making this impossible.}\label{fig:NonMainifoldCond2}
\end{figure}

Figure \ref{fig:NonMainifoldCond1} shows examples of non-manifold meshes where condition (1) is violated. Polylidar3D handles violations of (1)  using methods from our previous work \cite{polylidar2D}. The missing triangles (shown as white) are explicitly captured as holes inside a polygon for (\subref{fig:NonMainifoldCond1A}), (\subref{fig:NonMainifoldCond1B}), and (\subref{fig:NonMainifoldCond1C}), while the mesh is split into two polygons for (\subref{fig:NonMainifoldCond1D}). Figure \ref{fig:NonMainifoldCond2} shows cases of non-manifold meshes that violate condition (2). No mesh generated per Sections \ref{sec:methods_mesh_upc} and \ref{sec:methods_mesh_opc} will violate (2)  because triangulation occurs in 2D space so all edges share at most two triangles. However a user-provided 3D mesh may not satisfy condition (2).

The half-edge array $\mathcal{HE}$ used for neighbor expansion during planar segment extraction in Section \ref{sec:methods_polylidar_plane_extraction} only maps twin half-edges, making condition (2) mesh violations problematic. Three options can handle cases when more than two shared edges exist:

\begin{enumerate}
    \item Store only the first pair of edges found and ignore any others.
    \item Select the pair of edges that are most similar. Similarity between edges is defined by comparing angular distance of their owning triangle normals. 
    \item Ignore all of them by labelling all as boundary edges.
\end{enumerate}

Option one is advantageous in speed and will generally have minimal consequences in the event an incorrect half-edge pairing is chosen, e.g., a green and orange triangle edge are linked in Figure \ref{fig:NonMainifoldCond2}. If green and orange triangle normals are sufficiently different then planar segment extraction will not connect them. However there is no guarantee that this may occur and may fail as in (\subref{fig:NonMainifoldCond2C}). Option two attempts to remove the issue entirely by connecting only the pair of edges that are most similar (edges shared by green triangles). This technique will work for (\subref{fig:NonMainifoldCond2A}) and (\subref{fig:NonMainifoldCond2B}) but will fail once again on (\subref{fig:NonMainifoldCond2C}). Finally option three is the safest, it links none of the shared edges and treats them as border edges (edges sharing no neighbor).  This keeps the critical invariant that no condition (2) violation will exist in an extracted planar mesh. However superfluous border edges will exist which can be handled downstream. Currently only option (1) is implemented in Polylidar3D \cite{polylidarcode} with future plans to allow the user to choose between any of the three proposed solutions. 


\section{Mesh Smoothing}\label{sec:methods_mesh_smoothing}

Mesh smoothing for user-provided triangular meshes is performed using Intel Open3D smoothing procedures \cite{zhou2018open3d}.  The sections below describe our implementation of Laplacian and bilateral filtering for the organized point cloud meshes created in Section \ref{sec:methods_mesh_opc}.  Our implementation is open source and provides single-threaded CPU, multi-threaded CPU, and GPU accelerated routines \cite{organizedpointfilters}. 

\subsection{Laplacian Filter}\label{sec:methods_mesh_smoothing_laplacian}

We implement the standard Laplacian filter for organized point clouds with the benefit that no explicit triangular mesh is required, only the point cloud itself. The filtering, as described in Equation \ref{eq:laplacian_vertex}, results in smoothed vertices of the mesh, i.e, the point cloud is denoised. Vertex neighborhood information is defined implicitly by the image space indices of the organized point cloud.  The neighborhood size is configured by adjusting the kernel size of the filter, e.g, a kernel size of three implies eight vertex neighbors. Filtering this way offers the following benefits:

\begin{enumerate}
    \item Neighboring vertices do not need to be found through lookup over $\mathcal{T}$, $\mathcal{HE}$, or an adjacency list. 
    \item Neighborhood size can be increased by adjusting filter kernel-size. Increasing the kernel size is critical for extremely dense and noisy point clouds.
    \item Parallelization is trivial, similar to image filters, with all necessary neighborhood data for a vertex located close in memory.
\end{enumerate}

The amount of filtering is controlled by $\lambda$, the kernel size, and the number of iterations. As kernel size and number of iterations increase the computational demand of the filter also increases. Mesh borders in image space have no defined neighbors on the exterior thus are not filtered. This gives a negative drawback of a noisy border but a positive benefit of reducing the mesh shrinkage inherit to Laplacian filtering. One may think of the fixed border as ``pinning'' the mesh to prevent overshrinkage.

\subsection{Bilateral Filter}\label{sec:methods_mesh_smoothing_bilateral}

We implement the bilateral mesh filtering algorithm presented in \cite{zheng_bilateral_2011} but for organized point clouds. Smoothing occurs on the implicit fully-connected organized mesh $\mathcal{T}_{FC}$, described in Section \ref{sec:methods_mesh_opc}.  Recall the mesh spatial structure is defined through image indices ($u,v$) with a final index $k \in \{0,1\}$ representing the first or second triangle in a 2X2 quad (see Figure \ref{fig:OPC_HE}). Bilateral filtering per Equation \ref{eq:bilateral} requires data structures for each triangle's centroid and normal, which we denote as $\mathcal{C}$ and $\mathcal{N}$. These are constructed in parallel (if multi-core CPU is available) and laid out in contiguous memory with the same indexing scheme as $\mathcal{T}_{FC}$, i.e., the centroid of triangle $t_{u,v,k}$ is $c_{u,v,k}$. If any of the vertices of a triangle in $\mathcal{T}_{FC}$  are NaN then the associated centroid and normal will also be NaN.

\begin{figure}[ht]
    \centering
    \includegraphics[width=.60\linewidth]{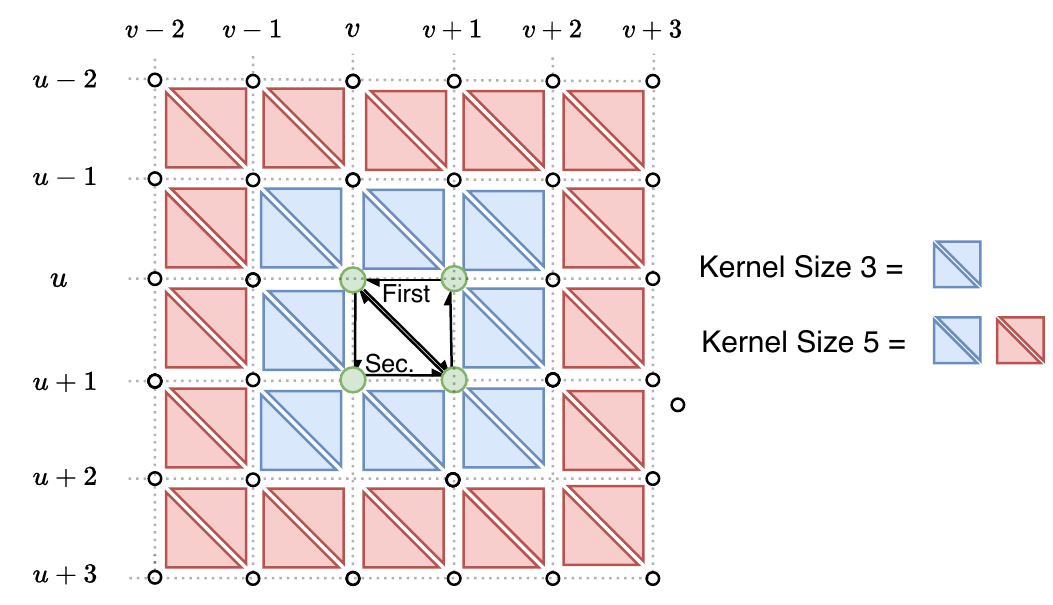}
    \caption{Visualization of a triangle's neighborhood during bilateral filtering of an organized point cloud.  Each 2 $\times$ 2 point group forms two triangles creating a mesh. Each triangle's neighbors are defined by the kernel size. For example a kernel size of five includes all blue and red triangles.}
    \label{fig:smooth_bilateral}
\end{figure}

The algorithm partitions triangle smoothing in image space coordinates, smoothing both the first and second triangles as one unit of work. Each triangle's normal is updated using centroid and normal information from neighboring triangles. The neighbors of a triangle are determined by a user-configurable kernel size as shown in Figure \ref{fig:smooth_bilateral}.  Note that defining neighbors in this way mixes both $n$-ring and $(n+1)$-ring triangle neighbors. However the exponential decay of the bilateral filter in Equations \ref{eq:bilateral_centroid} and \ref{eq:bilateral_normal} ensures that only triangles of similar properties (close in position and orientation) will be integrated into the smoothed normal. Neighbors are not integrated if they have NaN values for their centroid/normal. The end result is smoothed normals for $\mathcal{T}_{FC}$; however, what is actually desired are smoothed normals for $\mathcal{T}$ per Figure \ref{fig:OPC_Mesh}. This is quickly achieved by using $\mathcal{T}_{map}$ to identify the valid normals for $\mathcal{T}$. Ref. \cite{zheng_bilateral_2011} follows up with vertex updating, but we do not perform this step. Vertex updating is an expensive operation which provides minimal benefit for triangle region growing downstream. Only the smoothed normals are needed in Polylidar3D. 

The advantage of implementing bilateral smoothing in this manner is that all information is laid out in contiguous memory to reduce the need for branching code. These are important characteristic for CPU and especially GPU performance. However nontrivial excess work is performed if most of the point cloud is invalid, e.g, invalid depth measurements in a range image. The entire procedure is controlled by $\sigma_c^2$, $\sigma_s^2$, kernel size, and number of iterations.

\section{Dominant Plane Normal Estimation}\label{sec:methods_fastga}

We present a new method for constructing and using a Gaussian Accumulator to identify dominant plane normals in a scene. We call this method the Fast Gaussian Accumulator (FastGA). The input to this method is a list of $k$ unit normals $\mathcal{N} = \{\hat{n}_0, \ldots, \hat{n}_{k-1} \}$ which have been sampled from a scene. Use of denoised data is advantageous but not required. Sections \ref{sec:methods_fastga_histogram} and \ref{sec:methods_fastga_peak} discuss constructing the Gaussian Accumulator and performing peak detection, respectively. 

\subsection{Gaussian Accumulator}\label{sec:methods_fastga_histogram}

The following subsections describe the process to approximate a sphere using an icosahedron, construct the Gaussian Accumulator, and our method to integrate information into the accumulator. 

\subsubsection{Refined Icosahedron}\label{sec:methods_fastga_ico}

\begin{figure}[ht]
    \centering
    \includegraphics[width=.60\linewidth]{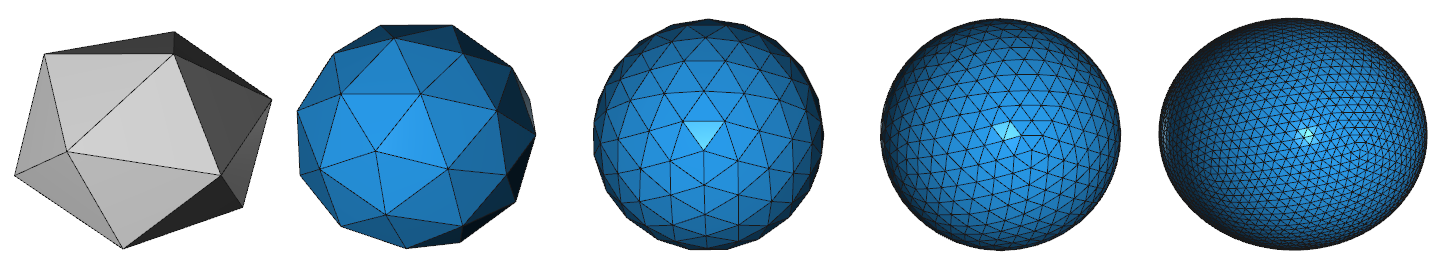}
    \caption{Approximation of the unit sphere with an icosahedron. The level 0 icosahedron is shown on the  left with increasing refinements to the right.  Triangle cells become buckets of a histogram on S2. }
    \label{fig:fastga_refinement}
\end{figure}

A geodesic polyhedron is first constructed by using an icosahedron as the base model approximation of a unit sphere. The icosahedron is composed of 12 vertices and 20 faces and can be seen on the far left in Figure \ref{fig:fastga_refinement}. This polyhedron is refined by recursively dividing each face into four equilateral triangles and then projecting the new vertices onto the surface of a sphere. The number of iterations or levels of recursion is user configurable with higher levels better approximating a sphere. The Class I geodesic polyhedron is defined with the Schläfli symbol $\{3,5+\}_{1,0}$ with frequency doubling at each level \cite{wenninger1999spherical}. Figure \ref{fig:fastga_refinement} shows refinement up to level four while Table \ref{table:fastga_refinement} displays the change in number of vertices, triangles, and approximate angular separation between each triangle. We denote each triangle as the cell or bucket of the histogram of S2. The number of cells, $n$, and their properties described below are  fixed once a refinement level is chosen.

\begin{table}[ht]
\centering
\caption{Levels of Refinement for an Icosahedron.}\label{table:fastga_refinement}
\begin{tabular}{@{}cccc@{}}
\toprule
Level & \# Vertices & \# Triangles & Separation \\ \midrule
0     & 12          & 20           & $41.8^{\circ}$          \\
1     & 42          & 80           & $18.0^{\circ}$            \\
2     & 162         & 320          & $6.9^{\circ}$              \\
3     & 642         & 1280         & $3.1^{\circ}$              \\
4     & 2562        & 5120         & $1.5^{\circ}$              \\ \bottomrule
\end{tabular}
\end{table}

\subsubsection{Gaussian Accumulator Properties}\label{sec:methods_fastga_ga}

\begin{figure}[ht]
    \centering
    \includegraphics[width=.20\linewidth]{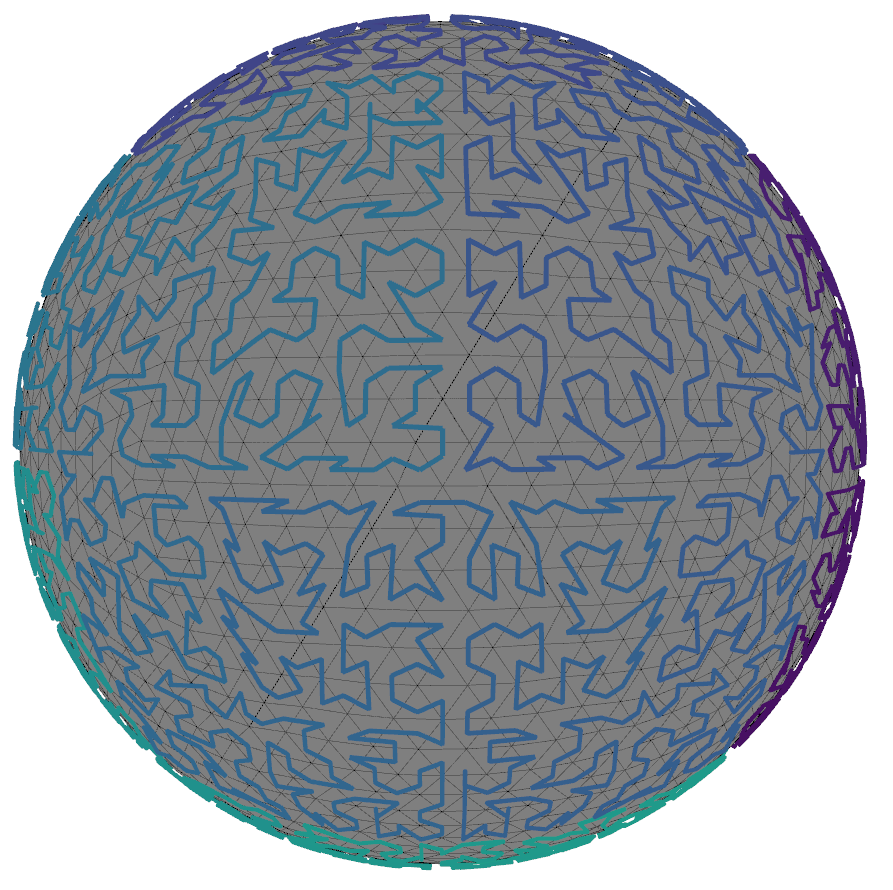}
    \caption{Space filling curve (SFC) of a level 4 icosahedron generated using the S2 Geometry Library. Each cell's surface normal is mapped to an integer creating a linear ordering for a curve. The curve is colored according to this mapping and traverses each cell.   }
    \label{fig:fastga_hilbert}
\end{figure}

A space-filling curve (SFC) maps a multi‐dimensional space into a one‐dimensional space, e.g., $\mathcal{R}^2 \rightarrow \mathcal{R}$.  Hilbert curves are a widely used SFC because they preserve locality well during transformation \cite{mokbel_space-filling_2008}. This means that points close in 1-D space are close in $N$-D, though the converse is not guaranteed to be true. In practice a SFC is approximated using discrete integers. The S2 Geometry library \cite{s2geometry} provides a SFC routine that transforms any real-valued unit normal $\hat{n}_i \in \mathcal{R}^3$ to a 64 bit unsigned integer. The method works by projecting the unit sphere to a cube, creating 2D $\rightarrow$ 1D Hilbert curves for each of the six faces, and finally stitching them together to make one unbroken linear chain.  Each cell in the refined icosahedron has a surface normal $\hat{n}^{c}_{i}$ that can be mapped to a unique ID denoted $s2id$ using this procedure. This generates a one-dimensional thread that passes through every cell such that each cell is visited exactly once as seen in Figure \ref{fig:fastga_hilbert}.

The final Gaussian Accumulator (GA) is then an ordered array of $Cells = [ c_i, \ldots, c_{n-1}] $. Each cell contains its surface normal $\hat{n}^{c}_{i}$, unique $s2id_i$, and an accumulating integer $count_i$. The cell array is sorted by $s2id$, creating the invariant that cells close together in the array are close in physical space. A neighborhood data structure $Nbrs_{i,j}$ is constructed as an N$\times$12 matrix in which the $i^{th}$ row contains the 12 neighboring cell indices of the $i^{th}$ cell in the $Cells$ array. Neighboring triangles are defined as those in the 1-ring vertex adjacency. A maximum of 60 triangles at any level of refinement have only 11 neighbors; all others have 12. For these cells the $12^{th}$ neighbor index is given a sentinel value of -1 to indicate no neighbor is present. 

\subsubsection{Integrating the Gaussian Accumulator with Search}\label{sec:methods_fastga_search}
Integrating a list of $k$ unit normals $\mathcal{N} = \{\hat{n}_0, \ldots, \hat{n}_{k-1} \}$ into the Gaussian Accumulator is done through a search that finds the corresponding cell whose surface normal is closest to an input normal $\hat{n}_i$ then incrementing the cell's $count_i$ member.
Instead of a $K$-$D$ tree search we propose combining a sorted integer search with a local neighborhood search. Though similar, there are nontrivial differences and optimizations that make our method faster.  The main components of the search are as follows: map $\hat{n}_i$ to an integer $s2id$, perform sorted integer interpolation search to reduce search bounds, perform branchless binary search within these bounds in the $Cells$ array, then  perform local neighborhood search to find the correct cell. Algorithm \ref{alg:find_cell_index} outlines this search routine and is explained below.

There exist several methods for sorted integer search such as interpolation and binary search. Interpolation search works by predicting the index of a value in sorted array by interpolating  between the first and last value of the sorted array (thus computing a slope). The process continues iteratively, each time reducing the search window and recomputing a new line for improved prediction. Interpolation search is best used for linear data but still often underperforms in comparison to binary search in practice due to its use of repeated computationally expensive calculations of slope \cite{van_sandt_efficiently_2019}.


\begin{figure}[ht]
        \centering
  \begin{subfigure}[t]{.30\linewidth}
    \centering\includegraphics[clip,trim=0.9cm 0cm 0.5cm 0cm,width=\linewidth]{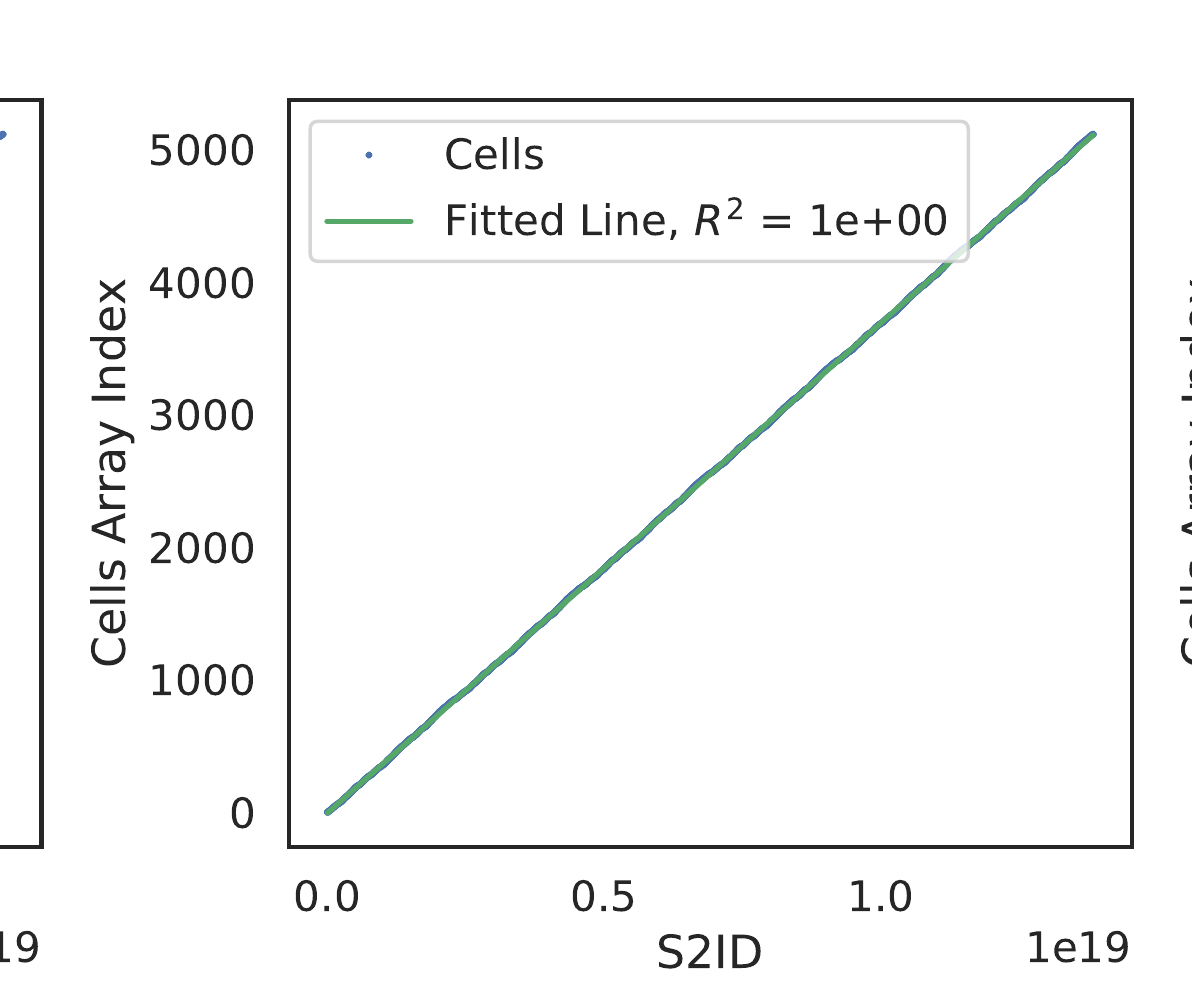}
    \caption{\label{fig:linear_interp_1}}
  \end{subfigure}
\quad \quad
  \begin{subfigure}[t]{.33\linewidth}
    \centering
    \includegraphics[width=\linewidth]{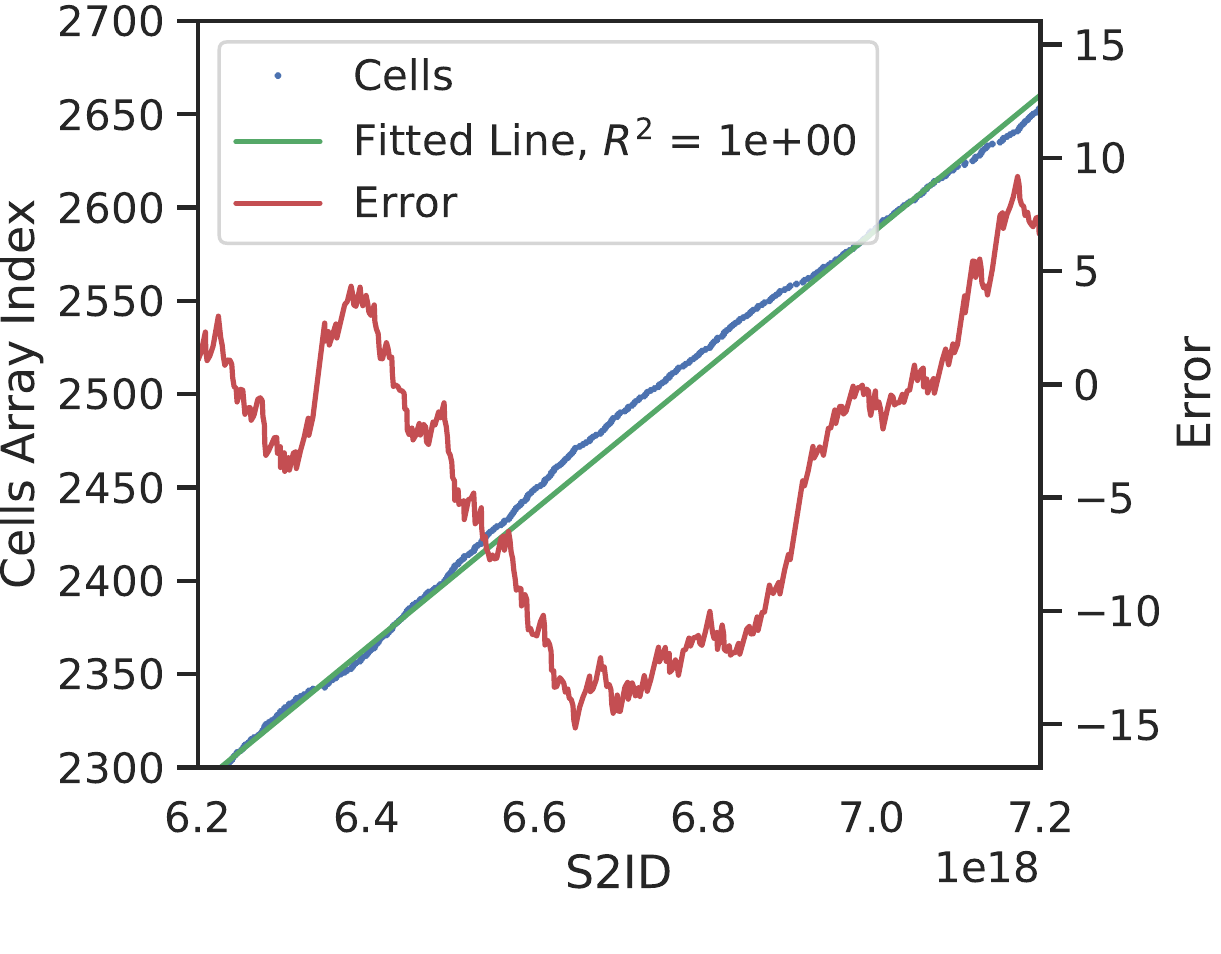}
    \caption{\label{fig:linear_interp_2}}
  \end{subfigure}
  \caption{Linear prediction model for a level four Gaussian Accumulator (GA). The 5120 GA cells are sorted in an array ($Cells$) by their corresponding spatial index $s2id$.  (\subref{fig:linear_interp_1}) Plot relating cell $s2id$ and index position in the $Cells$ array with a regressed line (green) to the data. (\subref{fig:linear_interp_2})  Zoomed-in view showing model error (red line) indicating  difference between predicted and actual index for each $s2id$.  
  }\label{fig:linear_interp}
\end{figure}

\begin{algorithm}[ht]
    \SetKwInOut{Input}{Input}
    \SetKwInOut{Output}{Output}
    \SetStartEndCondition{ }{}{}%
    \SetKwProg{Fn}{def}{\string:}{}
    \SetKw{KwIs}{is}{}
    \SetKwIF{If}{ElseIf}{Else}{if}{:}{elif}{else:}{}%

    \Input{Unit Normal: $\hat{n}_i \in \mathcal{R}^3$ \\
           Cells Array: $Cells$ \\
           Neighbor Matrix: $Nbrs$
           
        }
    \Output{Cell Index: $k_{best} \in [0, |Cells|]$}
    
    $s2id = \operatorname{GetS2ID}(\hat{n}_i)$ \\
    $[k_{min}, k_{max}] = \operatorname{SearchWindow}(s2id) $\\
    \tcc{get closest neighbor by s2id}
    $k' = \operatorname{BranchlessBinarySearch}(s2id, Cells, k_{min}, k_{max})$\\
    $k_{best} = k'$\\
    $dist_{best} = ||\hat{n}^c_{k'} - \hat{n}_i  ||$ \\
    \tcc{local neighbor search by actual distance}
    \For{$j \leftarrow 0$ \KwTo  $12$ \hspace{.1cm} }{
       $k_{nbr} = Nbrs_{k',j}$ \\
        \uIf{$k_{nbr}$ \KwIs $-1$ }{
            continue
        }
       $dist = || \hat{n}^c_{k_{nbr}} - \hat{n}_i ||$ \\
        \uIf{$dist < dist_{best}$ }{
            $k_{best} = k_{nbr}$\\
            $dist_{best} = dist$
        }
    }
    return $k_{best}$
    \caption{Find Cell Index}
    \label{alg:find_cell_index}
\end{algorithm}

Figure \ref{fig:linear_interp_1} shows a graph of cells in the Gaussian Accumulator where the $x$-axis is the $s2id$ and the $y$-axis is the corresponding index into the sorted $Cells$ array. 
We use least squares regression to fit a line to the data shown in Figure \ref{fig:linear_interp_1}, in contrast to only using the first and last values typical of interpolation search. Figure \ref{fig:linear_interp_1} shows this regressed line (green) accurately fits the data overall. In a zoomed plot (Figure \ref{fig:linear_interp_2}), model error, the difference between actual and predicted cell array index, is shown as the red line with values on the right vertical axis. Since the model/data domain and range are ordered and finite we can compute the negative and positive error bounds which is fixed once GA refinement level is chosen. For example, refinement level four with 5120 cells has maximal error bounds of -16 and +16 from any predicted position. This technique brings the  benefits of linear interpolation search without excess computational overhead.

This predicted index and maximal bounds are used to create a binary search window in the $Cells$ array, shown at Line 2 in Algorithm \ref{alg:find_cell_index}. A branchless binary search is used which is faster than standard binary search for arrays of small sizes that fit into CPU L1/L2 caches \cite{khuong_array_2017}. All the search windows at realistic levels of refinement are sufficiently small to meet this criterion. The output of branchless binary search is an index $k'$ into $Cells$ with $s2id$ closest to the mapped $s2id$ of $\hat{n}_i$ (Line 3). There is no guarantee this cell's surface normal $\hat{n}^c_{k'}$ is closest to $\hat{n}_i$ than neighboring cells though it is guaranteed to be close. Therefore a local neighbor search is performed where all 12 neighboring cells' surface normals are compared to $\hat{n}_i$. The cell index with closest surface normal is then returned.

\subsection{Peak Detection}\label{sec:methods_fastga_peak}

\begin{figure}[ht]
 \captionsetup[subfigure]{justification=centering}
  \begin{subfigure}[t]{.40\linewidth}
    \centering\includegraphics[clip,trim=0cm 0cm 0cm 0cm, width=.80\linewidth]{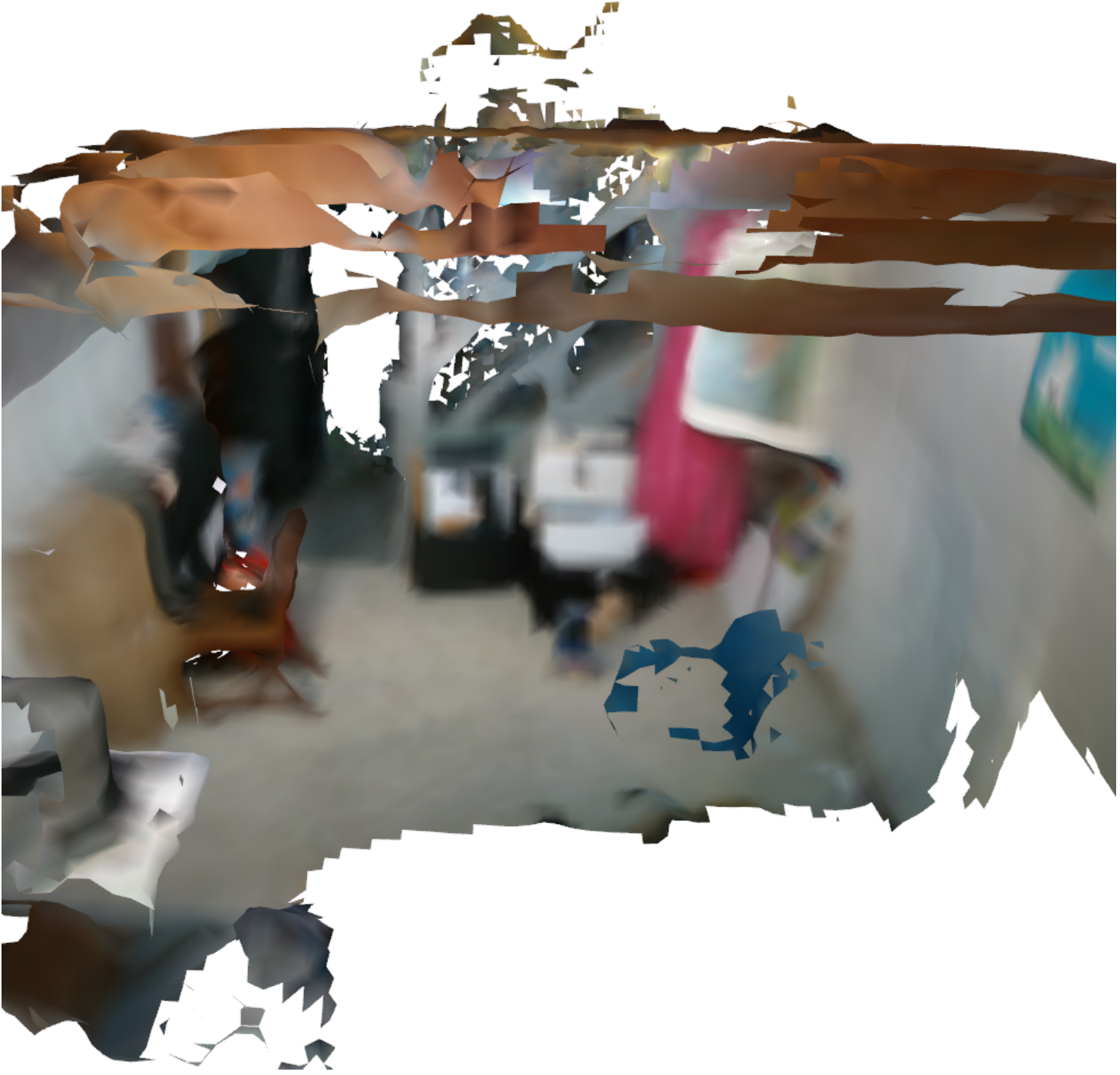}
    \caption{3D mesh of basement\label{fig:basement_mesh}}
  \end{subfigure}
  \hfill
  \begin{subfigure}[t]{.40\linewidth}
    \centering\includegraphics[clip,trim=0cm 0cm 0cm 0cm,width=.80\linewidth]{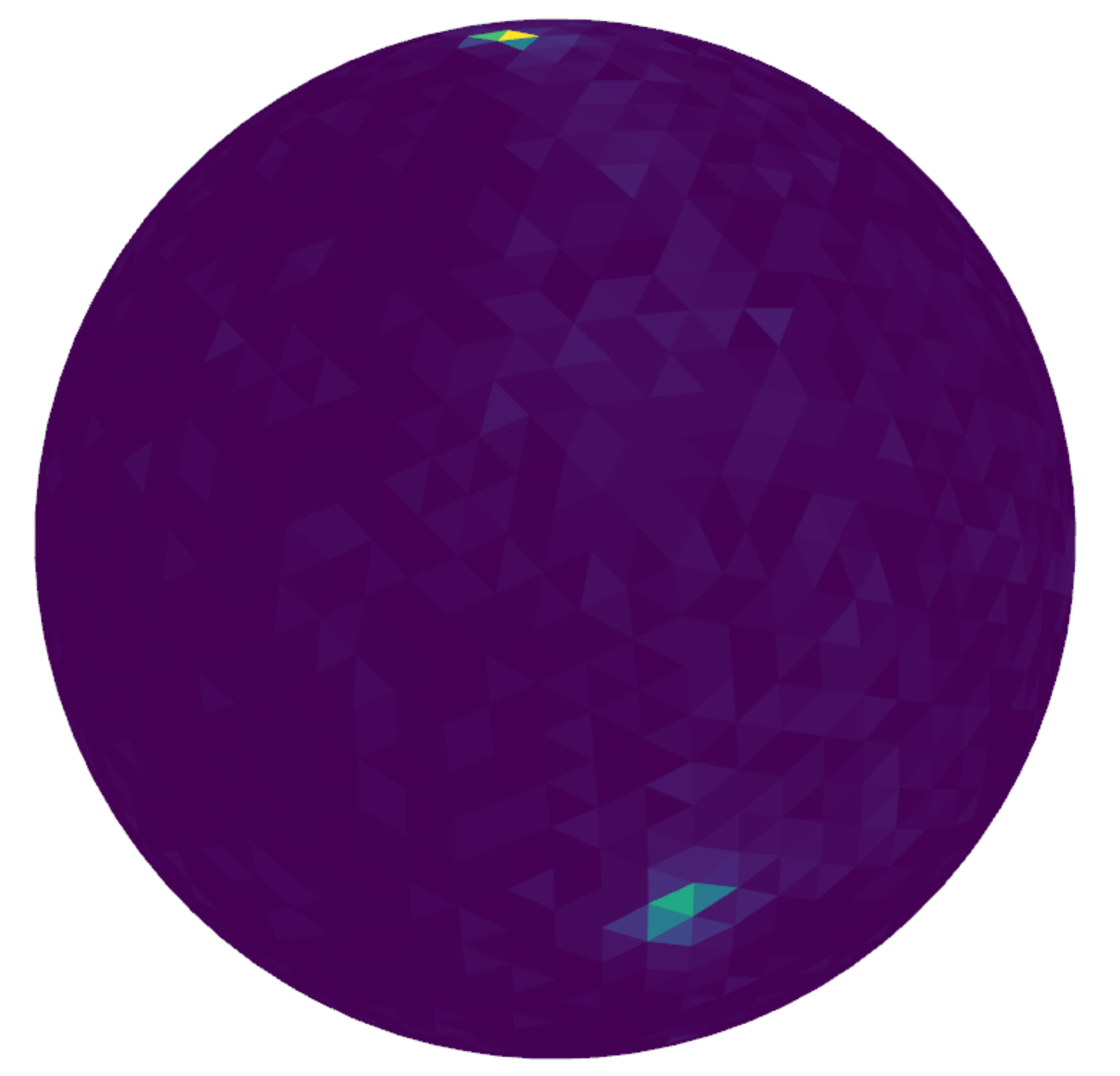}
    \caption{Colorized Gaussian Accumulator\label{fig:basement_ga}}
  \end{subfigure}
  \par\bigskip
  \begin{subfigure}[t]{.55\linewidth}
    \includegraphics[width=0.90\linewidth]{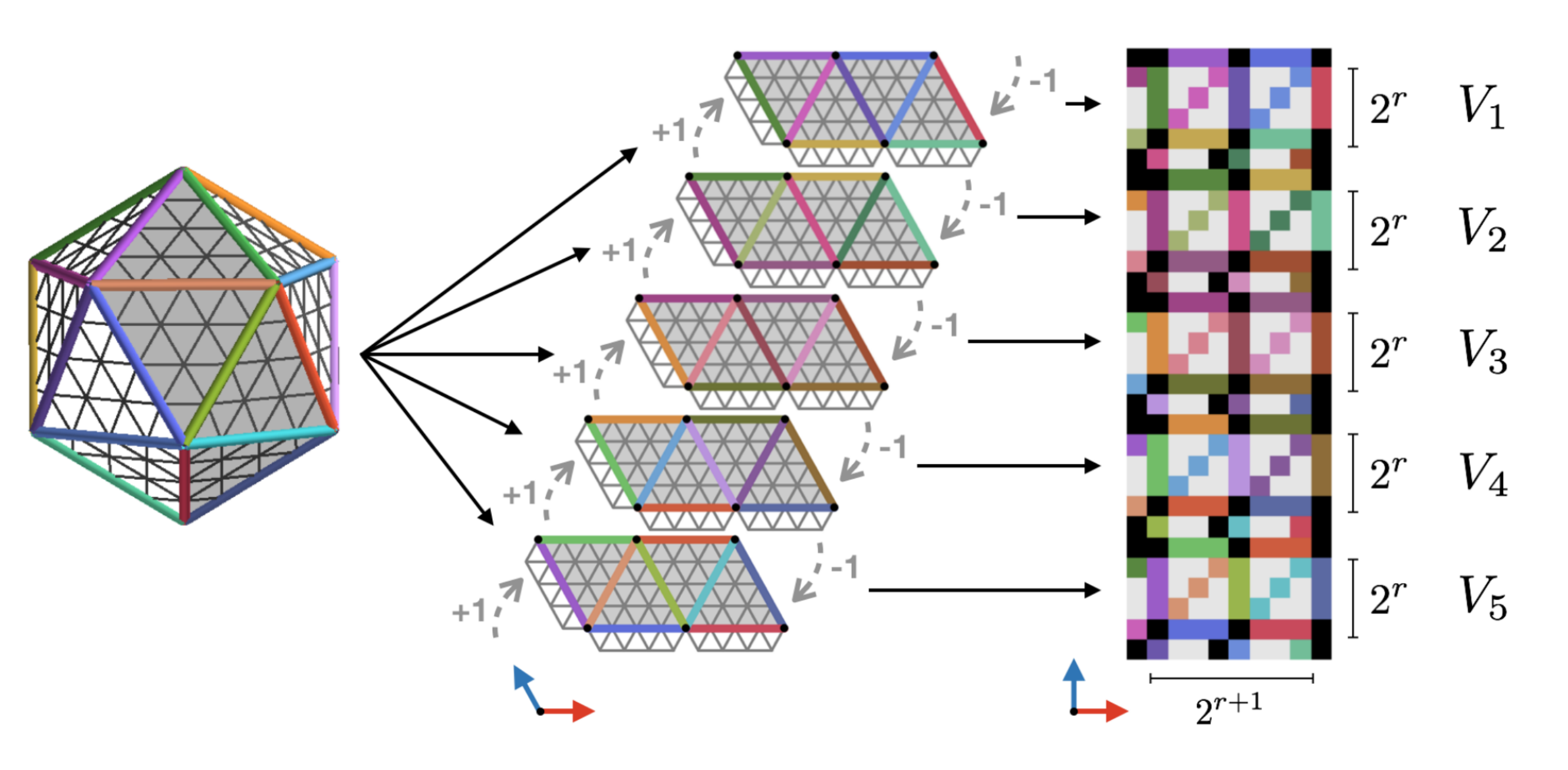}
    \caption{Unwrapping Process. Reprinted from Ref. \cite{cohen_gauge_2019} \\ Published open access under a CC-BY 4.0 license \\ http://proceedings.mlr.press/v97/cohen19d.html \label{fig:process_unwrap}}
  \end{subfigure}
  \hfill
  \begin{subfigure}[t]{.40\linewidth}
    \centering\includegraphics[clip,trim=0cm 0cm 0cm 0cm,width=.95\linewidth]{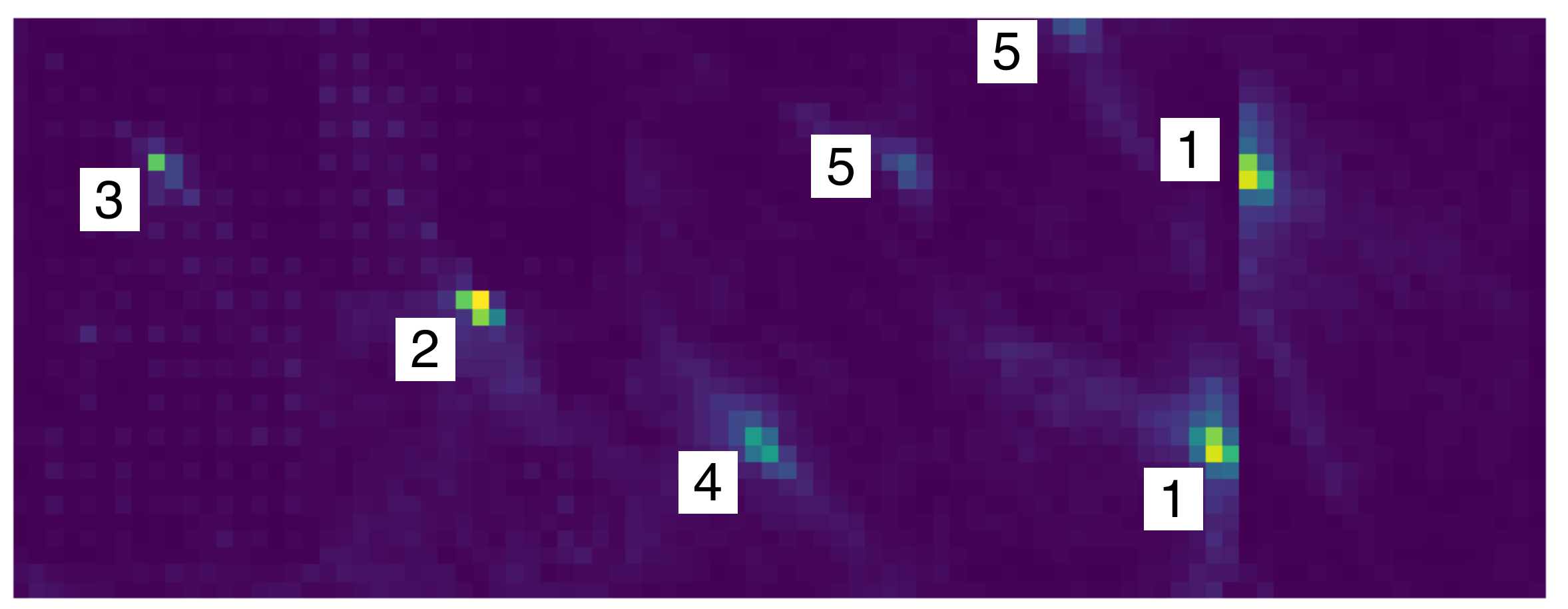}
    \caption{Unwrapped GA, rotated \label{fig:basement_unwrapped}}
  \end{subfigure}

  \caption{(\subref{fig:basement_mesh}) Example basement scene mesh. (\subref{fig:basement_ga}) Mesh triangle normals are integrated into the Gaussian Accumulator and colorized showing peaks for the floor and walls. (\subref{fig:process_unwrap}) Overview of unwrapping a refined icosahedron into a 2D image. Five overlapping charts are stitched together to create a grid. Padding between charts is accomplished by copying adjoining chart neighbors using the unwrapping process and its illustration from \cite{cohen_gauge_2019}. (\subref{fig:basement_unwrapped}) Unwrapped Gaussian Accumulator creating a 2D image used for peak detection.  White boxes indicate detected peaks. Duplicate peaks are merged (1 \& 5) with agglomerative hierarchical clustering. 
  }\label{fig:peak_detection}
\end{figure}


The histogram of the Gaussian Accumulator is normalized between the range [0-255]. Figure \ref{fig:basement_mesh} shows an example mesh of a basement where the dominant planes are the floor and walls. Figure \ref{fig:basement_ga} shows a colored visualization of the GA after integrating triangle normals of this mesh. Higher values are bright yellow; lower values are dark purple. Peaks representing the basement floor and walls are clearly visible near the top and side of the sphere, respectively. Note that more peaks exist on unseen sides of the sphere. We use the technique described in Ref. \cite{cohen_gauge_2019} to unwrap the refined icosahedron into a 2D image as shown in Figure \ref{fig:process_unwrap}.  The center image shows unwrapping of the icosahedron to create  five charts. The vertices of these five charts map to hard-coded correspondences of pixels in the right image. This requires every vertex take the average value of its neighboring triangles. Finally a one-pixel padding is performed on the edges of each chart by copying neighbors of adjoining charts. This creates duplicate pixel values on the bottom and left of the image as well as between charts. The end result is a 2D image guaranteed to provide equivariant convolution for kernels. The unwrapped image of the example GA is shown in the Figure \ref{fig:basement_unwrapped}.


We use a standard 2D peak detector algorithm to find local peaks in the image. A peak is in the center of a 3X3 pixel group if it is the maximum in the group and its value is higher than a user-configurable $v_{min}$. Once a peak is detected in the 2D image it is converted to its corresponding surface normal on the GA. Duplicate peaks may be detected near chart borders because of copy padding discussed above, or two peaks may be close together. In either case it is desirable to collapse them into a single peak.  Agglomerative hierarchical clustering (AHC) is used to merge these peaks and take their weighted average. AHC will only merge peaks whose Euclidean distance is less than $d_{peak}$. 

\subsubsection{Application to Polylidar3D}

Polylidar3D uses the Fast Gaussian Accumulator (FastGA) to estimate dominant plane normals. Triangle normals from the half-edge triangular mesh are input to FastGA. Not all triangle normals are needed to achieve acceptable results, so a user-configurable percent sampling parameter $sample_{pct}$ is used to reduce computational demand. After peak detection the $l$ unique dominant plane normals are returned as a list  $\mathcal{N}^d = \{\hat{n}^d_0, \ldots, \hat{n}^d_{l-1} \}$ for plane and polygon extraction. Note that alternative strategies of generating input normals such as fast RANSAC plane fitting with weighted voting may also be used \cite{limberger_real-time_2015}.

\section{Planar Segmentation and Polygon Extraction}\label{sec:methods_polylidar}

The following sections build upon our previous work in polygon extraction from 2D triangular meshes. Section \ref{sec:methods_polylidar_plane_extraction} describes planar segmentation while Section \ref{sec:methods_polylidar_polygon_extraction} outlines polygon extraction. 

\subsection{Planar Segmentation}\label{sec:methods_polylidar_plane_extraction}

The main input for planar segmentation is the half-edge triangular mesh, composed of $\mathcal{P}, \mathcal{T}$, $\mathcal{N}$, $\mathcal{HE}$, and the set of $l$ dominant plane normals $\mathcal{N}^d$. Polylidar3D performs parallelized and regularized triangle mesh region growing via partitioning with dominant plane normals. Triangles having similar normals to a dominant plane are grouped for region growing. Different groups are grown in parallel. This process is controlled through user-provided parameters including maximum triangle edge length $l_{max}$, minimum angular similarity $ang_{min}$, maximum point to plane distance $ptp_{max}$, minimum number of triangles $tri_{min}$, and minimum number of vertices in a hole $vertices^{hole}_{min}$. These parameters limit the maximum distance between points for spatial connectivity, ensure common normal orientation in planar segments, force planar constraints, and remove spurious/small planes and holes. Note that $ang_{min}$ is computed from the dot product between a triangle normal and its closest dominant plane normal; a value of 1.0 requires exact alignment while a value of 0.96 allows a $\approx 14^\circ$ difference. 

\begin{algorithm}[ht]
    \SetKwInOut{Input}{Input}
    \SetKwInOut{Output}{Output}
    \SetStartEndCondition{ }{}{}%
    \SetKwProg{Fn}{def}{\string:}{}
    \SetKw{KwIs}{is}{}
    \SetKwIF{If}{ElseIf}{Else}{if}{:}{elif}{else:}{}%
    \SetKw{KwContinue}{continue}{}

    \Input{Triangle Set: $\mathcal{T}$, Point Cloud: $\mathcal{P}$, Triangle Normals: $\mathcal{N}$\\
           Dominate Plane Normals: $\mathcal{N}^d$, Max Length: $l_{max}$, Min Angular Similarity: $ang_{min}$ \\
           }
    \Output{Triangle Group Set: $\mathcal{G}$}
    \texttt{SentinelValue} = 255 \\
    $k = |\mathcal{T}|$ \\
    $l = |\mathcal{N}^d |$ \\
    \tcc{Loop through every triangle}       
    \For{$t \leftarrow 0$ \KwTo  $k$ \hspace{.1cm} }{
        \texttt{edge\_length} = $\operatorname{GetMaximumTriangleEdgeLength}(t, \mathcal{T}, \mathcal{P})$ \\
        \uIf{\texttt{edge\_length} > $l_{max}$ }{
            $\mathcal{G}[t]$ = \texttt{SentinelValue} \\
            \KwContinue
        }
        \texttt{max\_similarity} = -1.0 \\ 
        \tcc{Loop through every dominant plane normal}   
        \For{$j \leftarrow 0$ \KwTo  $l$ \hspace{.1cm} }{
            \texttt{similarity} = $\hat{n}_t \cdot \hat{n}^d_j$ \\
            \uIf{\texttt{similarity} > \texttt{max\_similarity} }{
                $\mathcal{G}[t] = j $\\
                \texttt{max\_similarity} = \texttt{similarity}
            }
        }
        \uIf{\texttt{max\_similarity} < $ang_{min}$ }{
            $\mathcal{G}[t]$ = \texttt{SentinelValue}
        }
    }
    return $\mathcal{G}$
    \caption{Group Assignment}
    \label{alg:group_assignment}
\end{algorithm}

\begin{figure}[t]
  \begin{subfigure}[t]{.22\linewidth}
    \centering\includegraphics[clip,trim=0cm 0cm 0cm 0cm, width=.99\linewidth]{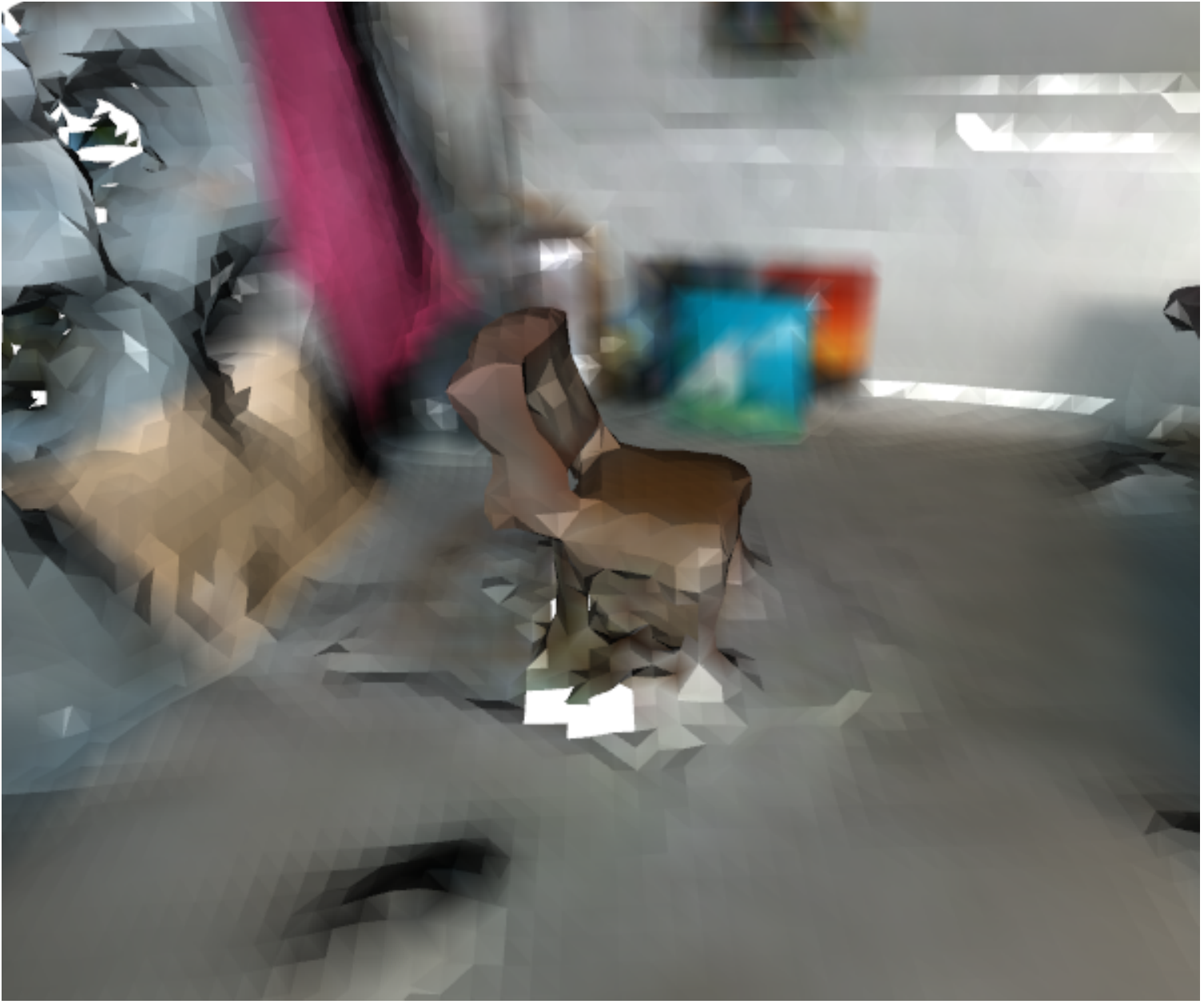}
    \caption{Mesh\label{fig:planar_seg_a}}
  \end{subfigure}
  \hfill
  \begin{subfigure}[t]{.22\linewidth}
    \centering\includegraphics[clip,trim=0cm 0cm 0cm 0cm,width=.99\linewidth]{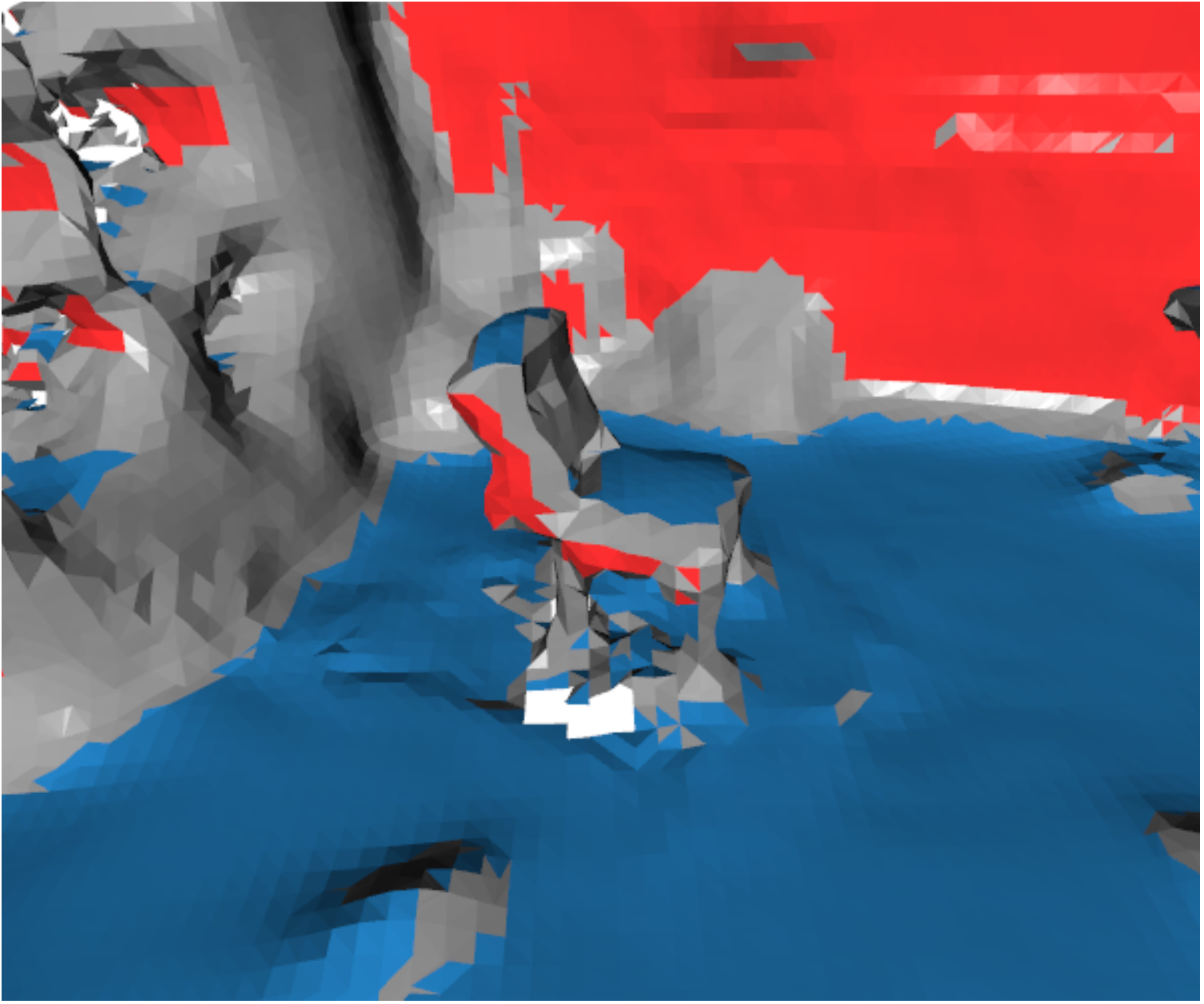}
    \caption{Group Assignment\label{fig:planar_seg_b}}
  \end{subfigure}
  \hfill
  \begin{subfigure}[t]{.22\linewidth}
    \centering\includegraphics[clip,trim=0cm 0cm 0cm 0cm,width=.99\linewidth]{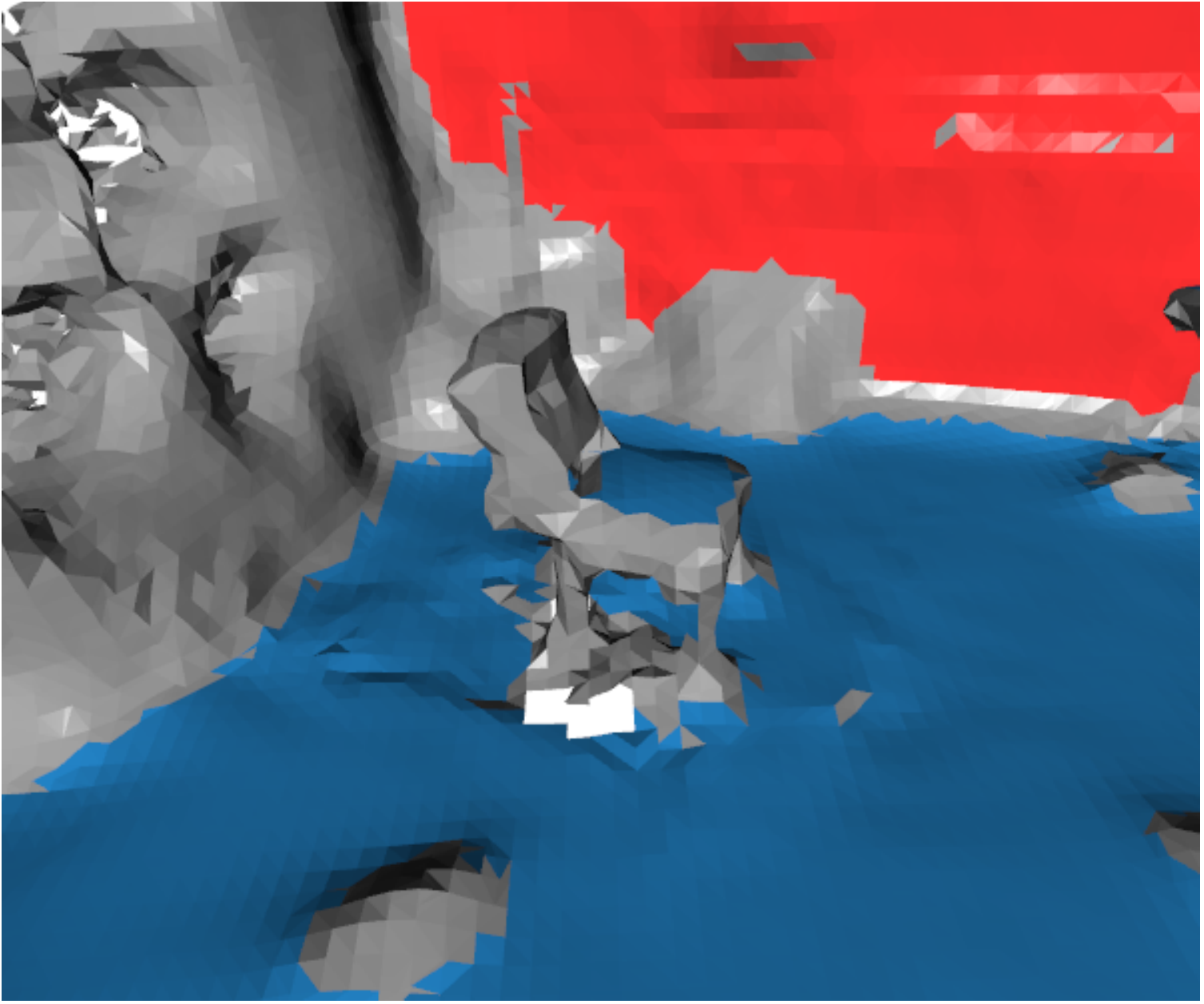}
    \caption{Planar Segments\label{fig:planar_seg_c}}
  \end{subfigure}
  \hfill
  \begin{subfigure}[t]{.22\linewidth}
    \centering\includegraphics[clip,trim=0cm 0cm 0cm 0cm,width=.99\linewidth]{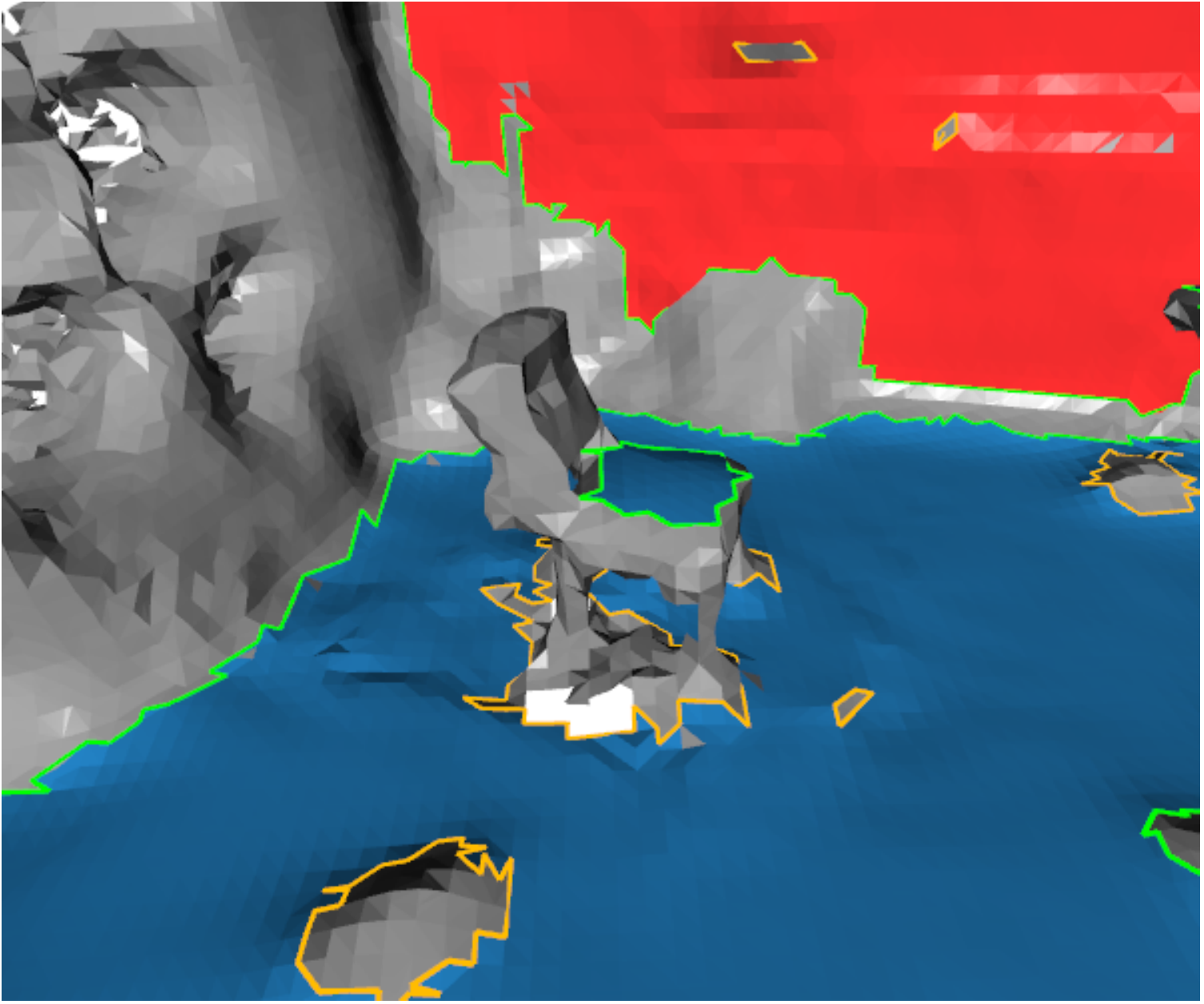}
    \caption{Extracted Polygons\label{fig:planar_seg_d}}
  \end{subfigure}

  \caption{(\subref{fig:planar_seg_a}) An example mesh to demonstrate planar segmentation and polygon extraction using two dominant plane normals, represented by the floor and wall.
  (\subref{fig:planar_seg_b}) Every triangle is inspected for filtering and clustered through group assignment. Blue and red triangles meet triangle edge length constraints and are within an angular tolerance of the floor and wall surface normals, respectively. (\subref{fig:planar_seg_c}) Region growing is performed in parallel for the blue and red triangles. The top chair surface and floor are distinct planar segments.  (\subref{fig:planar_seg_d}) Polygonal representations for each planar segment are shown. The green line represents the concave hull; the orange line depicts any interior holes. Note that small segments and small interior holes are filtered. 
  }\label{fig:planar_segmenti}
\end{figure}

We first create triangle group array $\mathcal{G}$ to store group labels for each of the $k$ triangles in $\mathcal{T}$. Algorithm \ref{alg:group_assignment} outlines this procedure and begins with iterating through all triangles (Line 4). $\mathcal{G}$ is composed of 8-bit unsigned integers [0-255] with 255 being a reserved sentinel value indicating a triangle does not belong to any planar segment. The following steps filter unused triangles and cluster triangles by normal orientation. The first geometric predicate (Line 5) removes triangles whose edge length exceeds a user-specified value.  Lines 9-15 iterate though all dominant plane normals finding the one most similar to the triangle's surface normal $\hat{n}_t$. Line 16 performs a check to ensure the triangle normal is within an angular tolerance of its nearest dominant plane normal. If a triangle is assigned the group 255 it will not participate in subsequent region growing. Using 8-bit integers limits the maximum number of dominant plane normals extracted to 254. This procedure is iteration-independent and is parallelized by OpenMP \cite{openmp08}.  Figures \ref{fig:planar_seg_a} and \ref{fig:planar_seg_b} show an example input mesh and color-coded group assignments, respectively. In this example the floor (blue) and the wall (red) are the two dominant plane normals to be extracted. Note that the seat of the chair is assigned the same group label as the floor, and that superfluous triangles are also assigned in the top left of Fig. \ref{fig:planar_seg_b}. 

Region growing is decomposed using \emph{task-based} parallelism, where $l$ dominant plane normals create $l$ separate tasks of regions growing. These tasks are executed in parallel by a threadpool and can themselves spawn additional dynamic tasks \cite{Huang2019}. Each independent task performs a serial region growing procedure that is similar to our previous work on 2D meshes \cite{polylidar2D} and was inspired by \cite{cao_roof_2017}. Algorithm \ref{alg:region_growing} outlines this procedure for a single group $g$. The routine begins by creating empty sets to store planar triangular segments and their corresponding polygonal representations, denoted $\mathcal{T}^g$ and $\mathcal{PL}^g$.  An iterative plane extraction procedure begins with a seed triangle $t$  verified to belong to group $g$ (Line 5). Subroutine \texttt{ExtractPlanarSegment} uses the seed triangle to create edge-connected triangular subsets from $\mathcal{T}$ which have the same group label in $\mathcal{G}$ and meet user-provided planarity constraints (Line 7). If a user-specified minimum number of triangles is met then this set, $\mathcal{T}^g_i$, is added to $\mathcal{T}^g$. A dynamic task is then created to perform polygon extraction for this segment (Line 10). This procedure call is non-blocking; the region growing task continues to extract any remaining spatially connected planar segments before terminating. This means planar segmentation and polygon extraction may occur in parallel if multi-core is enabled.

Figure \ref{fig:planar_seg_c} shows three planar segments extracted that represent the floor, chair seat, and wall. The floor and chair surfaces have similar surface normals but are not spatially connected so independent planar segments and corresponding polygons are created. The small bump on the floor did not meet the planarity constraints (configured with $ptp_{max}$) thus is not included in the floor planar segment. This hole in the mesh will be extracted as an explicit interior hole of a polygon. The wall surface belongs to a separate group and is extracted in parallel with the floor and chair. 


\begin{algorithm}[ht]
    \SetKwInOut{Input}{Input}
    \SetKwInOut{Output}{Output}
    \SetStartEndCondition{ }{}{}%
    \SetKwProg{Fn}{def}{\string:}{}
    \SetKw{KwIs}{is}{}
    \SetKwIF{If}{ElseIf}{Else}{if}{:}{elif}{else:}{}%
    \SetKw{KwContinue}{continue}{}

    \Input{Triangle Set: $\mathcal{T}$, Point Cloud: $\mathcal{P}$, Half Edge Set: $\mathcal{HE}$, Triangle Group Set: $\mathcal{G}$ \\
           Dominate Plane Normal: $n^d$, Dominate Plane Label: $g$ \\
           Point To Plane: $ptp_{max}$, Min Triangles: $tri_{min}$, Min Hole Vertices: $vertices^{hole}_{min}$ \\
           }
    \Output{Planar Segment Set: $\mathcal{T}^g$, Polygon Set: $\mathcal{PL}^g$}
    $\mathcal{T}^g$ = $\emptyset$ \\
    $\mathcal{PL}^g$ = $\emptyset$ \\
    $k = |\mathcal{T}|$ \\
    \tcc{Loop through every triangle}       
    \For{$t \leftarrow 0$ \KwTo  $k$ \hspace{.1cm} }{
        \uIf{$\mathcal{G}[t]$ $\neq g$}{
            \KwContinue
        }
        $\mathcal{T}^g_i = \operatorname{ExtractPlanarSegment}(t, \mathcal{T}, \mathcal{P}, \mathcal{HE},\mathcal{G},  n^d, ptp_{max}$) \\
        \uIf{|$\mathcal{T}^g_i| > tri_{min}$}{
            $\mathcal{T}^g$ = $\mathcal{T}^g$ + $\mathcal{T}^g_i$ \\
            $\mathcal{PL}^g = \mathcal{PL}^g + \operatorname{SpawnTask}(\texttt{PolygonExtraction}, \mathcal{T}^g_i, vertices^{hole}_{min})$
        }
    }
    $\operatorname{WaitForTasks}$\\
    return $\mathcal{T}^g, \mathcal{PL}^g$
    \caption{Region Growing Task}
    \label{alg:region_growing}
\end{algorithm}

\subsection{Polygon Extraction}\label{sec:methods_polylidar_polygon_extraction}

\begin{figure}[t]
  \begin{subfigure}[t]{.33\linewidth}
    \centering\includegraphics[clip,trim=1cm 1cm 1cm 3cm, width=.99\linewidth]{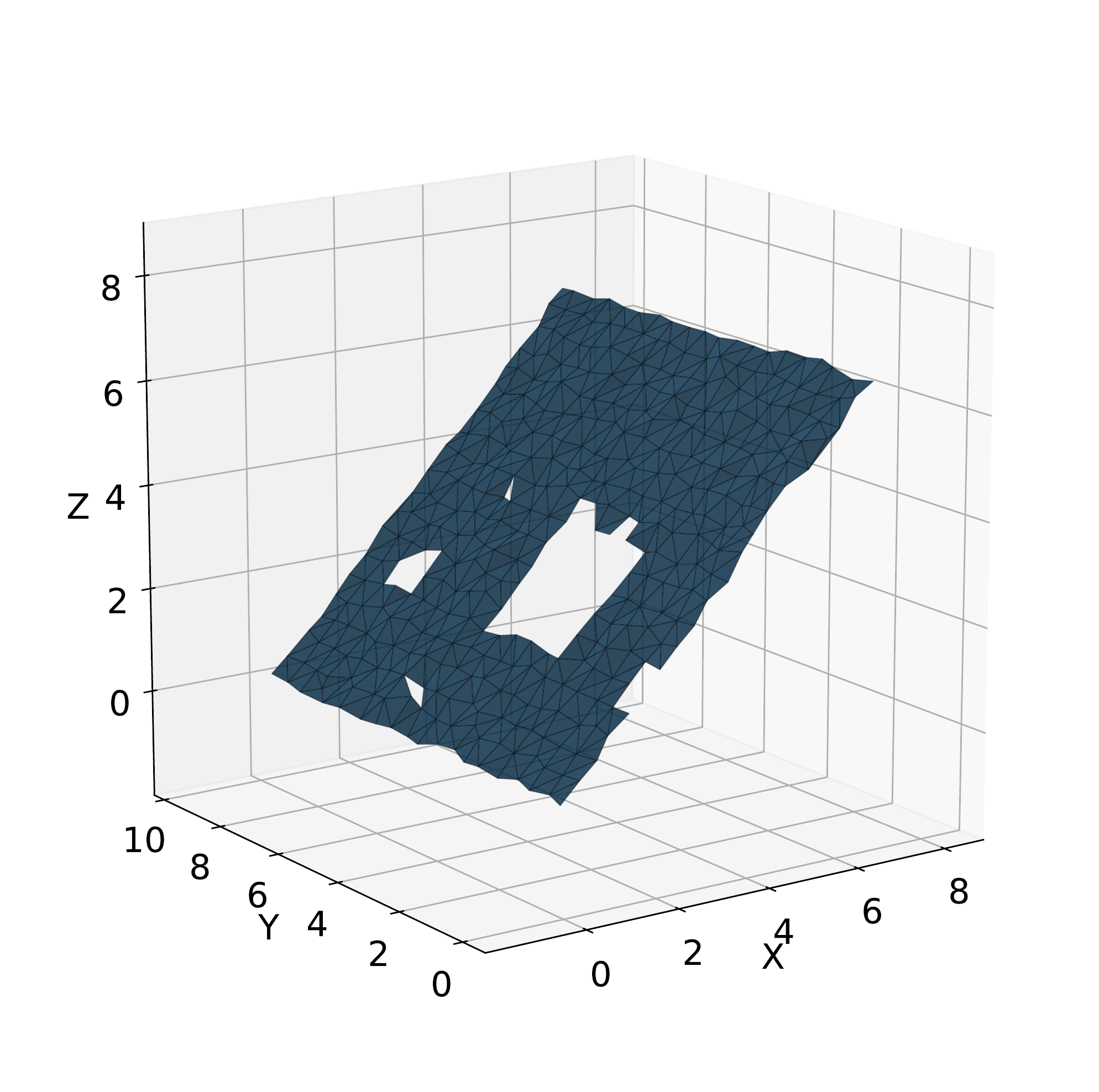}
    \caption{\label{fig:polygon_extraction_a}}
  \end{subfigure}
  \hfill
  \begin{subfigure}[t]{.30\linewidth}
    \centering\includegraphics[clip,trim=0cm 0cm 0cm 0cm,width=.99\linewidth]{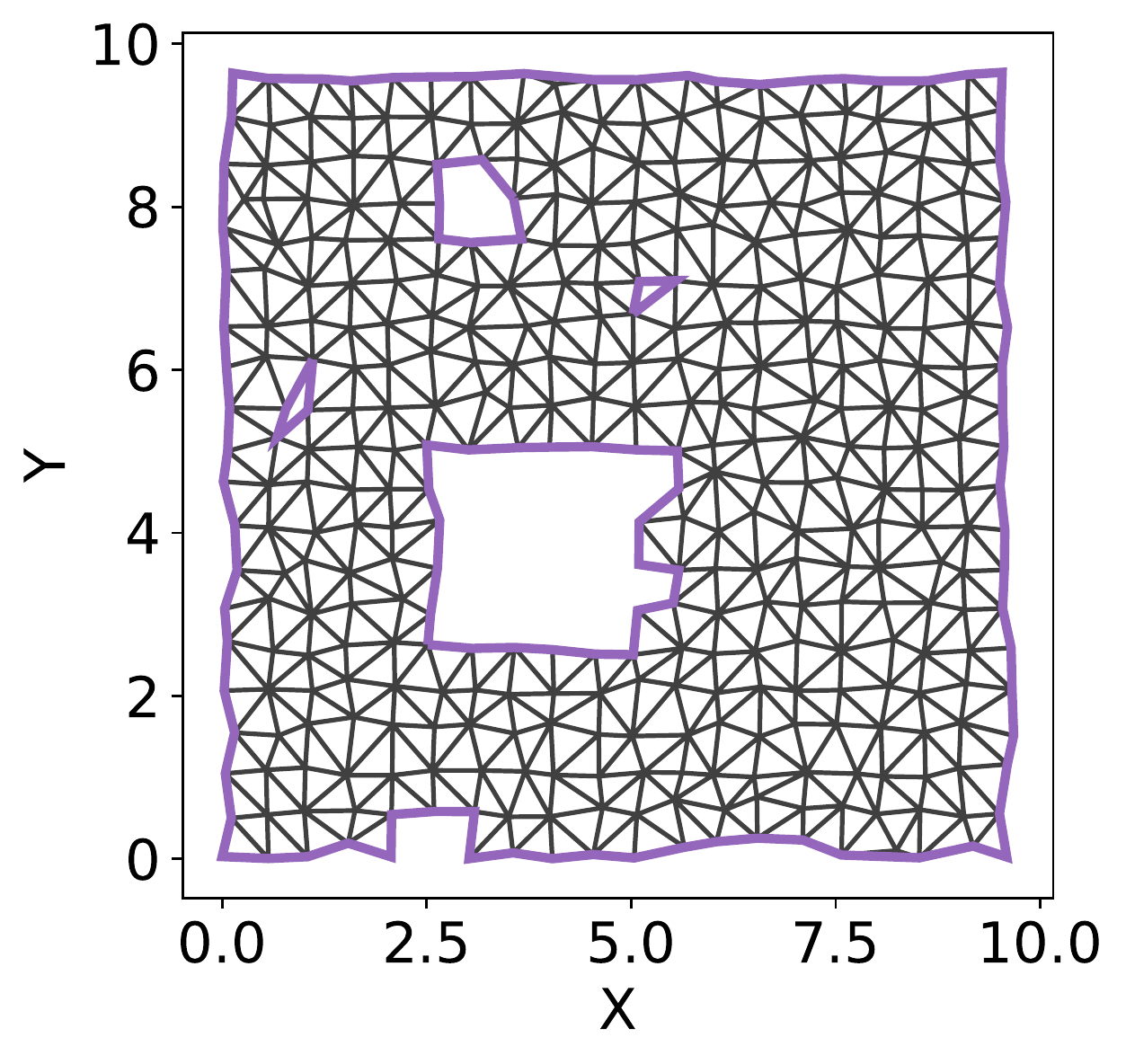}
    \caption{\label{fig:polygon_extraction_b}}
  \end{subfigure}
  \hfill
  \begin{subfigure}[t]{.30\linewidth}
    \centering\includegraphics[clip,trim=0cm 0cm 0cm 0cm,width=.99\linewidth]{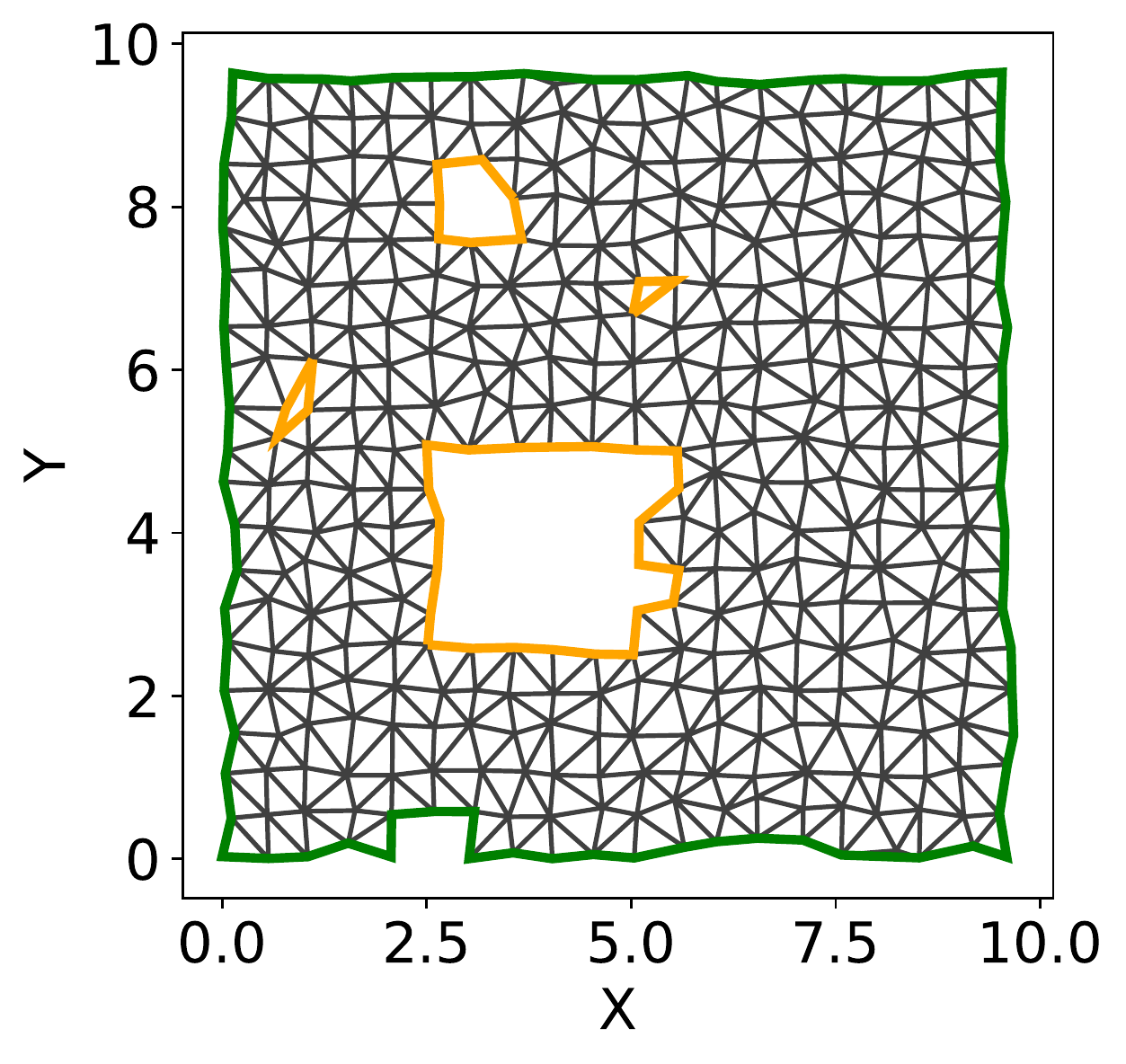}
    \caption{\label{fig:polygon_extraction_c}}
  \end{subfigure}
  \caption{(\subref{fig:polygon_extraction_a}) Example extracted planar triangular segment $\mathcal{T}^{g}_i$. Note the four holes in the mesh. (\subref{fig:polygon_extraction_b}) Projection of a triangle segment to a geometric plane. Only border edges (purple) are actually needed for projection. (\subref{fig:polygon_extraction_c}) A polygon is extracted from border edges with a concave hull (green) and multiple interior holes (orange). }\label{fig:polygon_extraction}
\end{figure}

Polygon extraction is performed on each planar mesh segment $\mathcal{T}^{g}_i$. Each polygon is defined by a single linear ring of points representing the concave hull/shell and a (possibly empty) set of linear rings representing interior holes. The same boundary following method we proposed in our previous work \cite{polylidar2D} is used with small modifications because triangular meshes are no longer 2D.  Polygons are defined in a 2D subspace and are provided explicit guarantees through their definition per \cite{herring2006opengis}. For example the edges in linear rings must not cross in this 2D space.  For this reason boundary following in polygon extraction is carried out in the 2D projection of $\mathcal{T}^{g}_i$ on its geometric plane. Note that only the boundary edges of $\mathcal{T}^{g}_i$ need to be projected. Figure \ref{fig:polygon_extraction} shows the projection of $\mathcal{T}^{g}_i$ to its geometric plane and extraction of its polygonal representation. The three main components of polygon extraction are:

\begin{enumerate}
    \item Data Structure Initialization
    \item Extract Exterior Hull/Shell
    \item Extract Interior Holes
\end{enumerate}

The data structure initialization identifies all boundary half-edges inside $\mathcal{T}^{g}_i$ which are highlighted in purple in Figure \ref{fig:polygon_extraction_b} and denoted $\mathcal{BE}$. Additionally a mapping between point indices and these boundary half-edges are created denoted $PtE$. Finally any point on the exterior on the shell is found denoted $pi_{xp}$. The outer exterior shell is then extracting beginning with $pi_{xp}$. Boundary following is performed by progressively building a linear ring by following each points outgoing half-edge(s) using $PtE$. Special routines handle scenarios when a hole is connected to the exterior hull. After the hull is extracted any interior holes remaining are extracted with the same special routines to handle rare scenarios when holes are connected.

The only modification to our previous work occurs in projecting the boundary edges. This is needed in finding $pi_{xp}$ and for the special routines in handling multiple outgoing edges during boundary following.
Note that the final polygons returned are represented as point indices in $\mathcal{P}$. The underlying 3D structure of the polygon is retained, i.e., it will follow a noisy surface per Figure \ref{fig:planar_seg_d}.  The polygon can also be projected to the surface's geometric plane as described in Section \ref{sec:methods_polylidar_polygon_filtering} for post-processing.

Given noisy and dense planar mesh segments, border edges may cross during projection to the geometric plane.  When this occurs an invalid polygon will be generated, most often a small self-intersection. This issue does not occur with unorganized point clouds because they are projected to the $x$-$y$ plane where triangulation has already taken place; this guarantees edges do not cross. However planar segments from user-provided meshes and organized point clouds may be projected to arbitrary geometric planes. Additionally the tolerance in ``flatness'' of the planar triangular segment is user-configurable.  This issue, if it occurs, is managed in polygon post-processing as described in Section \ref{sec:methods_polylidar_polygon_filtering}. Although rare, if this condition must be handled before post-processing one might instead project all vertices of $\mathcal{T}^{g}_i$ to the geometric plane and perform polygon extraction on the 2D point set as shown in our previous work \cite{polylidar2D}.

\subsection{Algorithm Parallelization}\label{sec:methods_polylidar_parallel}

Planar segmentation and polygon extraction use both data and task-based parallelism. We use \texttt{OpenMP} for data parallelism which is carried out in ``hot'' loops that are iteration independent, e.g., triangle group assignment in Algorithm \ref{alg:group_assignment} as well as computing triangle normals. We use the \texttt{MARL} library to handle task scheduling and synchronization primitives \cite{marl}. Note that region growing of a single dominant plane normal is still a serial process as is the polygon extraction process of a single planar triangular segment. Therefore if only one dominant plane normal exists than task-based parallelism will provide minimal speed up. However group assignment  is still fully parallelized. Benefits of parallelism are further explored in Section \ref{sec:results_meshes_parallel} experiments where speedup is calculated as number of threads and number of dominant plane normals vary.

\section{Post Processing}\label{sec:methods_polylidar_polygon_filtering}

The polygons returned by Polylidar3D can be further processed to improve visualization and filter superfluous polygons and/or holes. All operations are implemented on the 2D projection of the polygon on its geometric plane. The following sequential operations are executed:

\begin{enumerate}
    \item Polygon is simplified with parameter $\alpha$ 
    \item Polygon is buffered outward by parameter $\beta_{pos}$ meters 
    \item Polygon is buffered inward by  parameter $\beta_{neg}$ meters 
    \item Polygon is removed if its area is less than $\gamma$ meters 
    \item Interior holes are removed if area is less than $\delta$ meters 
\end{enumerate}

\begin{figure}[t]
  \begin{subfigure}[t]{.30\linewidth}
    \centering\includegraphics[clip,trim=0cm 0cm 0cm 0cm, width=.99\linewidth]{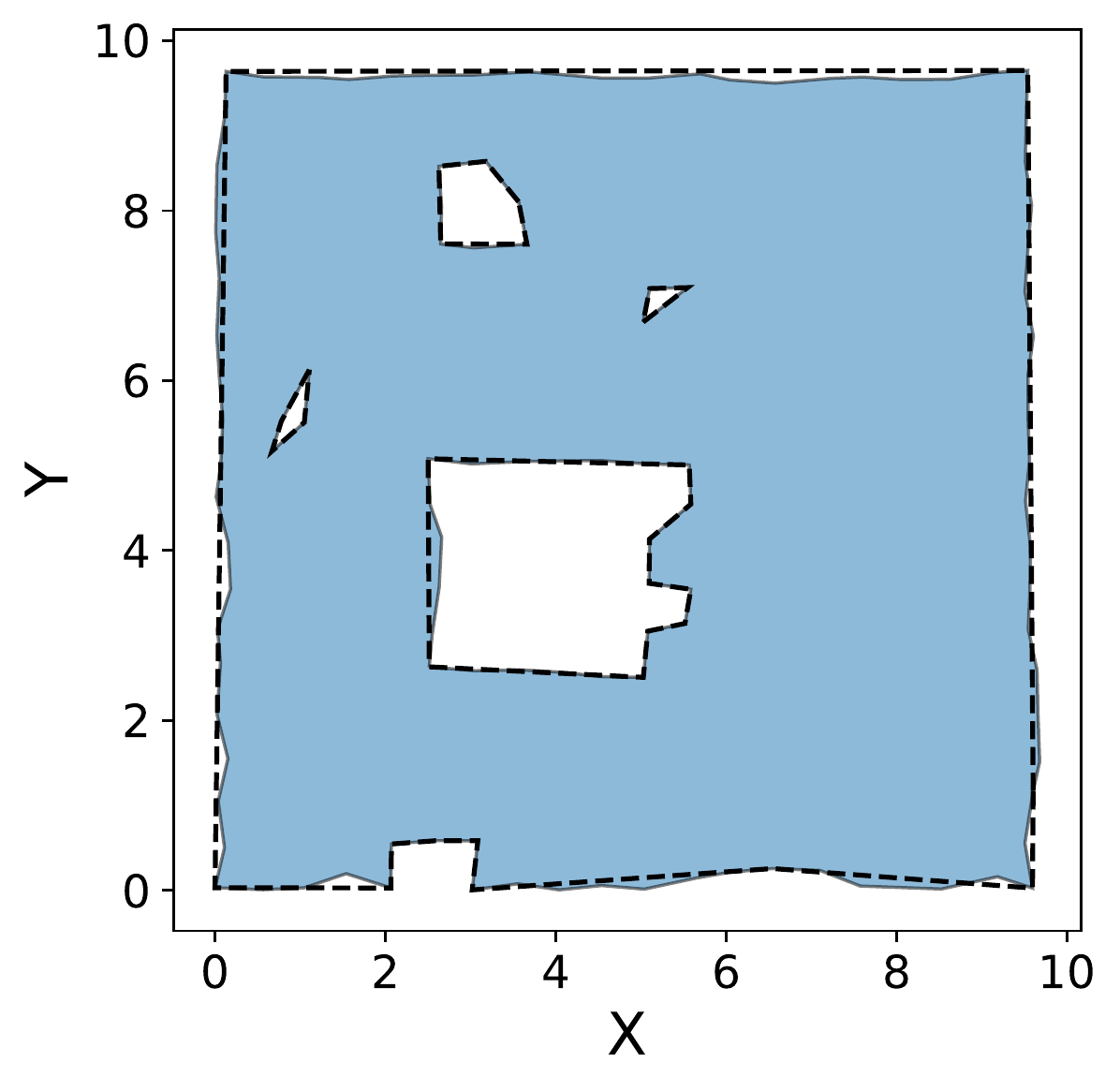}
    \caption{\label{fig:polygon_post_a}}
  \end{subfigure}
  \hfill
  \begin{subfigure}[t]{.30\linewidth}
    \centering\includegraphics[clip,trim=0cm 0cm 0cm 0cm,width=.99\linewidth]{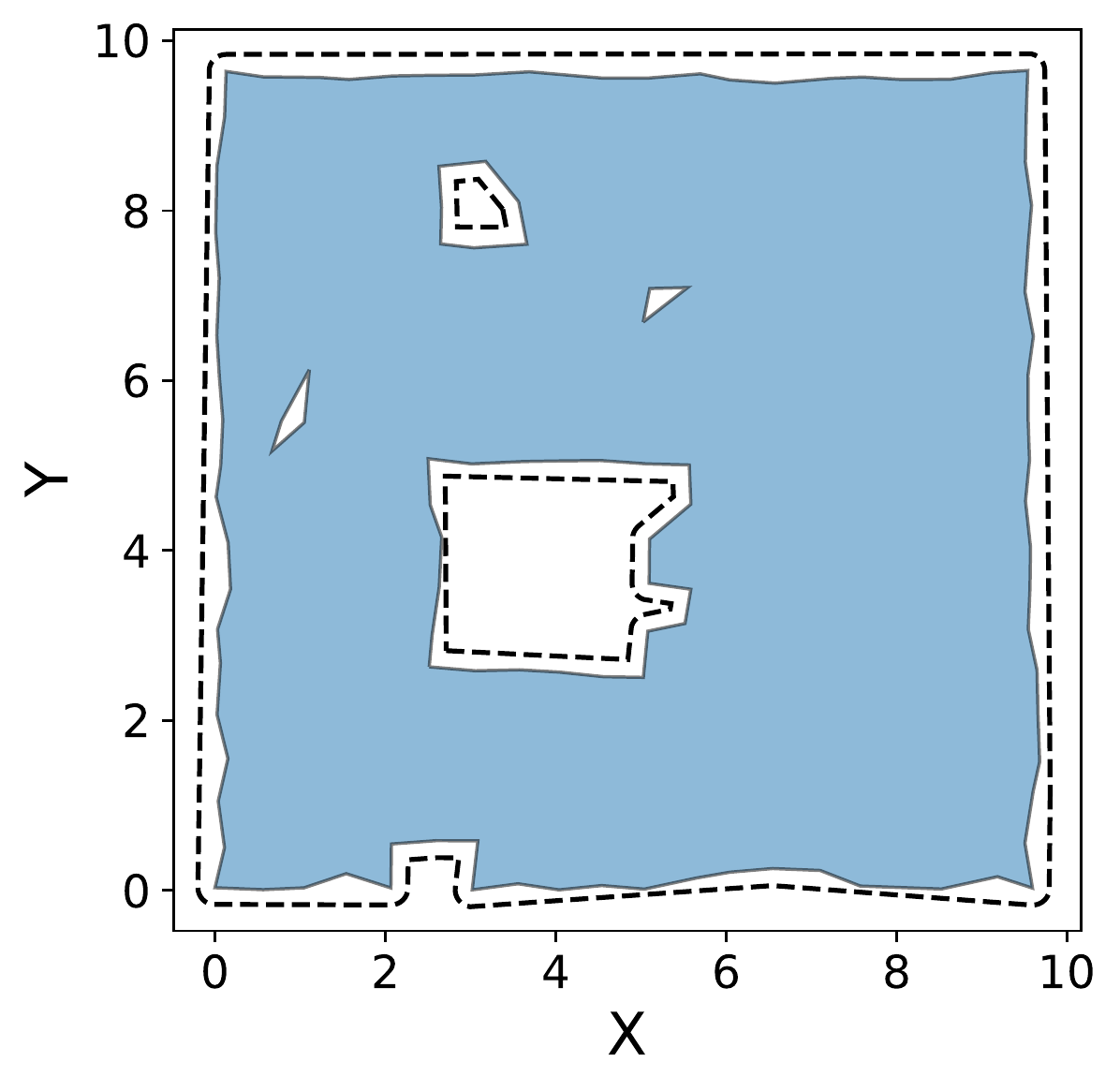}
    \caption{\label{fig:polygon_post_b}}
  \end{subfigure}
  \hfill
  \begin{subfigure}[t]{.30\linewidth}
    \centering\includegraphics[clip,trim=0cm 0cm 0cm 0cm,width=.99\linewidth]{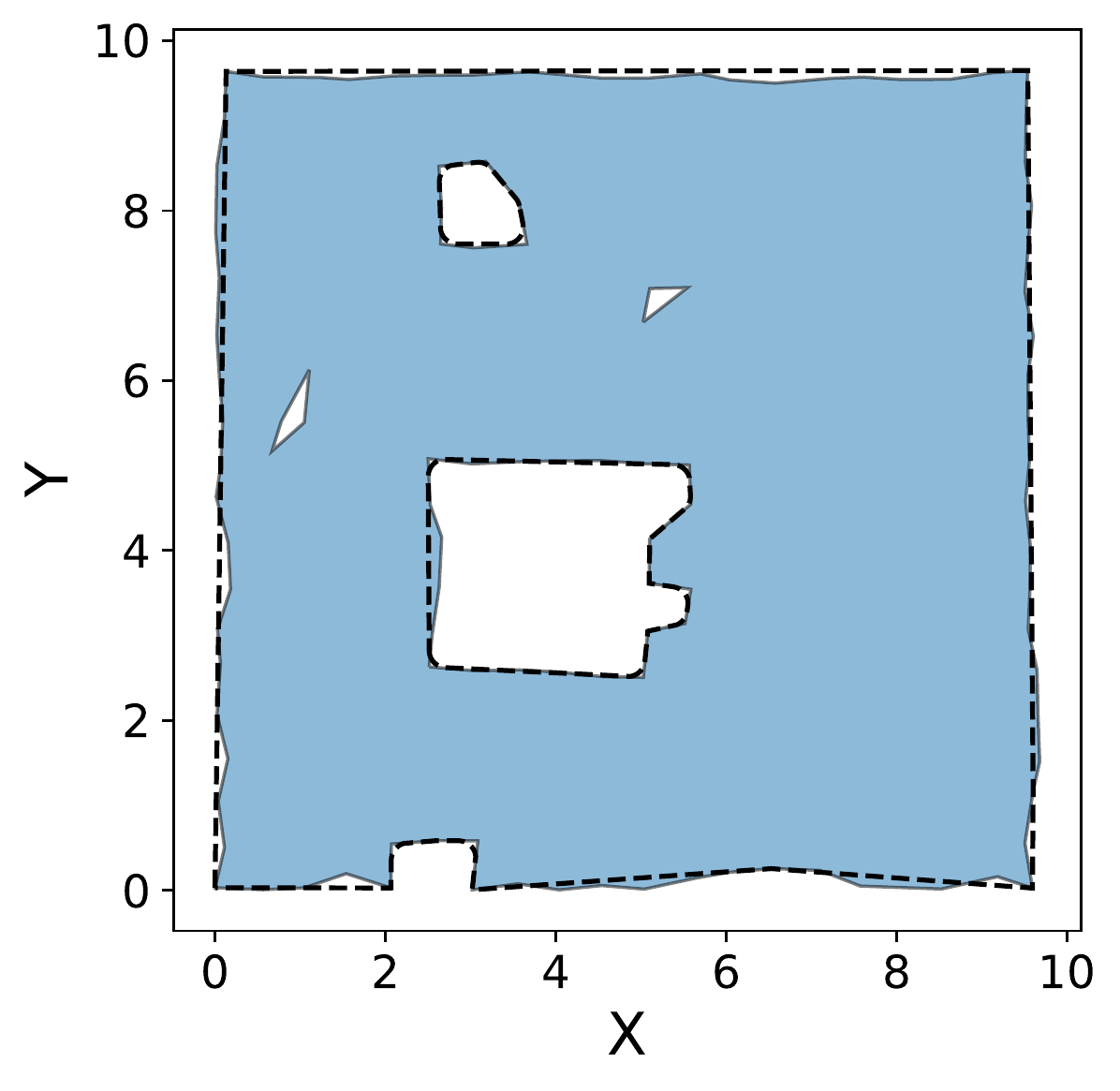}
    \caption{\label{fig:polygon_post_c}}
  \end{subfigure}
  \caption{An example of polygon post processing. The shaded blue polygon is the original polygon extracted (see Figure \ref{fig:polygon_extraction_c}). All dashed lines indicate a new polygon generated from a step of post processing. (\subref{fig:polygon_post_a}) The polygon is simplified. (\subref{fig:polygon_post_b}) The polygon is applied a positive buffer. Two small holes have been ``filled'' in. (\subref{fig:polygon_post_c}) The polygon is applied a negative buffer; only two holes remain. }\label{fig:polygon_post}
\end{figure}

The simplification algorithm is used to remove redundant vertices and ``smooth'' the polygon \cite{whyatt1988douglas}. The $\alpha$ parameter indicates the maximum distance between any point in the new polygon from the original. This reduction of superfluous vertices also decreases the computational demand for subsequent buffering. The buffering process is defined as the Minkowski difference of the polygon with a circle of radius equal to a buffer distance $\beta$ \cite{flato2000robust}. A positive buffer will expand a polygon and may fill in holes, while a negative buffer enlarges holes and recedes the concave hull.  A positive buffer will fix any small self-intersections that may have occurred during the projection. Small polygons and/or interior holes are then filtered by area. Currently all of these steps are single threaded and handled in Python using the geometry processing library Shapely which binds to the C++ GEOS library \cite{shapely}. An example of the first three steps of this process are shown in Figure \ref{fig:polygon_post}.

\section{Results}\label{sec:results}

We present several examples of our methods applied to real-world and synthetic 3D data. Section \ref{sec:results_dominant} provides execution time benchmarks evaluating the speed of our proposed Fast Gaussian Accumulator. Section \ref{sec:results_unorganized} shows examples of Polylidar3D applied to unorganized 3D points including airborne LiDAR point clouds and point clouds generated on a moving vehicle.  Section \ref{sec:results_orgnaized} shows Polylidar3D applied to organized point clouds including RGBD cameras as well as a challenging synthetic benchmark set. Finally Section \ref{sec:results_meshes_parallel} presents Polylidar3D applied to 3D meshes and explores how polygon extraction scales with additional CPU cores. 

All experiments/benchmarks use the same consumer desktop computer. The CPU is an AMD Ryzen 3900X 12 Core CPU with a frequency at 4.2 GHz equipped with 32 GB of RAM.  Note that all results obtained with CPU parallelization are annotated with number of threads used; the default is four. An NVIDIA GeForce RTX 2070 Super is used for GPU acceleration.

\subsection{Dominant Plane Normal Estimation}\label{sec:results_dominant}

This section evaluates the Fast Gaussian Accumulator (FastGA) proposed for dominant plane normal estimation. We specifically analyze CPU execution time needed to integrate a set of $k$ unit normals into the accumulator with $k$ varied over the tests. Per Section \ref{sec:methods_fastga_ga} we use sorted integer search coupled with local neighborhood search instead of $K$-$D$ trees \cite{toony_describing_2015}. To allow comparison, we created an alternative $K$-$D$ tree Gaussian Accumulator implementation that uses nanoflann, a high performance C++ $K$-$D$ tree library \cite{blanco2014nanoflann}. A leaf size of eight is used which offers the best results for our test cases.  Results were generated on two test sets with hundreds of runs to provide statistically significant results \cite{google_benchmark}. All benchmark code is open source \cite{fastgacode}. 

A GA with refinement level four (5120 triangle cells) was used for all tests. The first test generated 100,000 randomly distributed surface normals on the unit sphere and integrated them into the GA. The second test integrated all 60,620 triangle normals from the basement mesh previously shown in Figure \ref{fig:basement_mesh}. Recall that the GA is fixed once refinement level is chosen, so building the $K$-$D$ tree index is not part of execution timing. Results of integrating all $k$ normals for each test set are shown in Table \ref{table:results_fastga}. FastGA is more than two times faster than using a $K$-$D$ tree, though the $K$-$D$ tree implementation is also fast and could be used as an alternative method if desired.  We can conclude that exploiting the known fixed structure of triangular cells on S2 (using space filling curves and sorted integer search) outperforms a general purpose $K$-$D$ tree method. 

\begin{table}[!htb]
    \centering
    \caption{Execution Time Comparisons for Synthetic and Real World Datasets} \label{table:results_fastga}
    \begin{subtable}[t]{.5\linewidth}
        \centering
        \caption{Synthetic: 100,000 Random Normals}\label{table:results_fastga_a}
        \begin{tabular}{@{}ccc@{}}
        \toprule
        Algorithm & Mean (ms) & Std (ms) \\ \midrule
        $K$-$D$ tree    & 20.0      & 0.1      \\
        FastGA (ours)  & 9.1       & 0.1      \\ \bottomrule
        \end{tabular}
    \end{subtable}%
    \begin{subtable}[t]{.5\linewidth}
      \centering
        \caption{Real World: 60,620 Normals }\label{table:results_fastga_b}
        \begin{tabular}{@{}ccc@{}}
        \toprule
        Algorithm     & Mean (ms) & Std (ms) \\ \midrule
        $K$-$D$ tree & 9.7       & 0.2      \\
        FastGA (ours)      & 4.4       & 0.1      \\ \bottomrule
        \end{tabular}
    \end{subtable} 
\end{table}

Peak detection is currently implemented in Python using the \texttt{scikit-image} image processing library \cite{scikit-image}. Agglomerative hierarchical clustering (AHC) of any detected peaks is implemented in Python with the \texttt{scipy} library \cite{2020SciPy-NMeth}. The unwrapped 2D image of the icosahedron does not depend on the number of integrated normals but only the refinement level of the GA. Generated images are rather small (e.g., 90 $\times$ 34 pixels for a level four GA) resulting in very fast peak detection and clustering, e.g., it takes approximately 1 ms to detect peaks and perform AHC on a level four refined GA.  FastGA results from additional datasets are shown below. 

\subsection{Unorganized 3D Point Clouds}\label{sec:results_unorganized}

Sections \ref{sec:results_rooftop} and \ref{sec:results_kitti} describe results from Polylidar3D applied to airborne LiDAR point clouds and point clouds generated on a moving vehicle, respectively.  Both datasets offer real-world unorganized 3D point cloud evaluation of Polylidar3D.

\subsubsection{Rooftop Detection} \label{sec:results_rooftop}

\begin{figure}[t]
  \begin{subfigure}[t]{.33\linewidth}
    \centering\includegraphics[trim=5mm 5mm 5mm 5mm, clip, width=5cm]{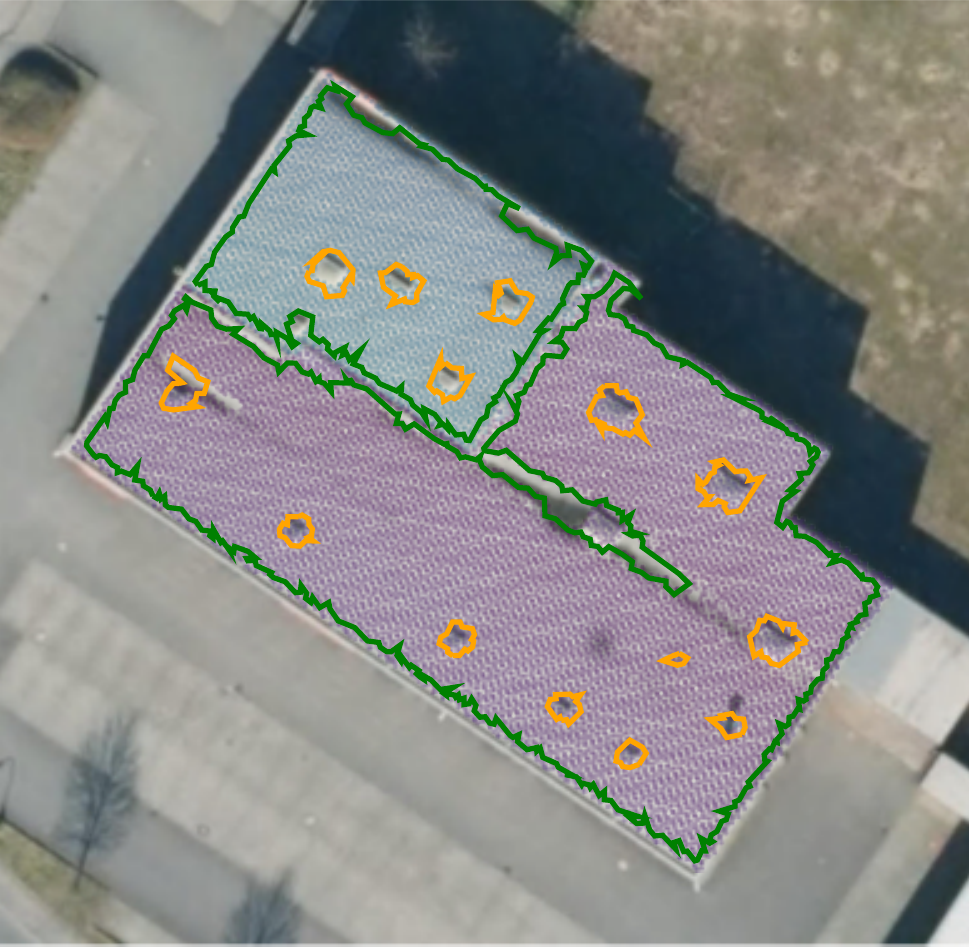}
    \caption{\label{fig:rooftop_a}}
  \end{subfigure}
  \hfill
  \begin{subfigure}[t]{.30\linewidth}
    \centering\includegraphics[trim=15mm 8mm 5mm 5mm, clip, width=4cm]{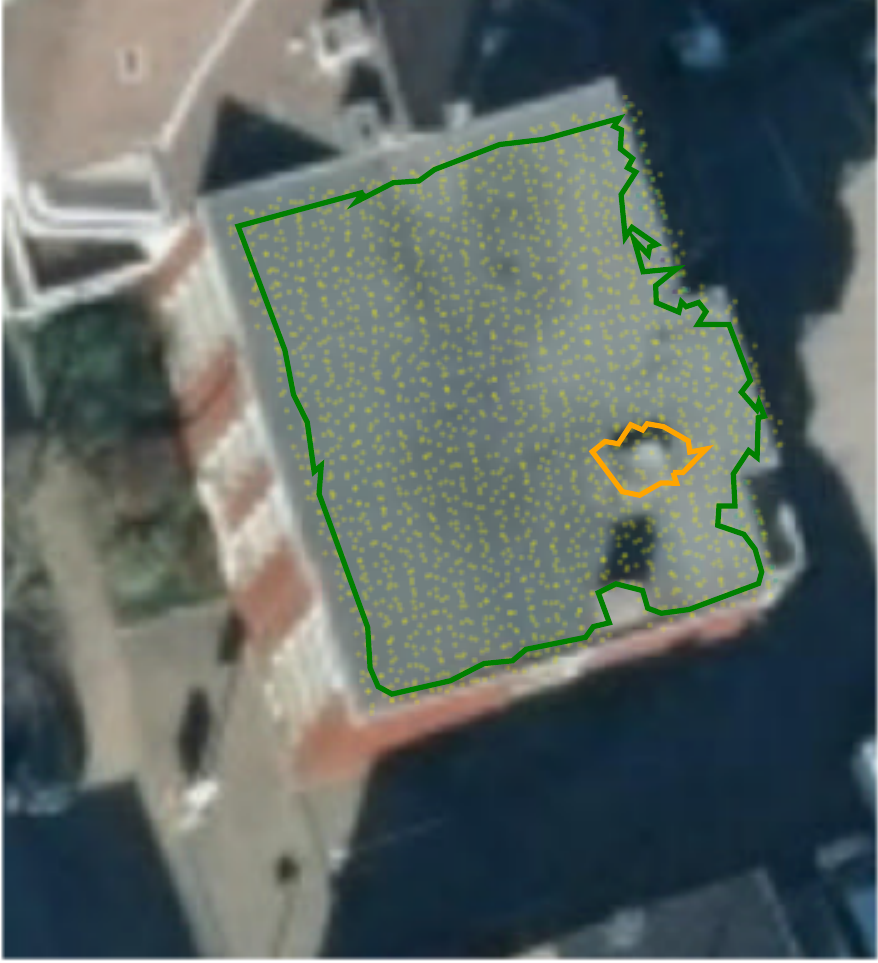}
    \caption{\label{fig:rooftop_b}}
  \end{subfigure}
  \hfill
  \begin{subfigure}[t]{.30\linewidth}
    \centering\includegraphics[trim=5mm 7mm 0mm 7mm, clip, width=4cm]{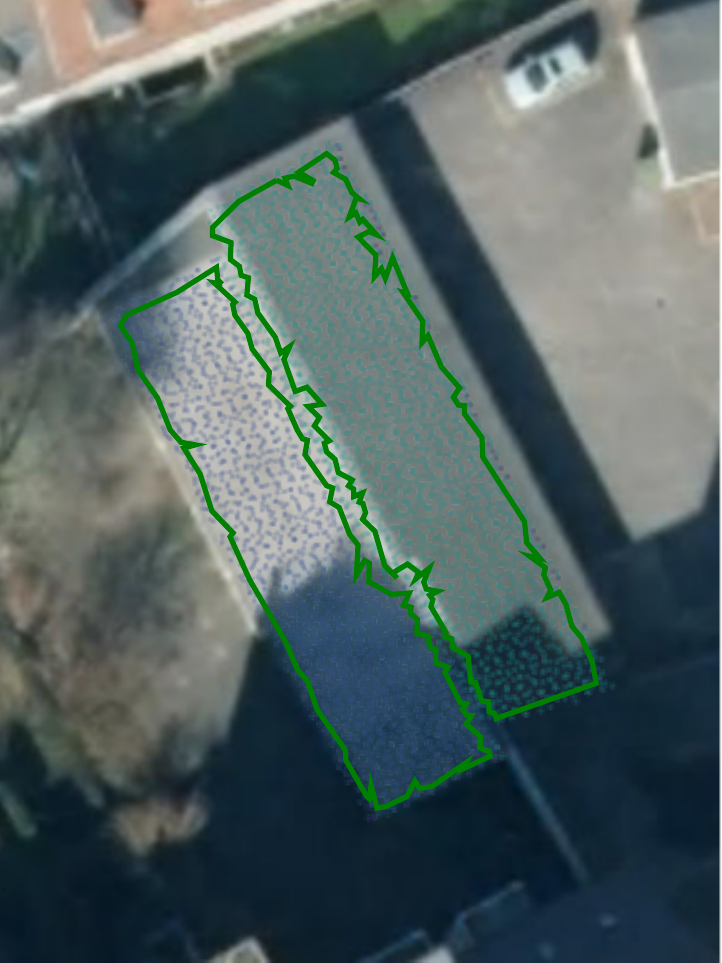}
    \caption{\label{fig:rooftop_c}}
  \end{subfigure}
    \par\bigskip
    \centering
  \begin{subfigure}[t]{.55\linewidth}
    \centering\includegraphics[trim=0mm 0mm 0mm 0mm, clip, width=8cm]{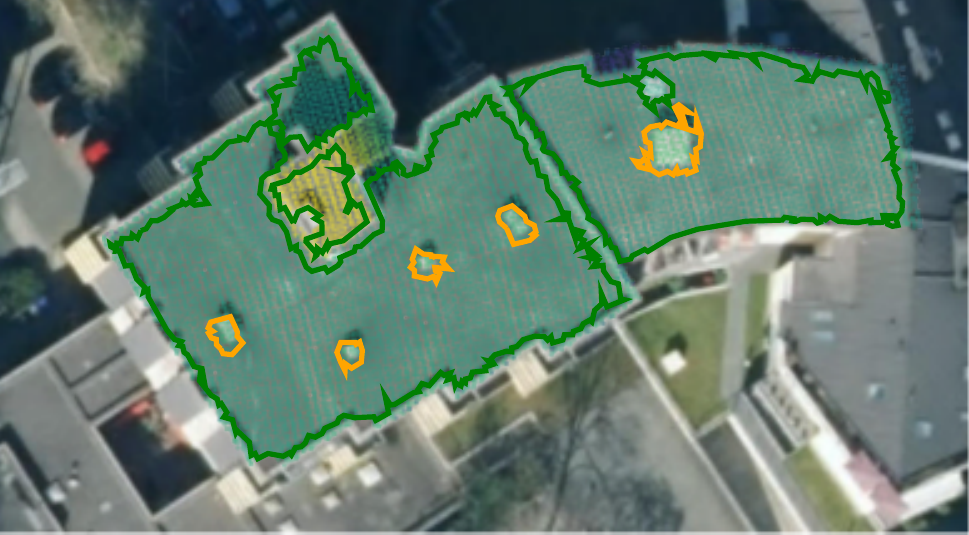}
    \caption{\label{fig:rooftop_d}}
  \end{subfigure}
  \hfill
  \begin{subfigure}[t]{.40\linewidth}
    \centering\includegraphics[trim=0mm 4mm 0mm 0mm, clip, width=6cm]{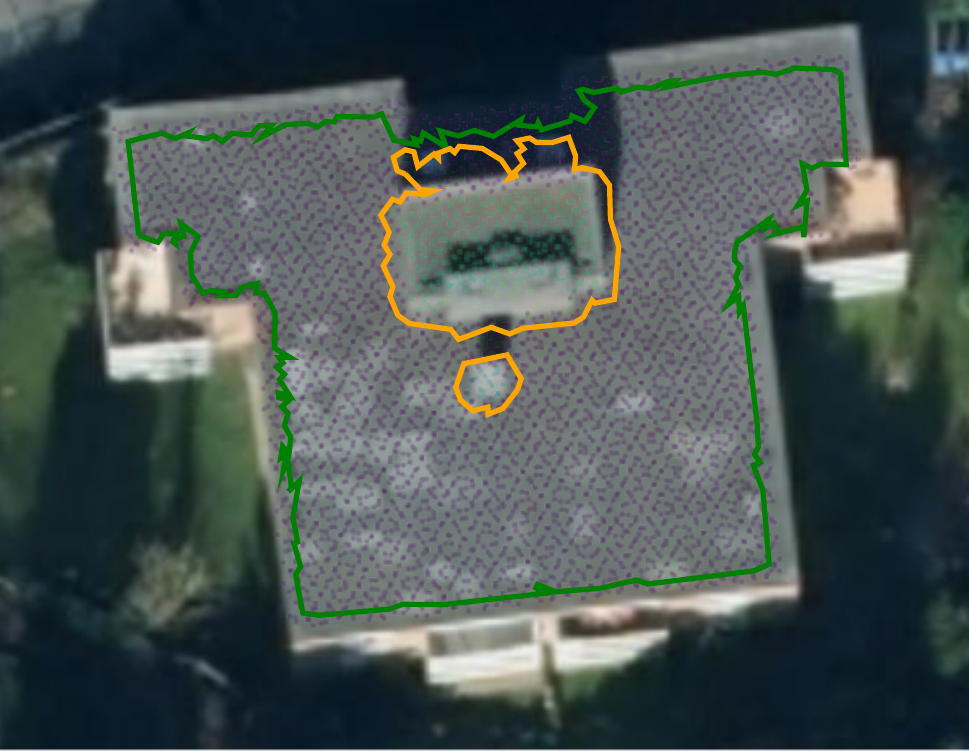}
    \caption{\label{fig:rooftop_e}}
  \end{subfigure}
  \caption{Polygon extraction of rooftops from unorganized 3D point clouds generated by airborne LiDAR.  Each figure shows the satellite image overlaid with the extracted polygons (green) representing flat surfaces with interior holes (orange) representing obstacles. A colorized point cloud is also overlaid ranging from dark purple to bright yellow denoting a normalized low to high elevation. LiDAR data and satellite images are provided from \cite{lidar_germany} and \cite{satellite_germany} respectively. }\label{fig:rooftop}
\end{figure}

This section presents qualitative results of Polylidar3D extracting flat rooftops in cities from airborne LiDAR point clouds of buildings. These unorganized point clouds are captured from an overhead viewpoint but are typically angled based on sensor location; wall surfaces can therefore be visible. Point cloud $xy$ components are in a planar projected coordinate system while the $z$ component represents elevation. Therefore points of a flat rooftop surface are already aligned with the $xy$ plane making them suitable for 2.5D Delaunay triangulation to create a half-edge triangular mesh.  The mesh is smoothed with Laplacian and bilateral filtering using Open3D \cite{zhou2018open3d}. Flat surfaces are then extracted as polygons and any non-flat obstacles on them become holes. Figure \ref{fig:rooftop} shows the extracted polygons of buildings in Witten, Germany. Satellite imagery is overlaid with colorized point clouds. 
Each flat surface is extracted as a polygon with holes. 
All parameters used for this dataset are shown in Table \ref{table:rooftop_parameters}. 

Figure \ref{fig:rooftop_a} shows a single building with two flat surfaces identifiable from the overlaid blue and purple points representing higher and lower elevation respectively.  Polylidar3D successfully separates both of these flat surfaces as two polygons. Rooftop obstacles such as air vents and A/C units are captured as holes. Figure (\subref{fig:rooftop_b}) and (\subref{fig:rooftop_c}) images show additional examples of obstacle detection and surface separation, respectively. The large building on the left in (\subref{fig:rooftop_d}) hosts a structure on top of its own flat rooftop (bright yellow points). This small structure is distinguished, and its own smaller flat rooftop is also extracted. The building in (\subref{fig:rooftop_e}) also has a small rooftop structure captured as a hole in the larger building's rooftop surface, but this structures rooftop is too small to meet the minimum area constraint used during polygon filtering. Note that Polylidar3D failed to extract several small obstacles in (\subref{fig:rooftop_d}) for the building on the right. Such obstacles are too small to be extracted after mesh smoothing. 

\begin{table}[ht]
\centering
\caption{Polylidar3D Parameters for Rooftop Detection.}\label{table:rooftop_parameters}
\begin{tabular}{@{}ll@{}}
\toprule
Algorithm        & Parameters                                                          \\ \midrule
Laplacian Filter & $\lambda$ = 1.0, iterations = 2   \\
Bilateral Filter & $\sigma_l$ = 0.1, $\sigma_a$ = 0.1 , iterations = 2 \\
Plane/Poly Extr.      & $tri_{min}$ = 200, $ang_{min}$ = 0.94, $l_{max}$ = 0.9, $vertices^{hole}_{min}$ = 8, $ptp_{max} = 0.20$     \\
Poly. Filtering      & $\alpha = 0.1$, $\beta_{neg}$ = 0.1, $\beta_{pos}$ = 0.00, $\gamma$ = 16, $\delta$ = 0.5  \\ \bottomrule
\end{tabular}
\end{table}


\subsubsection{Ground and Obstacle Detection}\label{sec:results_kitti}

\begin{figure}[th]

  \begin{subfigure}{.45\linewidth}
    \centering\includegraphics[clip, trim=0.1cm 0cm 0cm 0cm, width=.95\linewidth]{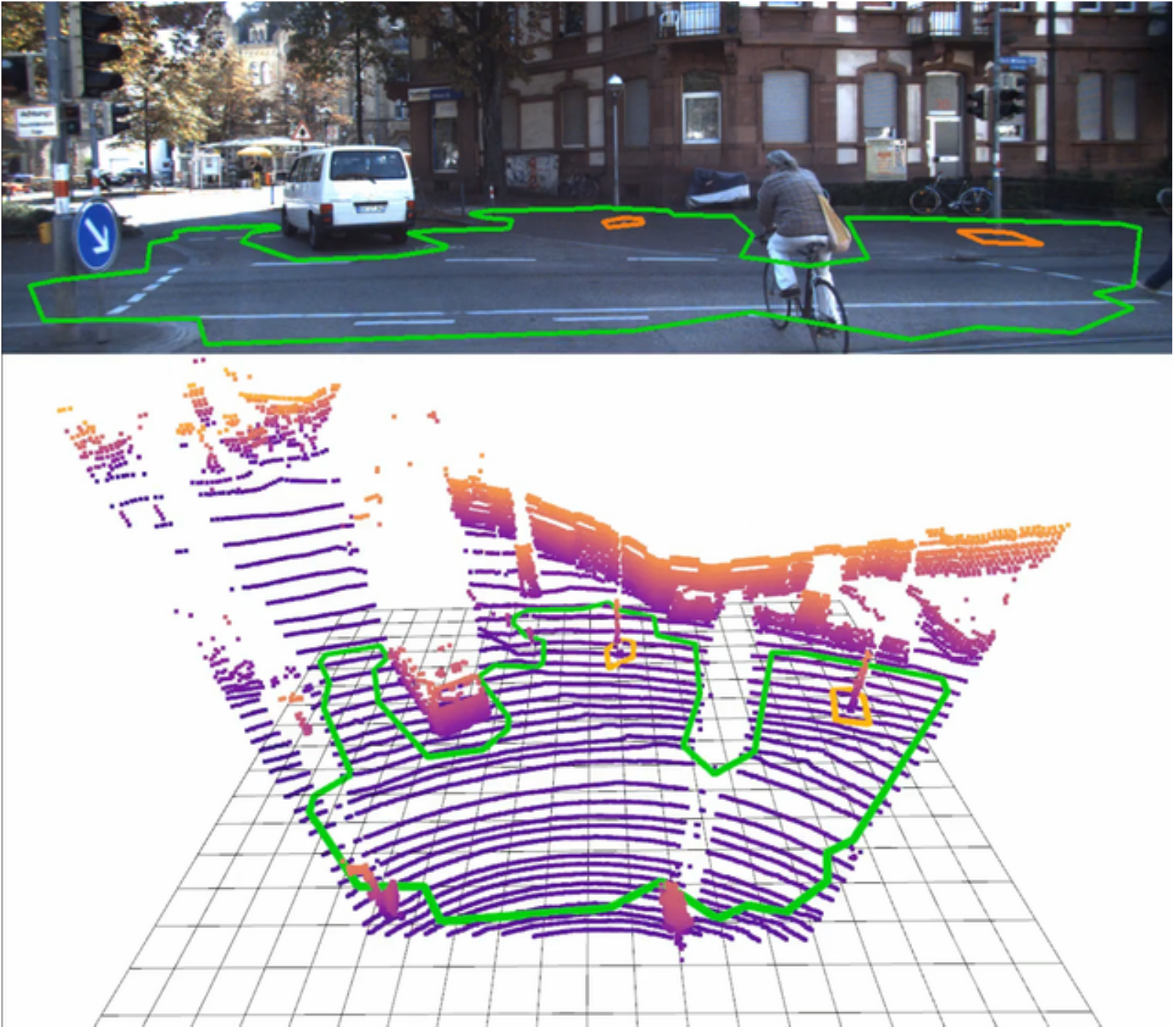}
    \caption{\label{fig:kitti_a}}
  \end{subfigure}
  \hspace{0.1cm}
  \begin{subfigure}{.45\linewidth}
    \centering\includegraphics[clip, trim=0.1cm 0.1cm 0.1cm 0cm, width=.95\linewidth]{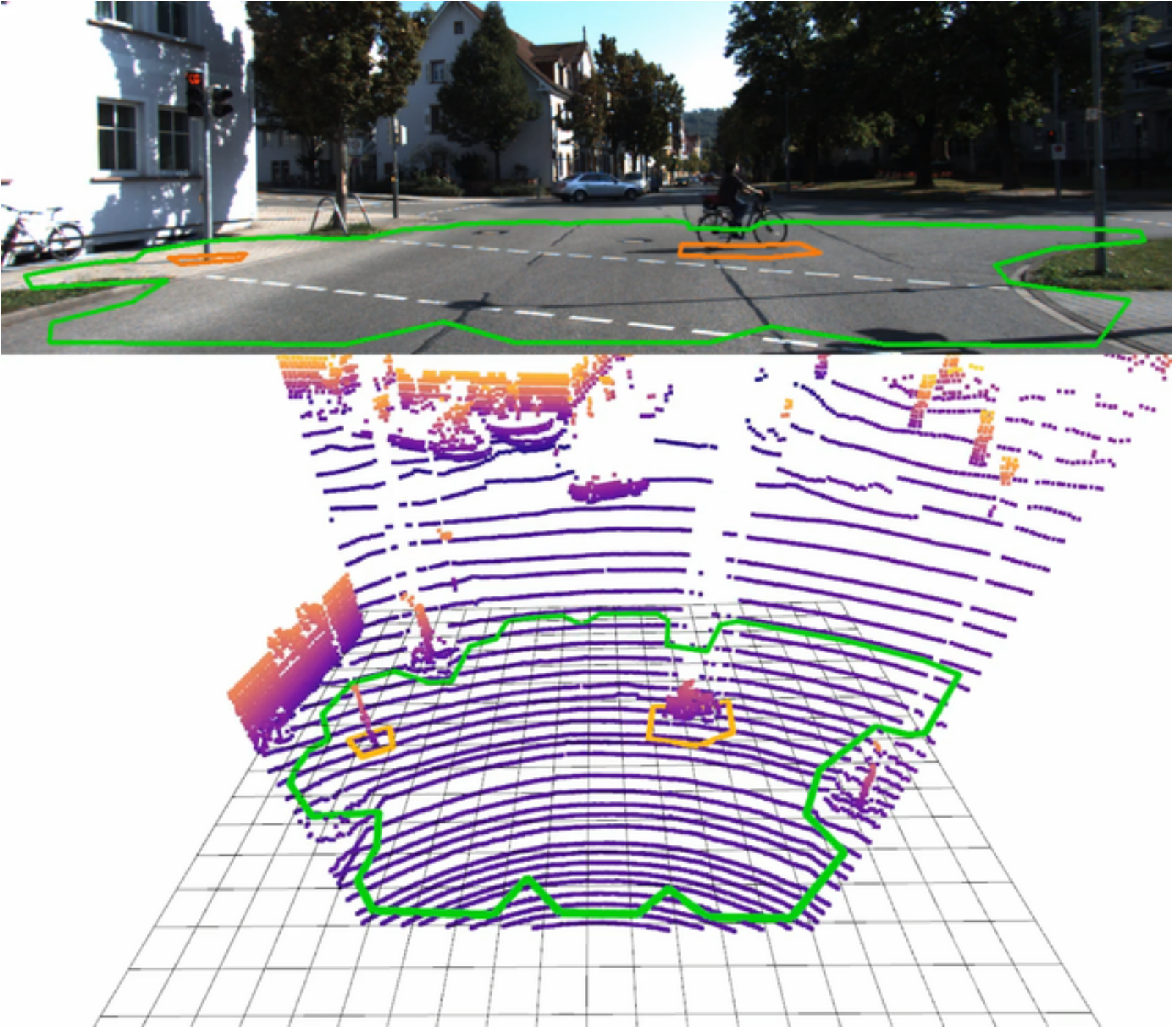}
    \caption{\label{fig:kitti_d}}
  \end{subfigure}
   \par\bigskip
  \begin{subfigure}{.45\linewidth}
    \centering\includegraphics[clip, trim=0.1cm 0cm 0.1cm 0cm, width=.95\linewidth]{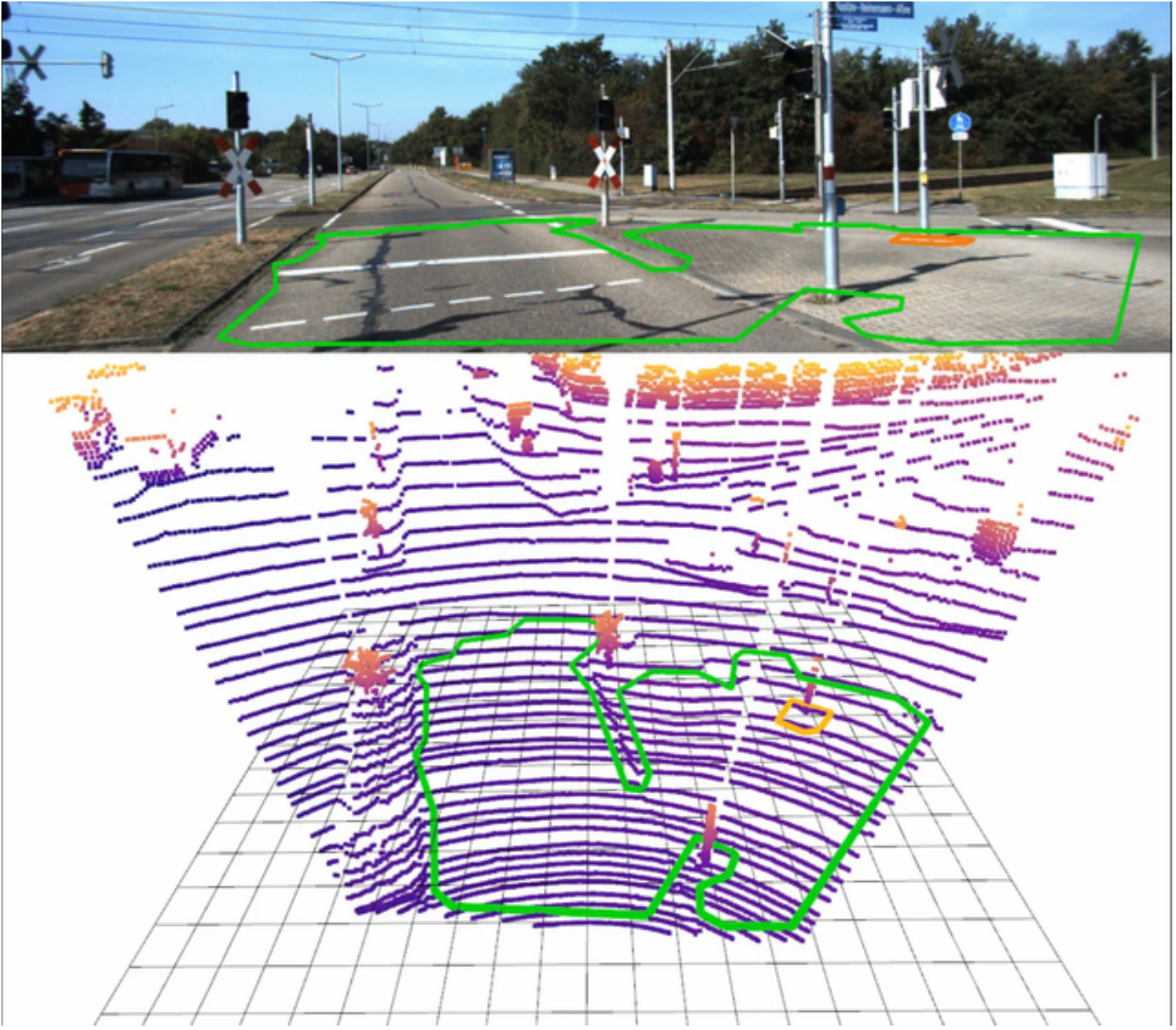}
    \caption{\label{fig:kitti_c}}
  \end{subfigure}
  \hspace{0.1cm}
  \begin{subfigure}{.45\linewidth}
    \centering\includegraphics[clip, trim=0cm 0cm 0.1cm 0.0cm, width=.95\linewidth]{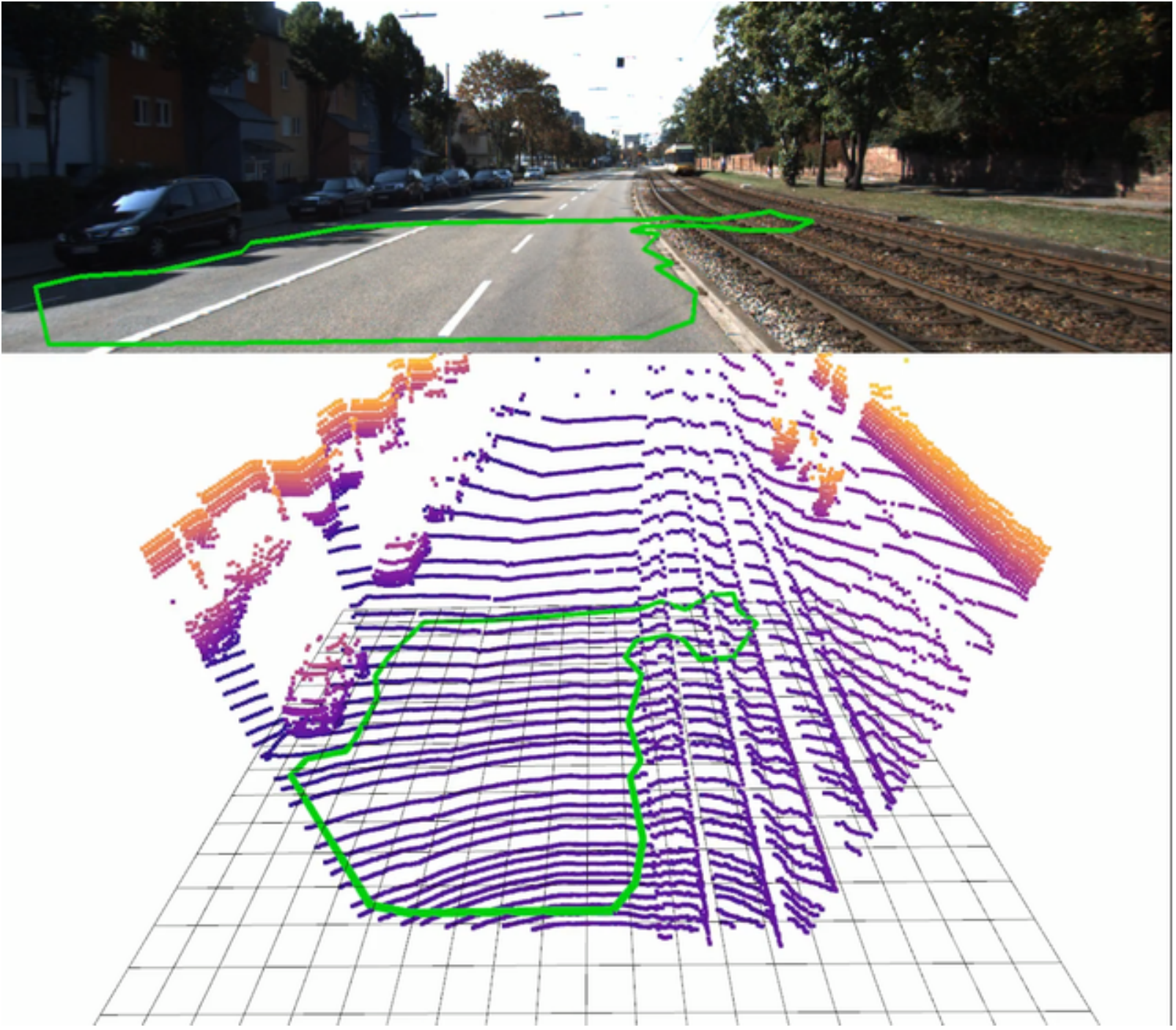}
    \caption{\label{fig:kitti_f}}
  \end{subfigure}
  \caption{Polygon extraction from KITTI unorganized 3D point cloud data acquired on a moving vehicle \cite{Geiger2013IJRR}. Four scenes (\subref{fig:kitti_a}, \subref{fig:kitti_d}, \subref{fig:kitti_d}, \subref{fig:kitti_f}) are shown, each with two subimages. The top subimage shows polygons projected into the color image while the bottom image shows 3D point cloud and surface polygon(s) from a bird's eye view. Obstacles on the ground such as the light and signal in (\subref{fig:kitti_a}) are extracted as (orange) holes. }\label{fig:kitti}
\end{figure}

The KITTI Vision Benchmark Suite provides raw datasets of Velodyne LiDAR point clouds, color video, and calibration data captured from a car while driving in Karlsruhe, Germany \cite{Geiger2013IJRR}.  Calibration data gives fixed transformations between vehicle body frame, Velodyne LiDAR frame, and camera frame. Raw point cloud is projected into the video image, and points outside the image are removed. Next, the point cloud is reduced to half its original size by skipping every odd point index. The filtered point cloud is then transformed to the vehicle body frame.
A single beam/point is deemed an outlier and removed if its left and right neighboring beams are part of a common flat surface and the point strongly deviates from this surface. The filtered point cloud is then sent to the Polylidar3D front-end for 2.5D Delaunay triangulation. Flat connected surfaces on the mesh are extracted as polygons, capturing any obstacles as interior holes. Polygons are filtered and simplified using methods in Section \ref{sec:methods_polylidar_polygon_filtering}. Filtered polygons are then transformed back to the camera frame and projected into the color image for visualization (e.g., top of Figure \ref{fig:kitti_a}). A second image is generated of the 3D point cloud and polygons from a bird's eye viewpoint (e.g., bottom of Figure \ref{fig:kitti_a}). This process is repeated for every frame of 24 distinct ``drives'' (continuous video/LiDAR sequences) provided by KITTI.  Visual qualitative results and execution timings are presented in Section \ref{sec:results_kitti_qual} and \ref{sec:results_kitti_exec} respectively. All code is open source \cite{polylidar_kitti} and the videos of the generated polygons in their entirety can be found here \cite{polylidar3d_kitti_videos}. 

\subsubsection{Qualitative Results}\label{sec:results_kitti_qual}

Roads are not truly flat; they often have elevation changes such as a raised center line for drainage. For this reason we do not use the point-to-plane distance parameter in Polylidar3D to allow flexibility in capturing semi-flat ground surfaces in street environments. Table \ref{table:kitti_parameters} shows the set of parameters used for plane/polygon extraction and filtering with KITTI. Parameter $tri_{min}$ filters out small planes, $ang_{min}$ provides the tolerance for flatness, $l_{max}$ sets the max distance between points, and  $vertices^{hole}_{min}$ filters small holes. The post-processing step of polygon filtering further removes spurious holes and polygons while simplifying polygons for visualization. Note that Polylidar3D is neither designed nor trained specifically to find road surfaces; it is designed to extract flat surfaces as polygons and capture obstacles as holes. The results below thus must not be misconstrued as author intent to apply Polylidar3D for standalone road detection. 

\begin{table}[ht]
\centering
\caption{Polylidar3D Parameters for KITTI.}\label{table:kitti_parameters}
\begin{tabular}{@{}ll@{}}
\toprule
Algorithm        & Parameters                                                          \\ \midrule
Plane/Poly Extr.      & $tri_{min}$ = 3500, $ang_{min}$ = 0.97, $l_{max}$ = 1.25, $vertices^{hole}_{min}$ = 6      \\
Poly. Filtering      & $\alpha = 0.2$, $\beta_{neg}$ = 0.3, $\beta_{pos}$ = 0.02, $\gamma$ = 30, $\delta$ = 0.5      \\ \bottomrule
\end{tabular}
\end{table}

Figure \ref{fig:kitti_a} shows Polylidar3D extracting the road and connected pedestrian walkway as one flat connected surface (green line). A light post and traffic signal are captured as holes because they are in the polygon interior. The cyclist and white vehicle are not captured as holes because they are exterior to the concave hull of the polygon. At greater distances vertical beam spacing becomes greater than $l_{max}$ preventing additional planar surface from being included in the polygon. Figure \ref{fig:kitti_d} shows a scene where a cyclist is explicitly captured as a hole. 
Figure \ref{fig:kitti_c} displays the street and a slightly elevated pedestrian walkway being extracted as one polygon. Surfaces are connected at the smooth wheel chair access transition. This occurs because without a point to plane distance constraint ``flat'' surfaces with similar normals and a smooth spatial connection will be extracted together.  However failures can occur when sensor noise inadvertently dominates a small height change between two surfaces. This is seen in Figure \ref{fig:kitti_f} when the road and part of the adjoining railroad tracks are extracted together. 


\subsubsection{Execution Timings}\label{sec:results_kitti_exec}

Polylidar3D processed 6608 frames from 24 recorded ``drives'' from the raw KITTI dataset. Mean execution timings for each processing step are presented in Table \ref{table:kitti_timings}.  The average size of the point cloud processed by Polylidar3D was 9316 points. No mesh smoothing is performed; the LiDAR is precise and has significant vertical spacing between beams such that the mesh is already sufficiently smooth. Only plane and polygon extraction are run in parallel with a maximum of four threads. The most demanding step is the post processing of polygons through filtering and simplification. 


\begin{table}[ht]
\centering
\caption{Mean Execution Timings (ms) of Polylidar3D on KITTI}
\label{table:kitti_timings}
\begin{tabular}{@{}ccccc@{}}
\toprule
Point Outlier Removal & Mesh Creation & Plane/Poly Ext.  & Polygon Filtering & Total   \\ \midrule
5.1           &    4.1             & 0.7             & 6.8                 & 16.7 \\ \bottomrule
\end{tabular}
\end{table}

\subsection{Organized 3D Point Clouds}\label{sec:results_orgnaized}

Sections \ref{sec:results_rgbd} and \ref{sec:results_synpeb} show Polylidar3D applied to RGBD imagery captured and a benchmark planar segmentation dataset, respectively.  Both datasets are stored as organized 3D point clouds.

\subsubsection{RGBD Cameras}\label{sec:results_rgbd}

We used an Intel RealSense D435i to capture depth and RGB frames in two home environment scenes. The D435i uses stereo infrared cameras to generate a depth map.  Depth noise grows quadratically with distance, and empirical evidence indicates as much as four centimeter RMS error at a two meter distance \cite{realsense_noise}. The Intel RealSense SDK provides denoising post-processing filters including decimation (downsampling), spatial bilateral smoothing, temporal filtering, and depth thresholding \cite{intel_realsense}. Parameters used for each of these filters are shown in Table \ref{table:realsense_parameters}, and Polylidar3D parameters used for captured RGBD data are shown in Table \ref{table:rgbd_parameters}. Each sensor is sampled at 424 $\times$ 240 resolution in a well-lit indoor environment shielded from direct sunlight. Raw data is recorded to assure all qualitative and quantitative results can be reproduced \cite{polylidar_realsense}.  

\begin{table}[ht]
\centering
\caption{Intel RealSense SDK Post-processing Filter Parameters.}\label{table:realsense_parameters}
\begin{tabular}{@{}ll@{}}
\toprule
Algorithm        & Parameters                                                          \\ \midrule
Decimation     & magnitude = 2   \\
Temporal      & $\alpha$ = 0.3, $\delta$ = 60.0, persistence = 2 \\
Spatial       & $\alpha$ = 0.35, $\delta$ = 8.0, magnitude = 2, hole fill = 1  \\
Threshold     & max distance = 2.5 m  \\ \bottomrule
\end{tabular}
\end{table}

\begin{table}[ht]
\centering
\caption{Polylidar3D Parameters for RealSense RGBD}\label{table:rgbd_parameters}
\begin{tabular}{@{}ll@{}}
\toprule
Algorithm        & Parameters                                                          \\ \midrule
Laplacian Filter & $\lambda$ = 1.0, kernel size = 3, iterations = 2   \\
Bilateral Filter & $\sigma_l$ = 0.1, $\sigma_a$ = 0.15, kernel size = 3, iterations = 2 \\
FastGA           & level = 3,  $v_{min} = 50, d_{peak} = 0.28, sample_{pct} = 12\%$            \\
Plane/Poly Extr.      & $tri_{min}$ = 500, $ang_{min}$ = 0.96, $l_{max}$ = 0.05, $ptp_{max} = 0.1$, $vertices^{hole}_{min}$ = 10  \\
Poly. Filtering      & $\alpha = 0.02$ , $\beta_{neg}$ = 0.02, $\beta_{pos}$ = 0.005, $\gamma$ = 0.1, $\delta$ = 0.1      \\ \bottomrule
\end{tabular}
\end{table}

\begin{figure}[ht]

  \begin{subfigure}[t]{.48\linewidth}
    \centering
    \includegraphics[width=.99\linewidth]{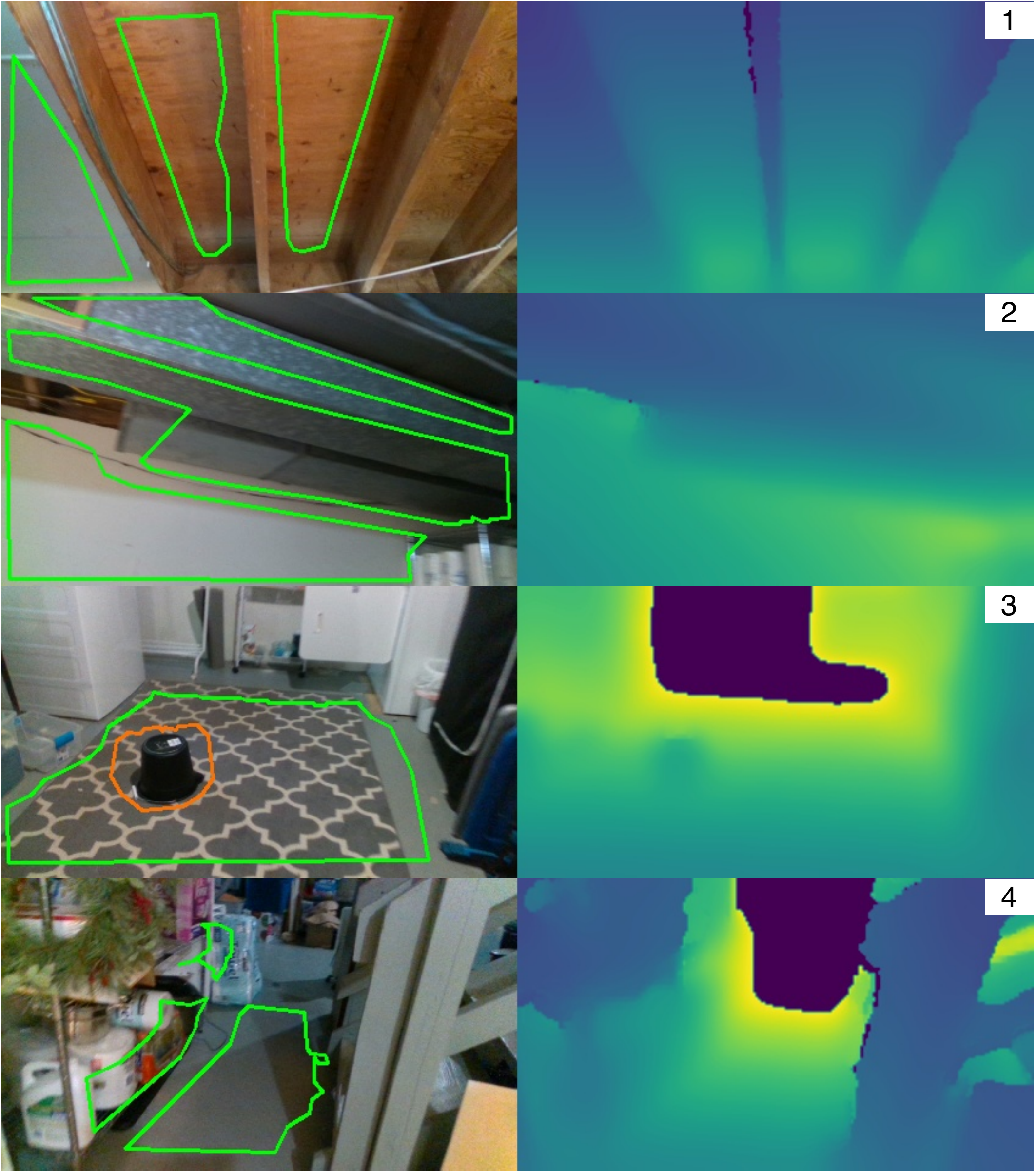}
    \caption{Basement}
    \label{fig:rgbd_basement}
  \end{subfigure}
  \hfill
  \begin{subfigure}[t]{.48\linewidth}
    \centering
    \includegraphics[width=.99\linewidth]{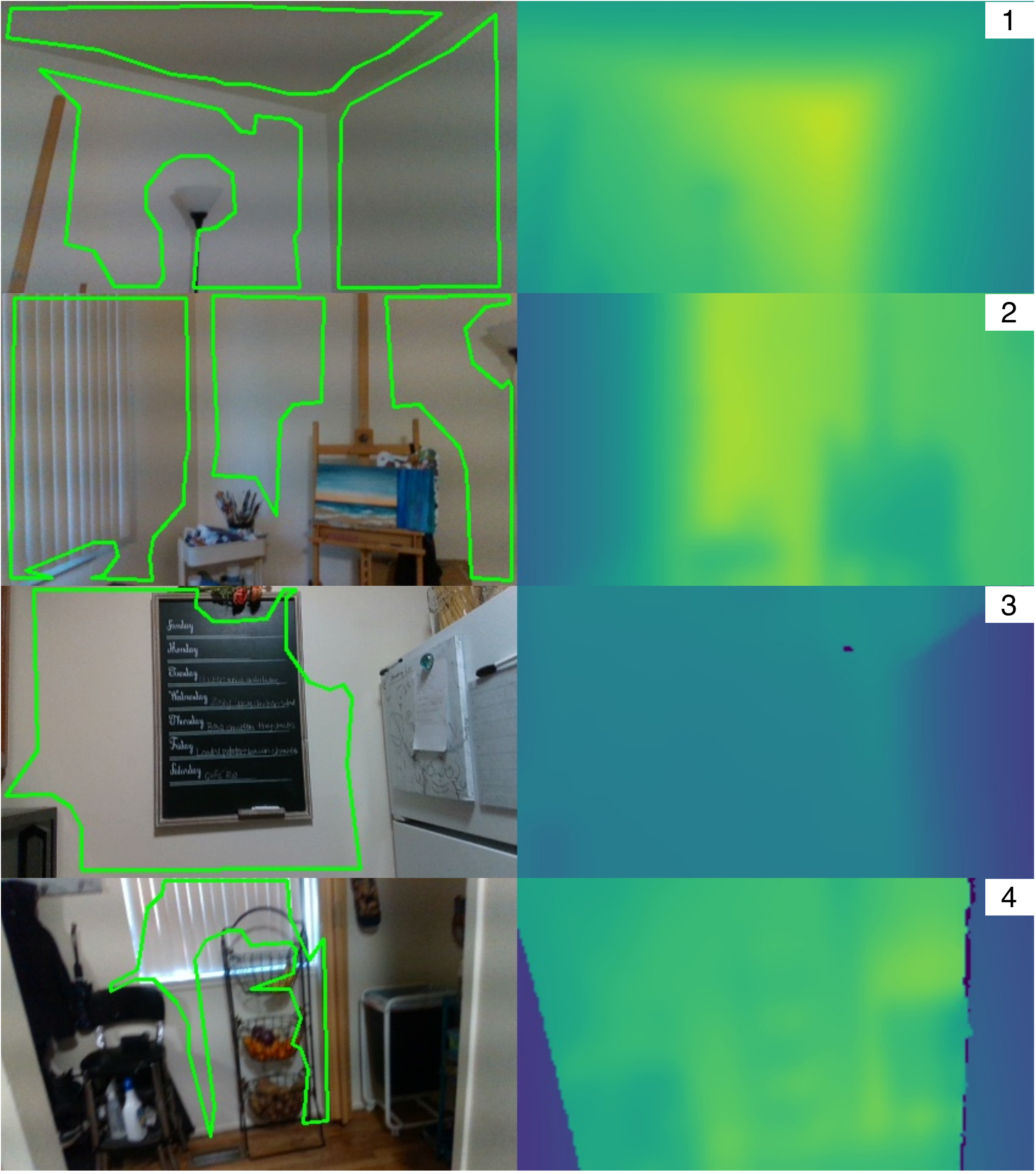}
    \caption{Main Floor}
    \label{fig:rgbd_main_floor}
  \end{subfigure}

  \caption{Real-time polygon extraction from an RGBD camera. Color and depth frames are shown side by side. Polygons are projected onto the color image. Green denotes the exterior hull, and orange denotes any interior holes in the polygon.}\label{fig:rgbd}
\end{figure}

The first scene is composed of 2246 frames (74 seconds) with the camera traversing from one side of a basement to the other. While walking the camera is pointed in many directions including the floor, ceiling, and walls.  Multiple dominant planar surfaces are captured at the same time.  Small surfaces are explicitly removed by filtering planes and polygons that do not meet minimum number of triangles and area constraints. Figure \ref{fig:rgbd_basement} shows several image pairs for this scene. The left image is the RGB video with overlaid 3D polygon projections; the right image is the associated filtered and colorized depth map. Image (1) shows three planar segments with a common surface normal being extracted from the ceiling. Image (2) shows three planar segments with different normals being extracted. Image (3) shows the floor being extracted with a hole representing a bucket obstacle. The last bottom image shows Polylidar3D incorrectly capturing items on a shelf as polygons which erroneously appear as ``planar'' surfaces. This occurs because the small gaps between the items on the shelf become smoothed and appear planar after RealSense post-processing of the depth image. 

The second scene is composed of 2735 frames (91 seconds) with the camera moving on the main floor through dining area, kitchen, and living room. The camera is pointed in many directions and with many dominant planar surfaces extracted as polygons. Figure \ref{fig:rgbd_main_floor} shows several still images captured. The first and second images show Polylidar3D capturing three planar segments on walls and a ceiling. A lamp and an art stand break up extracted planar segments with the polygons forming around them. For the third image, Polylidar3D does not distinguish the wall from the chalkboard because the depth difference is too small after the RealSense post-processing filters are applied.  The fourth image shows Polylidar3D extracting curtains and a lower portion of a wall as one flat surface which surrounds the fruit basket. The floor is not captured because its resulting polygon did not mean minimum area constraints. The adjacent wall to the left did not meet planarity constraints with the window forming a separate segment which in itself was too small and filtered. 

Table \ref{table:rgbd_timings} summarizes mean execution timings for each step of Polylidar3D over all frames of each respective RGBD scene. Polygon filtering is most computationally demanding and includes polygon buffering and simplification routines.  Image and mesh filtering (RealSense SDK and our mesh filtering) are extremely fast because they take advantage of the organized point cloud data structure. Dominant plane normal estimation with FastGA is quick and effective. Planar segmentation and polygon extraction are both completed in less than 2ms. Note that GPU acceleration is used for Laplacian and bilateral filtering, and four threads are used for plane/polygon extraction. All other steps are single-threaded.

\begin{table}[ht]
\centering
\caption{Mean execution timings (ms) of Polylidar3D with RGBD data collected from indoor settings.}
\label{table:rgbd_timings}
\begin{tabular}{@{}lllllllll@{}}
\toprule
Scene & RS Filters & Mesh & Laplacian & Bilateral & FastGA & Plane/Poly Ext.  & Poly. Filt.  & Total  \\ \midrule
Basement         &  2.4   &  0.4           & 0.4       & 0.5       & 1.2    & 1.7             & 4.8              & 11.4      \\
Main Floor       &  2.4   &  0.4           & 0.4       & 0.5       & 1.3    & 1.6             & 5.1              & 11.7       \\ \bottomrule
\end{tabular}
\end{table}



\subsubsection{SynPEB Benchmark}\label{sec:results_synpeb}

\begin{figure}[ht]

  \begin{subfigure}[t]{.45\linewidth}
    \centering\includegraphics[width=.95\linewidth]{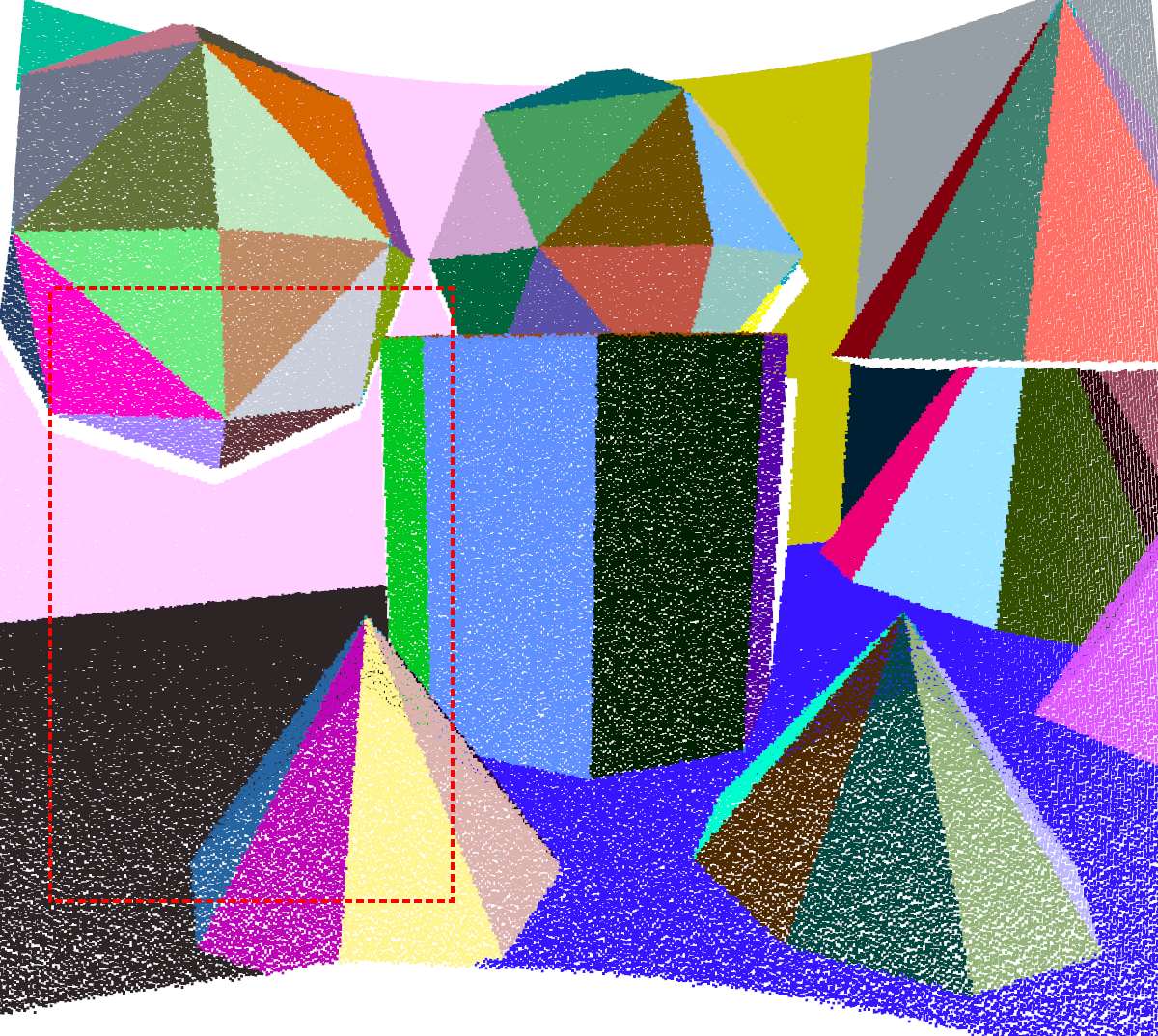}
    \caption{\label{fig:synpeb_a}SynPEB Organized Point Cloud}
  \end{subfigure}
  \hfill
  \begin{subfigure}[t]{.23\linewidth}
    \centering\includegraphics[width=.95\linewidth]{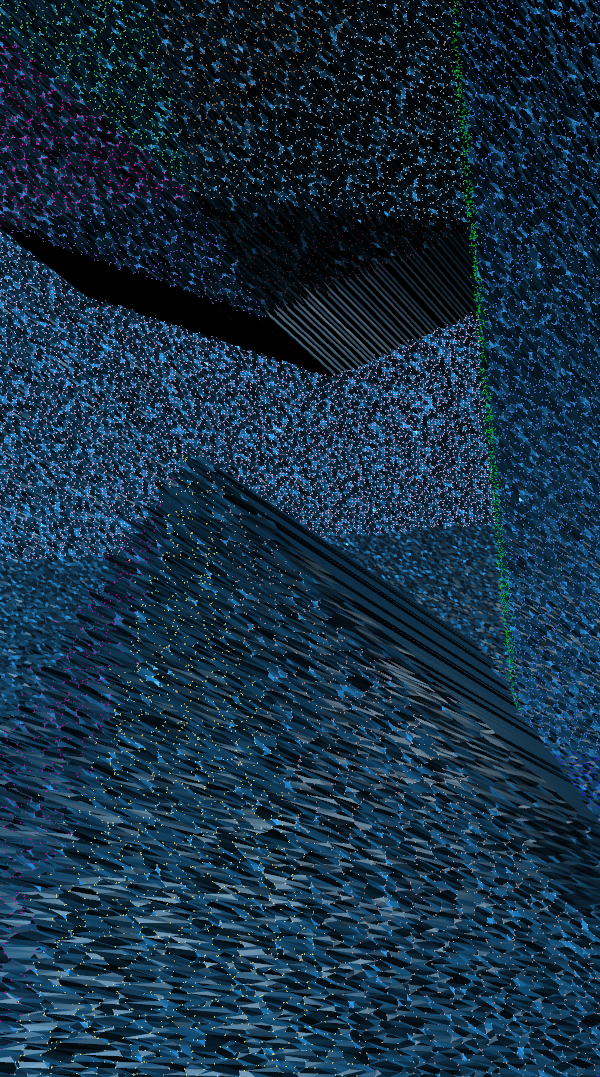}
    \caption{\label{fig:synpeb_b}Generated Mesh}
  \end{subfigure}
  \begin{subfigure}[t]{.23\linewidth}
    \centering\includegraphics[width=.95\linewidth]{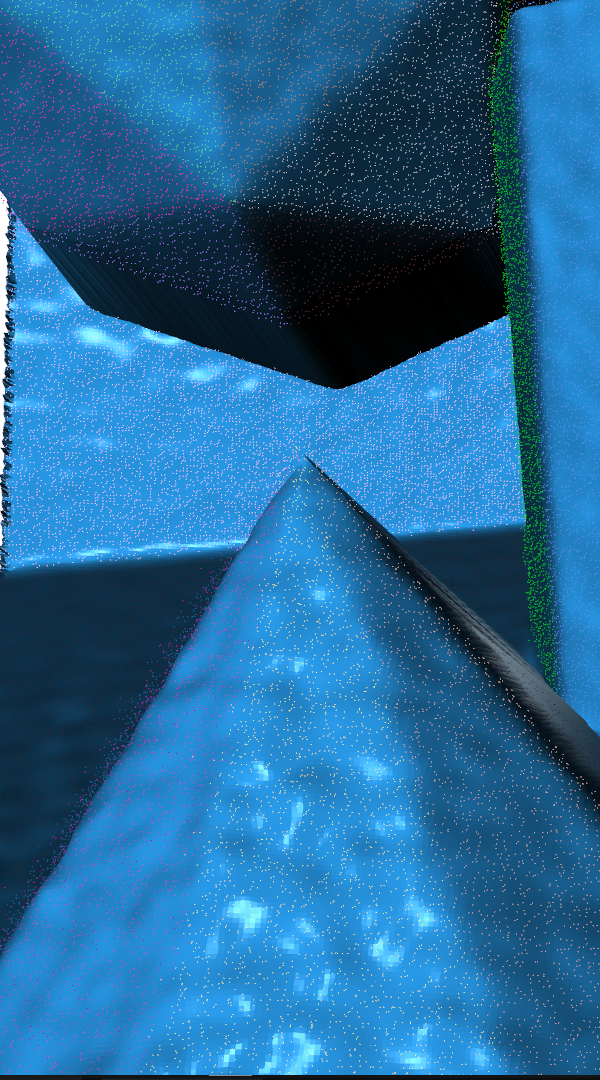}
    \caption{\label{fig:synpeb_c}Mesh Smoothing}
  \end{subfigure}
  \par\bigskip
    \centering
  \begin{subfigure}[t]{.45\linewidth}
    \centering\includegraphics[width=.95\linewidth]{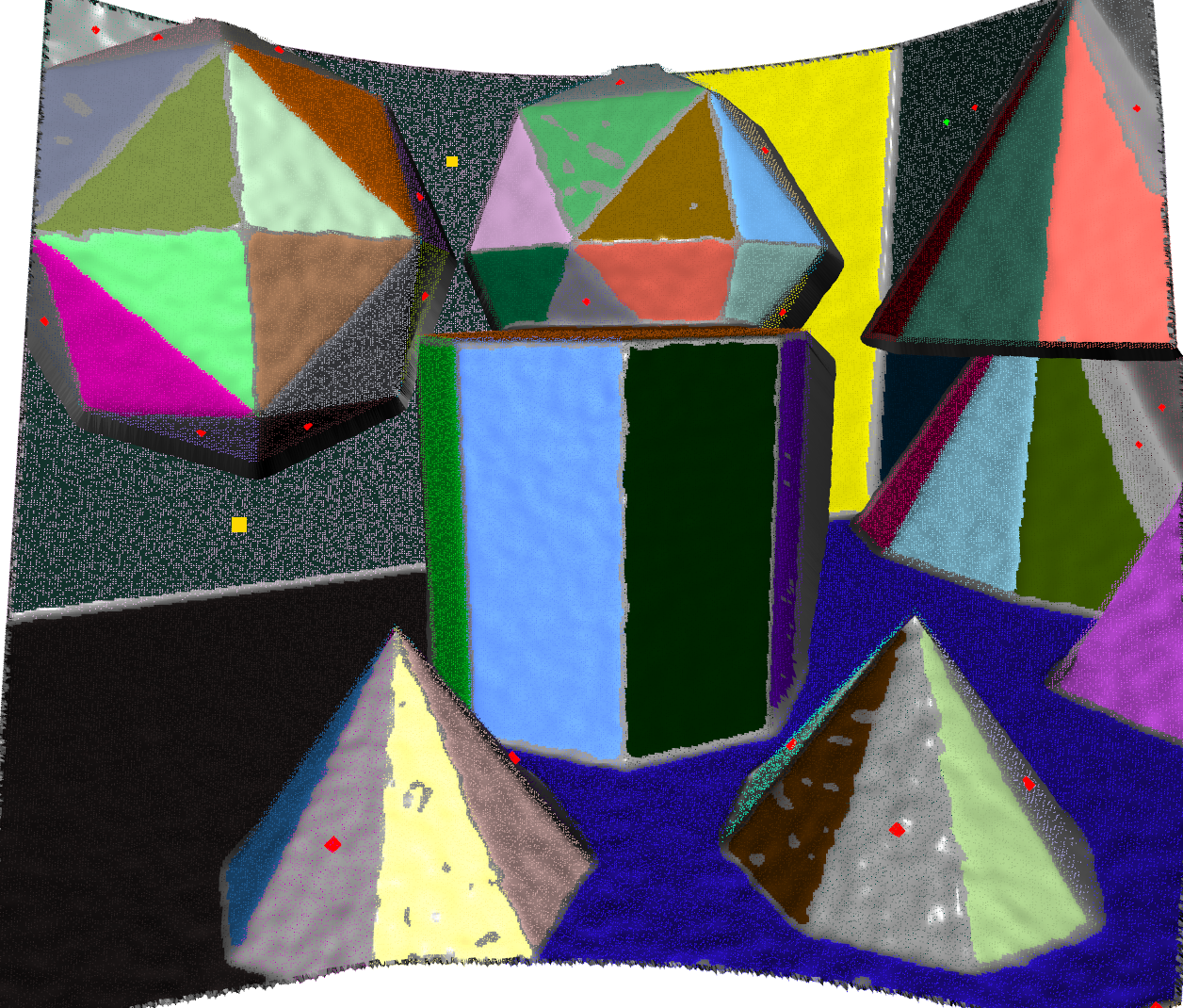}
    \caption{\label{fig:synpeb_d}Polylidar3D Generated Planes}
  \end{subfigure}
  \hspace{1cm}
  \begin{subfigure}[t]{.45\linewidth}
    \centering\includegraphics[width=.95\linewidth]{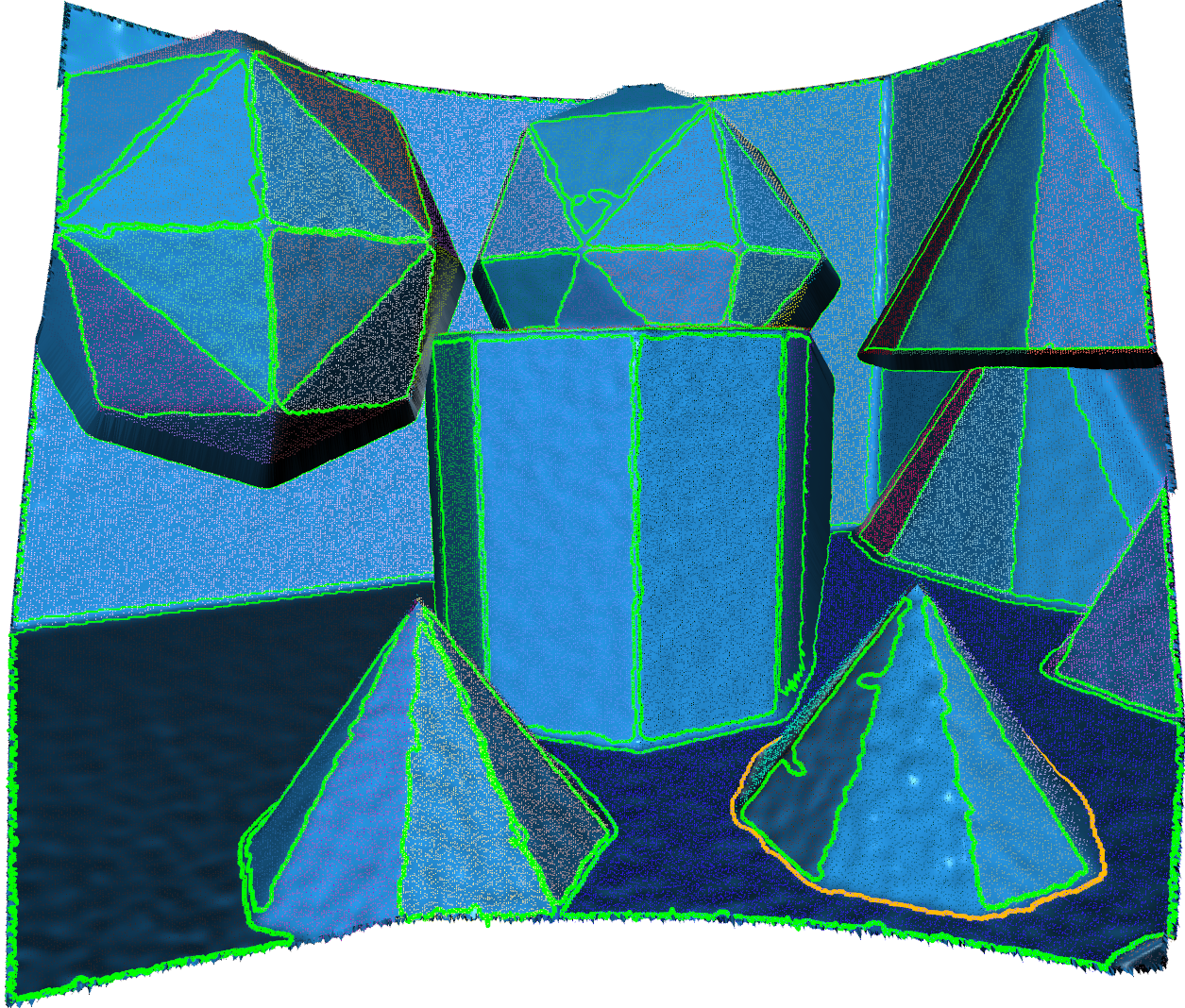}
    \caption{\label{fig:synpeb_e}Polylidar3D Generated Polygons}
  \end{subfigure}
  \caption{Example SynPEB scene with the highest noise level. Point cloud, generated mesh, and mesh  smoothed through Laplacian and bilateral filtering are shown in (\subref{fig:synpeb_a}), (\subref{fig:synpeb_b}), and (\subref{fig:synpeb_c}), respectively.  Planes and polygons generated by Polylidar3D are shown in (\subref{fig:synpeb_d}) and (\subref{fig:synpeb_e}). Red, green, and yellow blocks in (\subref{fig:synpeb_d}) represent missed, spurious, and oversegmented planes.}\label{fig:synpeb_pics}
\end{figure}

We also evaluated Polylidar3D on SynPEB, a challenging benchmark dataset used to evaluate plane segmentation algorithms, created by the authors of PPE  \cite{schaefer_maximum_2019}. This synthetic dataset is generated from a room populated with various polyhedra resulting with an average of 42.6 planes. LiDAR scans are simulated with different levels of normally distributed radial and tangential noise producing organized point clouds. There are four levels of tangential noise in the dataset with 0.5 mdeg, 1 mdeg, 2 mdeg, and 4 mdeg standard deviation.  Data is partitioned into a training set to tune algorithms parameters and a test set for evaluation. The combination of high-noise data and numerous small, connected, but distinct planes results in challenges for plane segmentation as shown in Figure \ref{fig:synpeb_pics}. 
The illustrated example uses the highest noise level (4 mdeg tangential standard deviation) from the benchmark set.

We used the training set to tune our methods parameters including mesh smoothing (Laplacian and bilateral filter), dominant plane normal estimation (FastGA), and plane/polygon extraction. We found that the most important parameter was the number of iterations of Laplacian smoothing needed. We trained a linear regression model to predict the most suitable number of iterations given an estimate of point cloud noise. All parameters used for test set reproduction are shown in Table \ref{table:synpeb_parameters}. Note that the significant number of noisy distinct planes, up to 72, required a higher than expected refinement level for FastGA and an increased focus on smoothing.

\begin{table}[ht]
\centering
\caption{Polylidar3D parameters for the SynPEB benchmark test set.}\label{table:synpeb_parameters}
\begin{tabular}{@{}ll@{}}
\toprule
Algorithm        & Parameters                                                          \\ \midrule
Laplacian Filter & $\lambda$ = 1.0, kernel size = 5, iterations = varies (predicted)   \\
Bilateral Filter & $\sigma_l$ = 0.1, $\sigma_a$ = 0.1 , kernel size = 3, iterations = 2 \\
FastGA           & level = 5,  $v_{min} = 2, d_{peak} = 0.1, sample_{pct} = 12\%$            \\
Plane/Poly Extr.      & $tri_{min}$ = 1000, $ang_{min}$ = 0.95, $l_{max}$ = 0.1, $ptp_{max} = 0.07$, $vertices^{hole}_{min}$ = 10     \\ \bottomrule
\end{tabular}
\end{table}

Table \ref{table:synpeb_results} shows benchmark test results of Polylidar3D against other plane segmentation methods. The results of other methods including timings are provided in \cite{schaefer_maximum_2019}.  Note that execution times cannot be directly compared but will give an idea of real-time capability. Polylidar3D produces both a point set and polygonal representation of identified planes, however this benchmark must be evaluated by the point set.   A ``plane'' is considered correctly identified if its point set overlaps with the ground truth plane with the standard 80\% threshold described in \cite{hoover_experimental_1996}. Key metrics are $f$ representing the percent of ground truth planes identified, $k$ indicating percent of the point cloud correctly identified, and RMSE quantifying accuracy of each plane fit. Variables $n_o$, $n_u$, $n_m$, and $n_s$ represent the absolute numbers of  oversegmented,  undersegmented,  missing,  and spurious  planes, respectively,  compared  to  the  ground-truth  segmentation. See \cite{schaefer_maximum_2019} and \cite{hoover_experimental_1996} for detailed definitions of these metrics. 
An $f$ metric of 48.6\% indicates that Polylidar3D did not capture most of the planes in the benchmark, however the $k$ metric of 82.0\% indicates our algorithm did well in capturing the large dominant planes comprising most of the point cloud.  Additionally there are fewer spurious, over segmented, and under segmented planes generated by Polylidar3D than with other methods. The RMSE value is also the lowest, indicating the predicated planes have a good fit. Plane segmentation is accomplished in significantly less time, especially in comparison to the front runner PPE. PPE's $f$ and $k$ metrics indicate it does an excellent job of capturing the numerous small planes in the scene, but it fails more often in capturing the large dominant planes.  Polylidar3D uniquely generates concave polygons which provide a condensed representation of identified planes.

\begin{table}[ht]
\centering
\caption{SynPEB Benchmark Results.  }
\label{table:synpeb_results}
\begin{tabular}{@{}ccccccccc@{}}
\toprule
Method                                  & $f$ {[}\%{]} & $k$ {[}\%{]} & RMSE {[}mm{]} & $n_o$ & $n_u$ & $n_m$ & $n_s$ & time\\ \midrule
PEAC \cite{feng_fast_2014}              & 29.1         & 60.4         & 28.6          & 0.7   & 1.0   & 26.7  & 7.4   & 33 ms\\
MSAC \cite{torr_mlesac_2000}            & 7.3          & 35.6         & 34.3          & 0.3   & 1.0   & 36.3  & 10.9  & 1.1 s\\
PPE   \cite{schaefer_maximum_2019}      & 73.6         & 77.9         & 14.5          & 1.5   & 1.1   & 7.1   & 16.5  & 1.6 hr\\
Polylidar3D (proposed)                  & 48.6         & 82.0         & 9.2           & 0.1   & 0.3   & 22.7  & 5.1  & 35 ms\\ \bottomrule
\end{tabular}
\end{table}



Table \ref{table:synpeb_timings} shows mean execution timings for each Polylidar3D method applied to SynPEB.  Each organized point cloud is 500X500 but can be efficiently downsampled by striding over rows and columns. This reduces computational complexity at the cost of reduced accuracy. Note that GPU acceleration is used for both Laplacian and bilateral filtering, while plane/polygon extraction is parallelized up to four CPU cores; all other algorithm steps are single threaded. 

\begin{table}[ht]
\centering
\caption{Mean Execution Timings (ms) and Accuracy of Polylidar3D on SynPEB.}
\label{table:synpeb_timings}
\begin{tabular}{@{}ccccccccc@{}}
\toprule
Point Cloud & Mesh Creation & Laplacian & Bilateral & FastGA & Plane/Poly Ext.  & Total  & $f$ {[}\%{]}  & $k$ {[}\%{]} \\ \midrule
500 X 500        & 9.3           & 1.2       & 3.1       & 6.6    & 14.9             & 35.1         & 48.6   & 82.0       \\
250 X 250        & 2.1           & 0.5       & 0.7       & 2.5    & 4.3             & 10.1          & 41.6    & 74.2      \\ \bottomrule
\end{tabular}
\end{table}

\subsubsection{Organized Point Cloud Mesh Smoothing}

This section provides execution timing analysis of our accelerated mesh smoothing algorithms on organized point clouds (OPC) per Section \ref{sec:methods_mesh_smoothing}. Laplacian and bilateral filtering are tested on two organized point clouds; one from a random scene in SynPEB and another random frame from our RGBD dataset.  Execution timing is most influenced by point cloud size which varies substantially for these two examples. For example the SynPEB OPC has $499 \cdot 499 \cdot 2 = 498,002$ triangles whereas the RGBD frame has at most 50,218 triangles. For each filter we report CPU single-threaded, CPU multi-threaded, and GPU accelerated timings. Only four threads are used in multi-threaded runs, and a kernel size of three is used in all runs. We compare our filters with Open3D's general purpose triangle mesh Laplacian filter \cite{zhou2018open3d}. Note that Open3D uses a general filter and does not take into account the organized structure of the mesh and must therefore create an adjacency list for each vertex to deterimine neighbors. Additional overhead occurs by returning a new triangle mesh whereas our Laplacian implementation returns only the smoothed vertices. Open3D does not have a bilateral filter implementation nor is its Laplacian filter CPU parallelized or GPU accelerated. The smoothed meshes produced by Open3D and ours are nearly the same except for a noisy one pixel border on the boundary of our mesh.  

Table \ref{table:results_lapalcian} shows the results of our Laplacian filter for one and five iterations with results separated by a semicolon. Our CPU single-threaded performance is faster than Open3D. This is mostly explained by not needing to compute a vertex adjacency list. Our CPU multi-threaded results nearly reach the ideal 4X speedup in most scenarios. GPU acceleration is quite fast but has substantial overhead on the first iteration of smoothing. This is because the OPC in CPU memory must be transferred to GPU memory which is an expensive operation. This penalty is only paid once no matter how many iterations of smoothing occur.  
\begin{table}[ht]
\centering
\caption{Execution timing (ms) for one and five iterations of Laplacian filtering.}\label{table:results_lapalcian}

\begin{tabular}{@{}ccccc@{}}
\toprule
                & \multicolumn{3}{c}{Ours} & Open3D \\
                  \cmidrule(lr){2-4}                  
                    \cmidrule(lr){5-5}
Data \& Size    & CPU-S     & CPU-M     & GPU         & CPU-S  \\ \midrule
SynPEB, 500X500 & 7.0; 35.0  & 2.0; 9.2   & 0.8; 0.9    & 205.9; 240.6  \\
RGBD, 120X212   & 0.7; 3.5  & 0.2; 0.9   & 0.1; 0.2    & 14.3; 17.5   \\ \bottomrule
\end{tabular}
\end{table}

Table \ref{table:results_bilateral} shows execution timing results of our bilateral filter for one and five iterations with results separated by a semicolon. This filter is substantially slower than the Laplacian filter primarily because both triangle normals and their centroids must be computed before this filter can run (included in timing). Additionally each of these data structures is nearly twice as large in memory as the input OPC ($\approx$ 2 triangles per vertex). The weighting of neighbors in Equations \ref{eq:bilateral_centroid} and \ref{eq:bilateral_normal} relies on an exponential function which is significantly slower than the floating point multiplication and division required for Laplacian filter per Equation \ref{eq:laplacian_weight}. Finally significantly more neighbors and data are used in Equation \ref{eq:bilateral} to produce a smoothed normal. A maximum of 16 triangle neighbors (accessing both their normals and centroids) are used for the bilateral filter whereas the Laplacian uses a max of eight vertex neighbors. The memory transfer from CPU to GPU is significantly higher as well because 4X as much memory is needed.


\begin{table}[ht]
\centering
\caption{Execution timing (ms) for one and five iterations of bilateral filtering.}\label{table:results_bilateral}
\begin{tabular}{@{}cccc@{}}
\toprule
Data \& Size    & CPU-S       & CPU-M           & GPU      \\ \midrule
SynPEB, 500X500 & 73.0; 354.0 & 19.4; 90.9      & 3.2; 4.4 \\
RGBD, 120X212   & 7.2; 35.0   & 1.9; 9.2        & 0.5; 0.5 \\ \bottomrule
\end{tabular}
\end{table}


\subsection{User-defined Meshes}\label{sec:results_meshes_parallel}

We apply Polylidar3D on two meshes of an indoor home environment.  Both meshes were generated by gathering color and depth frames from an Intel RealSense D435i camera and integrating them into a triangular mesh using methods from Ref. \cite{10.1145/2461912.2461919}  implemented in Open3D \cite{zhou2018open3d}. The method works by integrating the frames into a voxel grid and then performing marching cubes on the grid to create a triangular mesh. An important parameter in this process is the voxel size which if small makes a denser mesh that may also integrate noise from the sensor. The first mesh is of a basement and has a 5cm voxel size leading to a smoother approximation. Note this spacing is significantly higher than the noise of the sensor. The second mesh is of the main floor and is much larger and denser with 1cm voxel spacing leading to significantly more noise from the RGBD camera. Only the main floor mesh is post-processed with Laplacian filtering using Open3D. The basement mesh is composed of 60,620 triangles while the main floor mesh has 3,618,750 triangles. The parameters for Polylidar3D for both meshes are shown in Tables \ref{table:mesh_parameters_a} and \ref{table:mesh_parameters_b}. The most significant difference in parameter sets is that the basement is configured to capture small surfaces (lower $tri_{min}$ and $\gamma$) in comparison to the main floor.

\begin{table}[ht]
\centering
\caption{Polylidar3D Parameters for the Basement Mesh.}\label{table:mesh_parameters_a}
\begin{tabular}{@{}ll@{}}
\toprule
Algorithm        & Parameters                                                          \\ \midrule
FastGA           & level = 4,  $v_{min} = 2, d_{peak} = 0.1, sample_{pct} = 12\%$            \\
Plane/Poly Extr.      & $tri_{min}$ = 80, $ang_{min}$ = 0.95, $l_{max}$ = 0.1, $ptp_{max} = 0.08$, $vertices^{hole}_{min}$ = 6     \\
Poly. Filtering      & $\alpha = 0.01$, $\beta_{neg}$ = 0.025, $\beta_{pos}$ = 0.0, $\gamma$ = 0.07, $\delta$ = 0.05     \\ \bottomrule
\end{tabular}
\end{table}

\begin{table}[ht]
\centering
\caption{Polylidar3D Parameters for the Main Floor Mesh.}\label{table:mesh_parameters_b}
\begin{tabular}{@{}ll@{}}
\toprule
Algorithm        & Parameters                                                          \\ \midrule
FastGA           & level = 4,  $v_{min} = 2, d_{peak} = 0.1, sample_{pct} = 12\%$            \\
Plane/Poly Extr.      & $tri_{min}$ = 1000, $ang_{min}$ = 0.95, $l_{max}$ = 0.1, $ptp_{max} = 0.08$, $vertices^{hole}_{min}$ = 6     \\
Poly. Filtering      & $\alpha = 0.02$, $\beta_{neg}$ = 0.05, $\beta_{pos}$ = 0.02, $\gamma$ = 0.25, $\delta$ = 0.1    \\ \bottomrule
\end{tabular}
\end{table}

\begin{figure}[ht]
  \begin{subfigure}[t]{.45\linewidth}
    \centering\includegraphics[clip,trim=0cm 0cm 0cm 0cm,width=.99\linewidth]{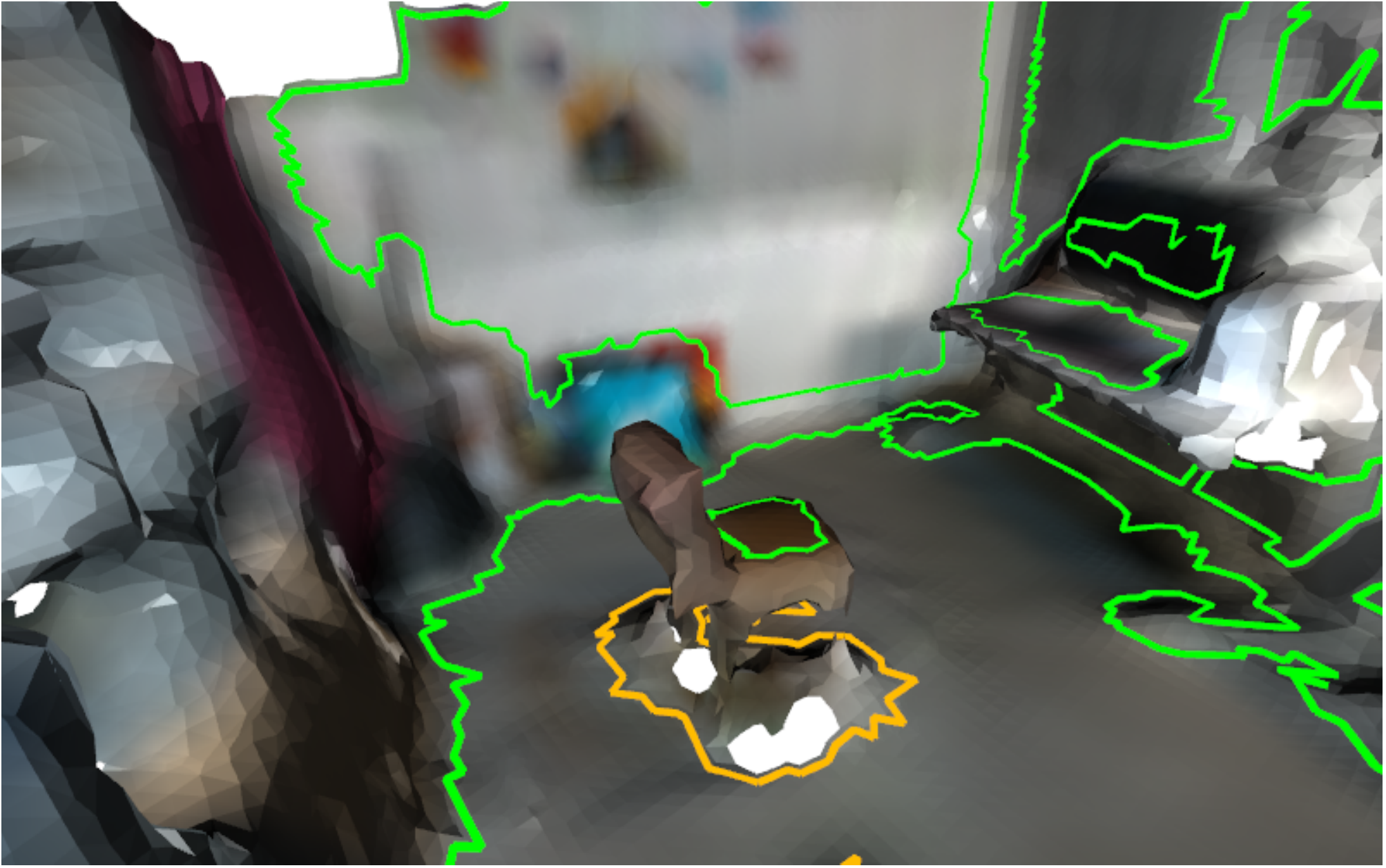}
    \caption{\label{fig:mesh_example_a}}
  \end{subfigure}
  \hfill
  \begin{subfigure}[t]{.45\linewidth}
    \centering\includegraphics[clip,trim=0cm 0cm 0cm 0cm,width=.99\linewidth]{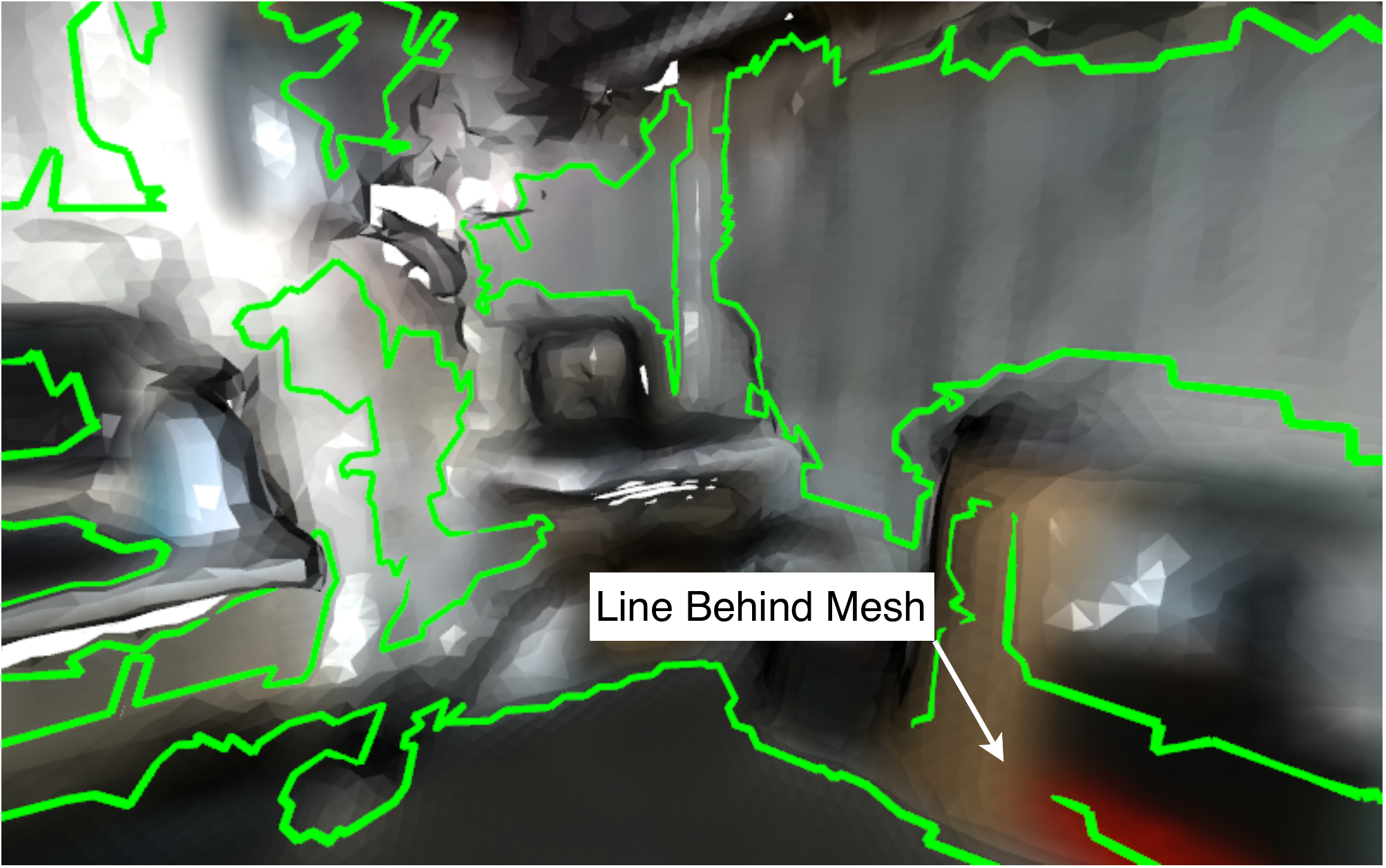}
    \caption{\label{fig:mesh_example_b}}
  \end{subfigure}
  \par\bigskip
  \begin{subfigure}[t]{.45\linewidth}
    \centering\includegraphics[clip,trim=0cm 0cm 0cm 0cm, width=.99\linewidth]{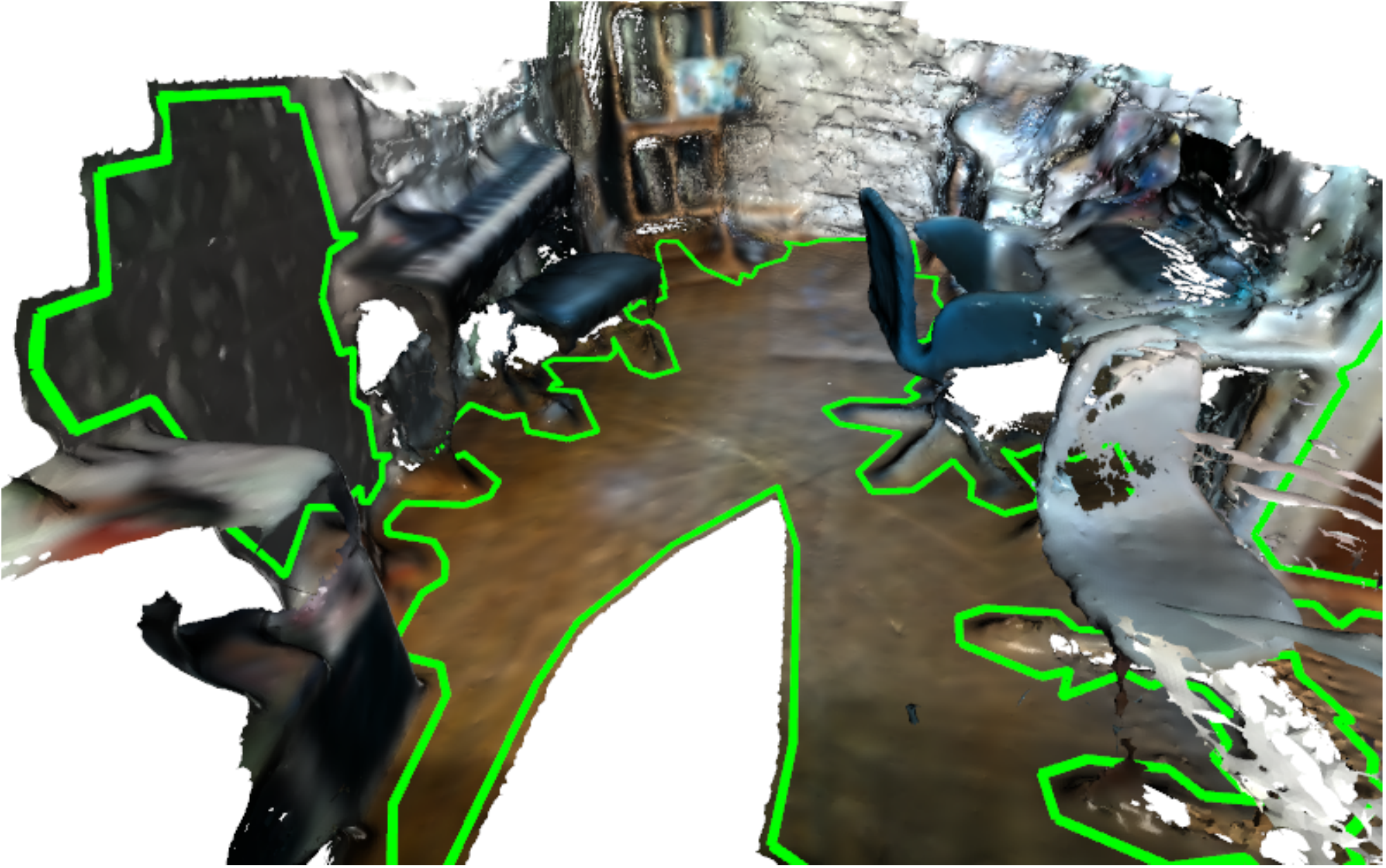}
    \caption{\label{fig:mesh_example_c}}
  \end{subfigure}
  \hfill
  \begin{subfigure}[t]{.45\linewidth}
    \centering\includegraphics[clip,trim=0cm 0cm 0cm 0cm,width=.99\linewidth]{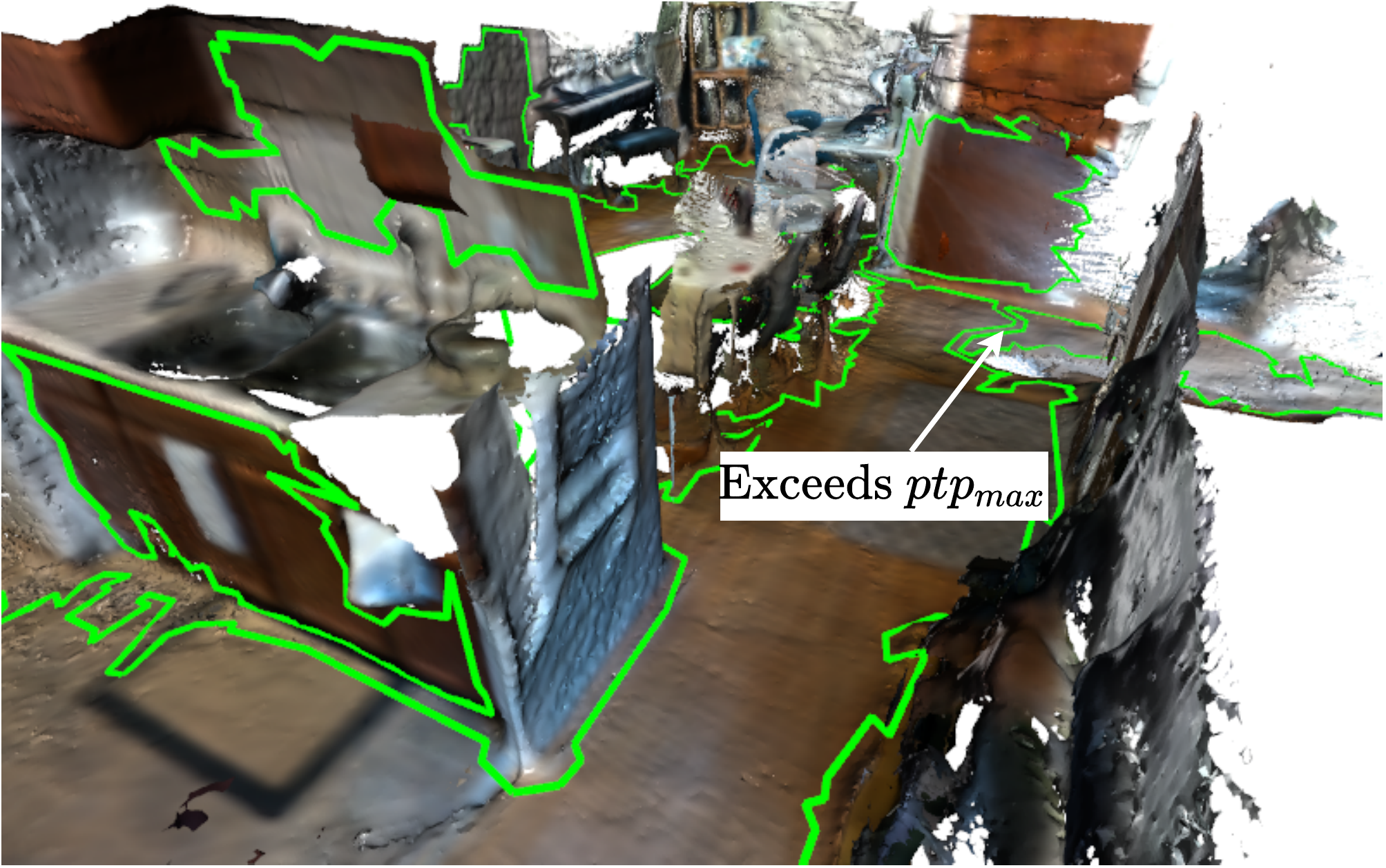}
    \caption{\label{fig:mesh_example_d}}
  \end{subfigure}
  \caption{Polygon extraction from meshes of an indoor home environment. Figures (\subref{fig:mesh_example_a}) and (\subref{fig:mesh_example_b}) show results on a basement mesh which has a smoother approximation and less noise.  Figures (\subref{fig:mesh_example_c}) and (\subref{fig:mesh_example_d}) show results on a significantly larger, denser, and nosier mesh of the main floor. }\label{fig:mesh_example}
\end{figure}

Figures \ref{fig:mesh_example_a} and \ref{fig:mesh_example_b} show the polygons output from Polylidar3D on the basement mesh.  The floor and all walls are appropriately captured as well as any obstacles on their flat surfaces. The top surface of the chair, table top, and monitor have also been captured. However there are several small planar segments on an occluded wall in (\subref{fig:mesh_example_b}) which may not be desirable for capture. In this same image a collection of stacked boxes are not truly flat and the polygon line segment goes ``behind`` the mesh surface.  Figures \ref{fig:mesh_example_c} and \ref{fig:mesh_example_d} show  polygons output for the dense first floor mesh. The floor and most walls have been successfully captured. However some walls have too much noise thus do not meet planarity constraints, e.g., the far wall in (\subref{fig:mesh_example_c}). The ground floor is not extracted as one continuous polygon instead separating at the edge of the mesh in (\subref{fig:mesh_example_d}).  This occurs because the floor areas have differences in height (in the mesh, not in reality); the point-to-plane distance constraint is exceeded between these two surfaces causing two extractions. This can be remedied by increasing $ptp_{max}$ by 1cm but is left here to highlight the issue.

\subsubsection{Parallelization Analysis}

\begin{figure}[ht]
  \begin{subfigure}[t]{.49\linewidth}
    \centering\includegraphics[width=.95\linewidth]{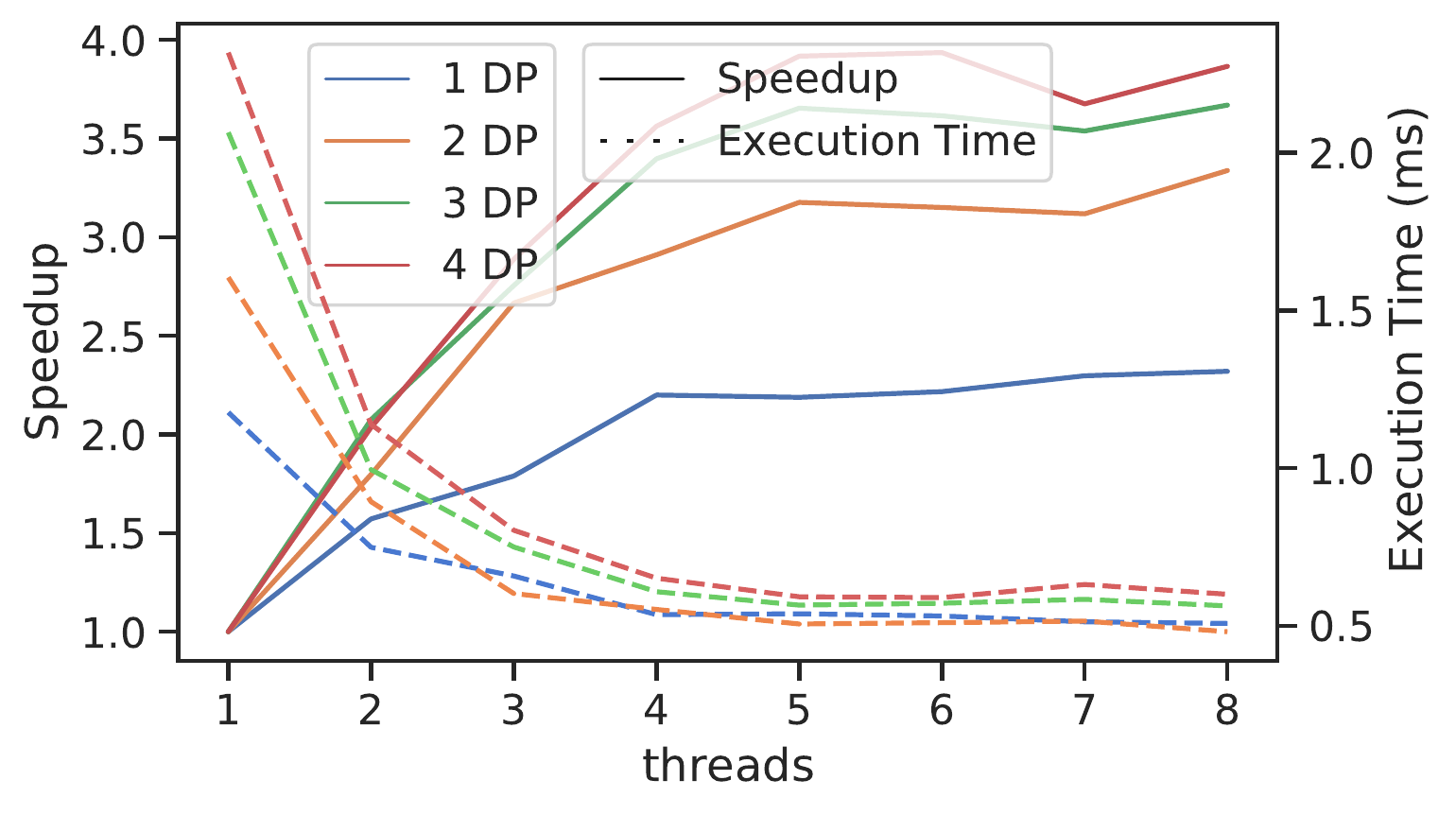}
    \caption{\label{fig:mesh_parallel_a}Basement (Sparse Mesh)}
  \end{subfigure}
  \begin{subfigure}[t]{.49\linewidth}
    \centering\includegraphics[width=.95\linewidth]{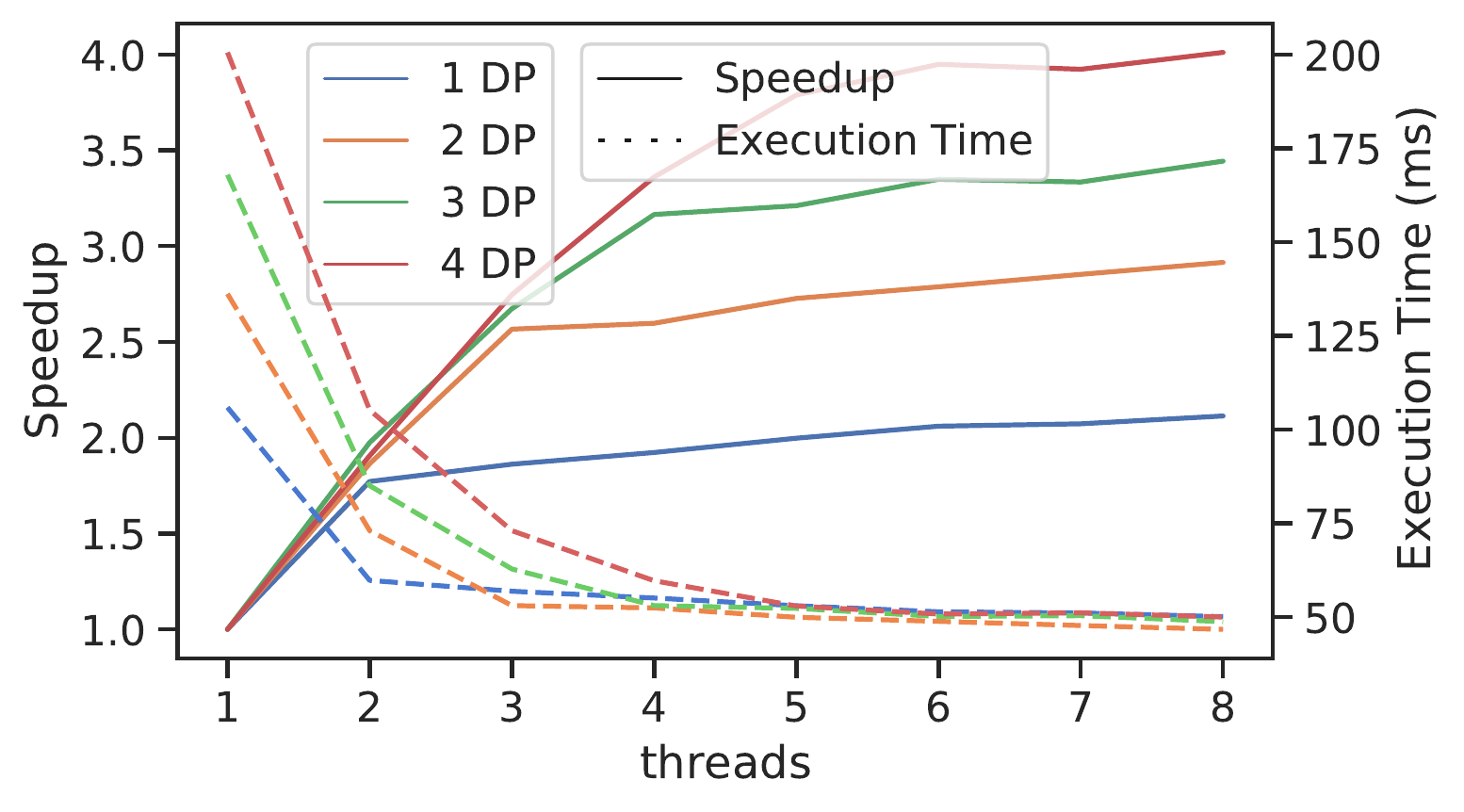}
    \caption{\label{fig:mesh_parallel_b}Main Floor (Dense Mesh)}
  \end{subfigure}
  \caption{Results of parallel speedup and execution timing of plane/polygon extraction with up to eight threads. Both basement (\subref{fig:mesh_parallel_a}) and main floor meshes (\subref{fig:mesh_parallel_b}) are analyzed. Solid lines indicate parallel speedup and link to the left $y$-axis while the dashed lines indicate execution time and link to the right $y$-axis. The color indicates number of dominant plane (DP) normals extracted.}\label{fig:mesh_parallel}
\end{figure}

This section explores how Polylidar3D scales with additional CPU threads. We specifically focus on plane/polygon extraction in Polylidar3D's back-end. Both meshes are used in these steps and we limit the dominant plane normals to the top four in the scene, i.e., only floors and walls are extracted. Figure \ref{fig:mesh_parallel} shows the parallelization speedup and execution timing of plane/polygon extraction as up to eight threads are provided. The color of the line indicates how many dominant plane normals are requested for extraction, i.e., blue indicates only the floor while orange indicates both the floor and one wall. The more dominant plane normals requested the more CPU cycles are needed.  

Figure \ref{fig:mesh_parallel_a} shows the speedup and execution timings of the sparse basement mesh. The parallel speedup does not go any higher than 2.4 with one dominant plane normal (blue-solid) and reaches approximately 4.0 with four dominant planes (red-solid). The execution timings (dashed lines) clearly show the diminishing returns as more threads are provided and plateaus around 0.5ms at 4 threads. Figure \ref{fig:mesh_parallel_b} shows a similar trend for the much larger and more dense main floor mesh. The trends are clear that greater speedup is possible as more unique dominant planes normals are requested because this work gets partitioned to independent tasks. However there is a  limit to this parallelism as not all procedures within the tasks are themselves parallelized. This is a clear example of Amdahl's law in effect which explains a theoretical limit to speedup as a function of the percent of a program that is actually parallelizable \cite{10.1145/1465482.1465560}. In essence the theoretical speedup is always limited by the serial tasks, which in our case becomes (roughly) the combined execution time of planar segmentation of the single largest dominant plane normal and the polygon extraction of its largest planar segment. New threads do not reduce the time to complete these tasks because their algorithms are serial.

\section{Discussion}\label{sec:discussion}



Results show Polylidar3D successfully extracts flat surfaces as polygons with interior holes from unorganized/organized point clouds and user-provided meshes. One of Polylidar3D's primary strengths is its polygon extraction speed.  The key to this speed is the fast construction of half-edge triangular meshes used directly in polygon extraction. No secondary re-triangulation is necessary after planar segmentation. Our Fast Gaussian Accumulator was benchmarked against competing $K$-$D$ tree methods and shown to be two times faster and effective at identifying dominant plane normals. Data and task-based parallelism is also exploited to efficiently allocate work to available CPU cores.



Results also illustrate limitations. First, rooftop and ground extraction in Section \ref{sec:results_unorganized} shows that only one plane normal can be extracted from unorganized 3D points clouds. As described in our methods the front-end currently performs 2.5D Delaunay triangulation which requires 3D $\rightarrow$ 2D projection. This projection is most suitable when the sensing viewpoint and flat surface of interest are aligned, as is for airborne LiDAR point clouds. However this is not a hard requirement as shown with ground detection from the KITTI dataset. We chose 2.5D Delaunay triangulation for its speed, however other methods may be used such as the ball pivot algorithm \cite{bernardini_ball-pivoting_1999} or Poisson surface reconstruction \cite{kazhdan_screened_2013}. These methods created 3D meshes which could then be processed by Polylidar3D.

Polylidar3D planar segmentation expects a mesh to be reasonably smoothed. The amount of smoothing depends on user-specified parameters for surface extraction and the noise of the input data. If only large distinct flat surfaces are required then minimal smoothing is necessary. We define distinct surfaces as plane normals that are well-separated on the Gaussian Accumulator (e.g., $90^\circ$). This smoothing aids in GA peak detection and appropriately groups triangles during planar segmentation. 

The Fast Gaussian accumulator can only detect plane normals; it currently has no concept of origin offset.  This means that if there are two flat surfaces separated far from each other in a scene, with similar (but not the same) surface normals, it is possible they will appear near each other on the GA and be merged. Noise in the mesh affects how close these two peaks can be on the GA and still be detected as distinct peaks. As the mesh is further smoothed (with edge-preservation filters) the noise is reduced and the peaks become more defined. This is exactly what had to be done to detect the numerous noisy planes in the SynPEB benchmark. Also, group assignment in Algorithm \ref{alg:group_assignment} will assign common triangles to these detected peaks if they meet a user defined angular threshold $ang_{min}$ from a detected peak. This means any detected peaks should be greater than $2\cdot arcos(ang_{min})$ from each other to guarantee no overlap. Note also that $ang_{min}$ can be increased as the mesh is smoothed.

Only dominant planes, flat surfaces that account for most of the 3D data, can be reliably captured from organized point clouds and user-provided meshes. We see numerous qualitative examples of this from RGBD sensor data, the SynPEB benchmark, and user-provided meshes. Polylidar3D is only able to extract 48.6\% of the average 42.6 planes in the SynPEB test scenes. However the percent of point cloud metric $k$ at 82\% shows Polylidar3D doing an excellent job of capturing large dominant planes.  Scenes in this benchmark are the antithesis for what Polylidar3D was designed for (dozens of small noisy planes), yet we show Polylidar3D still performs well in important metrics such as minimizing the number of over/under-segmented planes, spurious predictions, and execution time. We believe these metrics taken as a whole demonstrate Polylidar3D's efficiency and reliability for polygon extraction of dominant planar surfaces.

\subsection{Future Work}\label{sec:dicsussion_future_work}

There are three significant techniques that will improve Polylidar3D's robustness in future work: polygon merging, time integration, and integrating intensity/color data. Many planar segmentation algorithms perform ``plane`` merging of extracted segments (point sets) which are deemed similar by Euclidean distance and plane-fit error tolerance \cite{feng_fast_2014,oesau_planar_2016, trevor2013efficient, biswas_planar_2012}. This is most often used to combine oversegmented predictions of a common surface. Polylidar3D can be extended to perform the same action with polygons. Detailed meta-data about each polygon can be stored to aid in the merging process including geometric plane normal, centroid, axis-aligned bounding box, and even the convex hull if necessary. This information will aid the pairwise matching between polygons in a scene before a possibly expensive polygon merger. There are several methods to perform a non-convex polygon merge including morphological operations such as dilation and erosion.

Polylidar3D processes each point cloud distinctly. Time integration incorporates data from multiple data frames in a sequence by filtering and refining extracted polygons based on previous results. In a static scene with fixed sensor viewpoint time integration can reduce the variance of polygons produced over time. All linear rings of the polygon (both hull and holes) can be explicitly tracked using meta-data previously discussed and removed if certain thresholds are not met.  With a dynamic scene or moving sensor time integration would require significant extension to Polylidar3D to incorporate additional data such as sensor (vehicle) motion estimates and even semantic scene information. Additional work investigating the use of Bayesian filtering will be done.

Data such as intensity and/or color of the point cloud can be used to further determine similarity between neighbors in the point cloud during region growing. Such data has been shown to improve results for point cloud registration \cite{rusinkiewicz_efficient_2001} and mesh smoothing \cite{lee_fast_2013}. Additionally, deep neural network may perform semantic segmentation on RGBD images to quickly output class labels for each pixel in the image \cite{pham_scenecut_2018}. This information can then be fused into Polylidar3D to better inform partitioning of work and similarity between neighboring triangles.



\section{Conclusions}\label{sec:conclusion}

This paper introduced Polylidar3D, a non-convex polygon extraction method  capturing flat surfaces from a variety of 3D data sources. Front-end methods transform unorganized point clouds, organized point clouds, and 3D triangular meshes to a common half-edge triangular mesh format. Back-end core algorithms perform mesh smoothing, dominant plane normal estimation, planar segmentation, and polygon extraction. A novel Gaussian accumulator, FastGA, was demonstrated robust and quick at detecting dominant plane normals in a 3D scene. These dominant plane normals are used to parallelize planar segmentation and polygon extraction. Polylidar3D is evaluated in five separate experiments with airborne LiDAR point clouds, automotive LiDAR point clouds, RGBD videos, synthetic LiDAR benchmark data, and meshes of indoor environments. Qualitative and quantitative results demonstrate Polylidar3D's speed and versatility. All of Polylidar3D is open source and available to be freely used and improved upon by the community \cite{polylidarcode, fastgacode, organizedpointfilters}.

\vspace{6pt} 



\authorcontributions{Conceptualization, J.C.; methodology, J.C.; software, J.C.; validation, J.C.; formal analysis, J.C.; investigation, J.C. and E.A.; resources, J.C. and E.A.; data curation, J.C; writing--original draft preparation, J.C.; writing--review and editing, J.C. and E.A.; visualization, J.C.; supervision, E.A.; project administration, E.A.; funding acquisition, E.A. All authors have read and agreed to the published version of the manuscript.}

\funding{This work was supported in part by NSF I/UCRC Award 1738714.}

\acknowledgments{The authors would like to acknowledge the input and support of Prince Kuevor of the University of Michigan Robotics Institute.}

\conflictsofinterest{The authors declare no conflict of interest.} 

\abbreviations{The following abbreviations are used in this manuscript:\\

\noindent 
\begin{tabular}{@{}ll}
MDPI    & Multidisciplinary Digital Publishing Institute\\
OPC     & Organized Point Cloud  \\
AHC     & Agglomerative Hierarchical Clustering \\
OGC     & Open Geospatial Consortium \\
GA      & Gaussian Accumulator \\
CPU     & Central Processing Unit \\
GPU     & Graphics Processing Unit \\
RAM     & Random Access Memory \\
GPGPU   & General Purpose GPU \\
SDK     & Software Development Kit \\

\end{tabular}}

\appendixtitles{no} 


\reftitle{References}


\externalbibliography{yes}
\bibliography{references.bib}





\end{document}